\documentclass[9pt,twocolumn,twoside]{osajnl_patched}
\journal{ol}

\usepackage{times}
\usepackage{dblfloatfix}
\usepackage{amsmath}
\usepackage{amssymb}
\usepackage{multirow}
\usepackage{multirow}
\usepackage{mathtools}
\usepackage{subfloat}
\usepackage{float}
\usepackage{xcolor}
\usepackage{url}
\usepackage{bbm}
\usepackage{algorithm}
\usepackage{algpseudocode}
\usepackage{pbox}
\usepackage{placeins}
\usepackage{bm}
\usepackage{acronym}
\usepackage{enumitem}

\usepackage{booktabs}
\usepackage{graphicx}
\usepackage{caption}
\usepackage{subcaption}
\usepackage{newfloat}
\usepackage{times}
\usepackage{amsmath}
\usepackage{amssymb}
\usepackage{caption}
\usepackage{bm}
\usepackage{url}
\usepackage{hyperref}
\usepackage[capitalise]{cleveref}
\usepackage{amsmath}
\usepackage{empheq}
\usepackage{multirow}
\usepackage{bm}
\usepackage{graphicx}
\usepackage{amssymb}
\usepackage[round-pad=false]{siunitx}
\usepackage{pifont} 
\usepackage{booktabs,makecell}
\setboolean{shortarticle}{false}

\usepackage{newfloat}
\usepackage{multicol}

\renewcommand{\v}{\mathbf{v}}
\renewcommand{\S}{\mathbf{S}}

\newcommand{\papertitle}{\bfseries \boldmath HuMBLE: Human Motion-Driven Behavior Learning for Embodied Locomotion}
\newcommand{\methodtitle}{\emph{HuMBLE}}
\newcommand{\numsamples}{528,214}
\newcommand{\tabularasa}{\emph{Tabula Rasa}}
\newcommand{\highlight}[1]{#1}
\newcommand{\customref}[2]{\hyperref[#1]{\Cref*{#1}#2}}
\newcommand{\piteacher}{\bm{\pi}_{\text{teacher}}}
\newcommand{\pistudent}{\bm{\pi}_{\text{prior}}}
\newcommand{\pideploy}{\bm{\pi}_{\text{deploy}}}
\newcommand{\pibaselineone}{\bm{\pi}_{\text{baseline}}}

\newcommand{\vteacher}{\bm{V}_{\text{teacher}}}
\newcommand{\vstudent}{\bm{V}_{\text{prior}}}
\newcommand{\vdeploy}{\bm{V}_{\text{deploy}}}

\newcommand{\action}{\bm a}

\newcommand{\numimu}{n_{\text{IMU}}}
\newcommand{\numjoint}{n_j}
\newcommand{\numaction}{n_a}
\newcommand{\numee}{n_e}

\newcommand{\position}{\bm r}
\newcommand{\gravity}{\bm g_{\mathcal{I}}}
\newcommand{\gravityref}{\hat{\bm g}_{\mathcal{I}}}
\newcommand{\linvel}{\bm v}
\newcommand{\angvel}{\bm \omega}
\newcommand{\jointpos}{\bm q_j}
\newcommand{\jointposref}{\hat{\bm q}_j}
\newcommand{\jointvel}{\dot{\bm q}_j}
\newcommand{\jointtorque}{{\bm \tau}_j}
\newcommand{\jerk}{\bm \zeta}

\newcommand{\odep}{\bm{o}_{\text{deploy}}}
\newcommand{\oteacher}{\bm{o}_{\text{teacher}}}
\newcommand{\oprop}{\bm{o}_{\text{prop}}}
\newcommand{\ocmd}{\bm{o}_{\text{cmd}}}
\newcommand{\oref}{\bm{o}_{\text{ref}}}
\newcommand{\opriv}{\bm{o}_{\text{priv}}}
\newcommand{\ocritic}{\bm{o}_{\text{critic}}}
\newcommand{\osup}{o_{\text{sup}}}
\newcommand{\oterrain}{\bm{o}_{\text{terrain}}}

\newcommand{\rfinetune}{r_{\text{fine-tune}}}
\newcommand{\rref}{r_{\text{ref}}}
\newcommand{\rteacher}{r_{\text{teacher}}}
\newcommand{\rgoal}{r_{\text{goal}}}
\newcommand{\rtrack}{r_{\text{track}}}
\newcommand{\rcommand}{r_{\text{cmd}}}
\newcommand{\rsurvival}{r_{\text{survival}}}
\newcommand{\rreg}{r_{\text{reg}}}
\newcommand{\rfinetunestructure}{
\rfinetune =
    \begin{cases}
        \rref   & \text{if env} \in {\envref} \\
        \rgoal  & \text{if env} \in {\envgoal}.
    \end{cases}
}

\newcommand{\ei}{\bm{e}_i}
\newcommand{\einormalized}{\tilde{\bm{e}}_i}

\newcommand{\setwo}{SE(2)}

\newcommand{\envref}{\mathbb{E}_{\text{ref}}}
\newcommand{\envgoal}{\mathbb{E}_{\text{goal}}}
\newcommand{\similarity}{s}

\newcommand{\teasercaption}{
\textbf{Summary of hardware results from \methodtitle.} \textbf{(A)} Human-like steerable locomotion on the Boston Dynamics Atlas R1 was demonstrated live to the public on stage. \textbf{(B)} The Atlas R1 robot following real-time user steering commands that vary continuously in speed and direction while maintaining stylistic consistency with the human reference motion dataset. \textbf{(C)} The Atlas D1 robot performs backward walking, circling, and sidestepping (from left to right) to demonstrate the \methodtitle~policy's coverage of the \setwo~steering command space. \textbf{(D)} Transition from walking to high-speed sprinting on the Unitree G1 robot. The gait transition emerges in response to an increasing in the commanded speed. \textbf{(E)} Integration of the \methodtitle~policy as an underlying locomotion layer within hierarchical control stacks for autonomous navigation (left) and box parkour (right) tasks.
}

\newcommand{\stroboscopiccaption}{
\textbf{Stroboscopic snapshot of Atlas R1 walking forward under different controllers.} We contrast the forward walking gaits, initiated from a standing configuration, across three different locomotion controllers: \textbf{(A)} the \methodtitle~policy, \textbf{(B)} an MPC controller, and \textbf{(C)} a \tabularasa~RL policy. Each frame captures the transition from swing phase to ground contact to highlight the synchronized arm swinging, heel-strike, and knee-lock.
}

\newcommand{\datafidelitycaptionnew}{
\textbf{Data fidelity analysis.} \textbf{(A)} Mean absolute error (MAE) of the torso linear velocity, angular velocity, and joint positions relative to the reference motion data, comparing the teacher and the final \methodtitle~deployment policies. Each bar illustrates the cumulative tracking error decomposed into linear velocity (\SI[per-mode=repeated-symbol]{}{\metre\per\second}), angular velocity (\SI[per-mode=repeated-symbol]{}{\radian\per\second}), and joint positions (\SI{}{\radian}). The MAE is calculated by averaging over all individual state dimensions and over the trajectory length following temporal alignment via DTW. \textbf{(B)} Comparison of the reference motion, full-body-reference-conditioned teacher policy, and command-conditioned deployment policy for \emph{Fast Forward Walk}. All plots display raw time-series data without applying DTW, where the x-axis represents time in seconds. Trajectories illustrate \setwo~torso velocities alongside selected joint angles chosen to highlight key biomimetic traits such as arm swing, hip sway, knee extension, and heel-to-toe rolling present in the human data and learned by the \methodtitle~policy.
}

\newcommand{\quantitativecaption}{
\textbf{Atlas R1 quantitative performance analysis.} We evaluate and compare the command responsiveness, command-tracking accuracy, and robustness to external pushes for the \methodtitle~and the \tabularasa~policies. \textbf{(A)} Relative command-tracking error in steady-state. The \setwo~command space (longitudinal velocity, lateral velocity, and turning rate) is visualized using three 2D slices, zeroing one dimension per column, from left to right: $\omega_{\text{turn}}=0$, $v_{\text{lat}}=0$ and $v_{\text{long}}=0$. Each cell in a plot compares the \tabularasa~baseline (outer square) to the \methodtitle~policy (inner circle), with lighter colors indicating better performance (lower tracking error or faster rise time). \textbf{(B)} Rise time in seconds. Similarly visualized as in \textbf{(A)}. \textbf{(C)} Robustness to external pushes. We evaluate stability against planar force pushes across three locomotion scenarios. The shaded regions approximate the successful recovery envelopes for the \methodtitle~policy (orange) and the \tabularasa~baseline (purple). A $\bigcirc$ symbol indicates successful push recovery, while a $\boldsymbol{\times}$ indicates a fall.
}

\newcommand{\ablationstudycaption}{
\textbf{Ablation study.} We conducted ablation studies of three crucial design choices within \methodtitle. \textbf{(A)} Role of Teacher--Student Distillation. We compare the data fidelity of the \methodtitle~prior policy obtained from the Teacher--Student distillation stage (orange) against a single-stage RL baseline (gray) that directly trains a command-conditioned policy. The DTW-aligned reference motion tracking errors are visualized as a stacked bar chart of the component-wise Mean Absolute Error (MAE). \textbf{(B)} Efficacy of Multi-Task RL. We evaluate data fidelity and steering command-tracking performance across varying environment allocation ratios for four canonical modalities, validating the efficacy of the multi-task RL framework in balancing these two competing objectives. \textbf{(C)} Impact of RL Fine-Tuning. We assess command-tracking accuracy and command coverage by measuring the steady-state relative command-tracking error of the final \methodtitle~deployment policy (inner circle) following RL fine-tuning, compared directly against the prior policy (outer square) from which fine-tuning was initialized. The heatmap visualizes the relative command-tracking error in steady-state for $\omega_{\text{turn}}=0$, $v_{\text{lat}}=0$, and $v_{\text{long}}=0$ from left to right. Gray squares indicate regions where the prior policy fails to robustly track the corresponding velocity command, resulting in a robot fall.
}

\newcommand{\simtorealcaption}{
\textbf{Sim-to-real analysis.} A set of representative canonical commands is employed to evaluate the sim-to-real transfer of the \methodtitle~policy. The command elements $v_{\text{long}}$, $v_{\text{lat}}$, $\omega_{\text{turn}}$, have units of m/s, m/s, and rad/s respectively. Each scenario starts with the robot standing at zero command. The command is then stepped to the target and held for a duration before stepping back to zero. We report the mean absolute error (MAE) over the entire scenario as well as the steady-state error. The linear and angular components of the tracking error are reported separately with units of \SI[per-mode=repeated-symbol]{}{\metre\per\second} and \SI[per-mode=repeated-symbol]{}{\radian\per\second}, respectively. We define the linear tracking error as the max of the longitudinal and lateral tracking errors.
}

\title{\papertitle} 

\author[1$^\dagger$]{Mike Zhang}
\author[1$^\dagger\ast$]{Dongho Kang}
\author[2$^\dagger$]{Kevin Bergamin}

\author[1$^\ddagger$]{Nicola Burger}
\author[2$^\ddagger$]{Robin Deits}
\author[1$^\ddagger$]{Jonathan Foster}
\author[1$^\ddagger$]{Bilal Hammoud}
\author[1$^\ddagger$]{Katie Hughes}
\author[1$^\ddagger$]{Francesco Iacobelli}
\author[2$^\ddagger$]{Twan Koolen}
\author[1$^\ddagger$]{M. Eva Mungai}
\author[1$^\ddagger$]{Zach Nobles}
\author[2$^\ddagger$]{Shane Rozen-Levy}
\author[1$^\ddagger$]{Jean Pierre Sleiman}
\author[1$^\ddagger$]{Fangzhou Yu}
\author[1$^\ddagger$]{Yunbo Zhang}

\author[1$^{\ddagger\ddagger}$]{Alfred Rizzi}
\author[1,3$^{\ddagger\ddagger}$]{Jessica Hodgins}
\author[2$^{\ddagger\ddagger}$]{Scott Kuindersma}
\author[2$^{\ddagger\ddagger}$]{Yeuhi Abe}
\author[1$^{\ddagger\ddagger}$]{Sylvain Bertrand}
\author[1$^{\ddagger\ddagger}$]{Farbod Farshidian}

\affil[1]{RAI Institute, USA}
\affil[2]{Boston Dynamics, USA}
\affil[3]{Carnegie Mellon University, USA}

\affil[$^\dagger$]{Core contributors are listed in order of relative contribution.}
\affil[$^\ddagger$]{Additional contributors are listed alphabetically.}
\affil[$^{\ddagger\ddagger}$]{Project leads are listed in reverse order of relative contribution.}
\affil[$^\ast$]{Corresponding author. Email: \texttt{dkang@rai-inst.com}}

\begin{abstract}
\textbf{Despite recent advances in humanoid locomotion, controllers optimized for command tracking and robustness tend to produce mechanical gaits, whereas controllers tied to human motion data often fail to generalize to commands outside the data distribution. This work introduces a learning framework that balances these competing objectives to synthesize real-time steerable, robust, and biomimetic locomotion policies from human data. Using an in-house curated locomotion dataset covering diverse speeds and directions, we first learn a natural locomotion prior policy through a teacher--student distillation process. Specifically, we train a full-body reference-conditioned policy with Reinforcement Learning (RL), then distill it into a lightweight prior policy conditioned solely on proprioception and a planar torso-velocity steering command. Next, we fine-tune the prior policy with multi-task RL to expand command coverage and robustness beyond the data distribution, pairing a goal-conditioned task that tracks arbitrary commands with a reference-guided task that tracks the human data as an explicit style regularizer. We validate our framework on three humanoid robots: the Boston Dynamics Atlas R1, Atlas D1, and Unitree G1.
Experimental results demonstrate robust performance across real-world scenarios, including direct user-controlled locomotion in indoor and outdoor environments, and integration as the locomotion layer within hierarchical control stacks. Benchmarks against \tabularasa~RL policies trained without human data and ablation studies confirm that our framework yields a lightweight, deployable policy that reconstructs coordinated whole-body behavior from a steering command, retaining the human gait characteristics while remaining robust and fully steerable. 
}
\end{abstract}

\DeclareFloatingEnvironment{Movie}

\begin{document}

\maketitle

\newcommand{\isArXiv}{1}


\section*{INTRODUCTION}

Enabling robots to work alongside people in shared environments has long motivated the field of humanoid robotics.
Within this broader goal, humanoid locomotion has been a central benchmark because walking is one of the most visible and socially salient forms of embodied behavior.
This pursuit has shaped generations of robotic research, from the pioneering, cautious footsteps of \emph{Honda ASIMO} in the early 2000s \cite{Sakagami2002ASIMO} to the robust and energetic gait of \emph{Boston Dynamics PETMAN} in the 2010s \cite{Nelson2012Petman}. 
Beyond aesthetic appeal, human-mimetic locomotion can provide familiar kinematic and social cues, such as step timing, body motion, and velocity transitions, that people can interpret through prior social and sensorimotor experience from human-human interaction \cite{Duffy2003Anthropomorphism, Chaminade2005Motor}.
As a result, such motions can make robot behavior easier to anticipate, interpret, and coordinate with, thereby improving legibility, trust, and usability in human-robot collaborative environments \cite{Dragan2013Legibility, Dragan2015Effects}.

However, achieving human-like locomotion remains an open challenge, particularly when natural full-body motion must be steered by high-level commands in real-time and transferred reliably to physical hardware.
This requirement exposes a fundamental trade-off, wherein controllers optimized primarily for command tracking and disturbance recovery often produce mechanical, unnatural gaits, whereas controllers tightly constrained to recorded human motion data can lose responsiveness when commands require transitions, speeds, or direction combinations that are underrepresented in the reference motion dataset used for development.
A deployable controller must satisfy three requirements simultaneously: (1) it must preserve the stylistic quality of human locomotion and natural gait transitions across speeds and directions; (2) it must respond accurately and stably to a broad range of user-specified velocity commands; and (3) it must execute with sufficiently low latency and modest memory and compute requirements for onboard deployment.
To address this challenge, we introduce \methodtitle~(\textbf{Hu}man \textbf{M}otion-driven \textbf{B}ehavior \textbf{L}earning for \textbf{E}mbodied Locomotion).
\methodtitle~is a learning-based framework that distills human-motion priors into a compact command-conditioned locomotion policy and subsequently expands command coverage while preserving stylistic consistency.
At runtime, the resulting policy maps proprioceptive observations and \setwo~velocity commands directly to joint-level actions, without requiring reference motions, latent motion sampling, or computationally intensive generative inference.


This broader pursuit of stable, responsive, and natural humanoid locomotion has led to a long line of work based on first-principles approaches to control and motion generation.
Historically, these approaches have included heuristic motion generation \cite{Raibert1991Animation, Coros2010Generalized}, high-level motion features \cite{Lasa2010Feature}, and biomechanically derived control objectives \cite{wang2012optimizing}.
These principles have shaped modern control frameworks, including model-based control frameworks \cite{Kuindersma2014QP, Kuindersma2016AtlasWBC, Chignoli2020MITHumanoid} and reward-engineered Reinforcement Learning (RL) approaches \cite{Radosavovic2024Real, seo2025learning}.
Such methods have demonstrated stable, responsive locomotion on humanoids, particularly for command tracking, balance, and disturbance recovery.
However, encoding the full variability of human gait using hand-designed objectives remains difficult, given that human-like walking depends on highly coupled full-body coordination, subtle nuances in contact timing, arm and leg swing trajectories, posture, and speed- and direction-dependent gait transitions.
As a result, controllers optimized primarily through analytic models or hand-crafted objectives may achieve strong tracking and recovery while failing to reproduce the stylistic variability and coordination patterns present in human locomotion.

An alternative approach uses human motion data directly as a source of supervision for producing naturalistic behavior. Human motion data is relatively easy to collect at scale and has long been central to character animation, from motion graphs and interactive motion synthesis \cite{Kovar2002MotionGraphs, Arikan2002Interactive, Lee2002Interactive} to modern kinematic methods such as \emph{Motion Matching} \cite{Clavet2016Motion, Bergamin2019DReCon, Holden2020Learned}. 
Recent work has increasingly shifted toward deep generative models, including Variational Autoencoders (VAEs) \cite{Kingma2014Auto}, Generative Adversarial Networks (GANs) \cite{Goodfellow2020GAN}, diffusion models \cite{Ho2020Diffusion}, and flow-matching models \cite{lipman2023flow}. In this setting, motion data is treated as samples from an underlying reference distribution rather than as target trajectories to be tracked. This paradigm shifts the objective from reproducing specific sequences to aligning the distribution of synthesized behaviors with that of the reference dataset. VAEs, in particular, have been widely used to embed high-dimensional motion into structured latent spaces that can be explored by higher-level policies \cite{Ling2020Character, Won2022Physics, Yao2022ControlVAE, Starke2024Categorical, luo2026sonic}. Related ideas have also been adapted to legged robots, either by coupling VAE-based motion generators with RL motion trackers \cite{kang2025learning} or by conditioning policies on learned gait embeddings \cite{LiuSkillLatent2025}. Despite their interpretability and controllability, VAE-based approaches can struggle to reproduce the kinematic structure and contact consistency of the demonstrated motions, resulting in artifacts such as foot slippage and unnatural joint configurations that undermine physical plausibility of the motions.

Adversarial Imitation Learning (AIL) provides an alternative approach for directly training physically grounded controllers using motion data.
Instead of first synthesizing kinematic motions and then tracking them with a separate controller, AIL methods train policies using a discriminator reward that evaluates how closely the executed behavior matches the demonstrations, optimizing until the two are indistinguishable.
The Adversarial Motion Prior (\emph{AMP}) \cite{Peng2021AMP} framework is a representative example wherein an RL policy is optimized using both task rewards for command tracking and a discriminator-based style reward derived from motion data.
This formulation has been applied to achieve stylized and steerable locomotion on quadrupedal and humanoid robots while avoiding an explicit kinematic motion-generation stage \cite{Escontrela2022Adversarial, Tang2024HumanMimic, shi2025adversarial}.
Extensions such as \emph{ASE} \cite{Peng2022ASE} and \emph{CALM} \cite{Tessler2023CALM} further introduce skill embeddings and hierarchical control to mitigate mode collapse and preserve behavioral diversity.
However, adversarial objectives introduce a min-max training problem that can be sensitive to reward scaling, discriminator design, and hyperparameter selection. Additionally, the learned style reward offers only indirect influence over the resulting behavior, which complicates fine-grained user control over emergent behaviors.

More recently, diffusion and flow-matching models have emerged as more expressive alternatives for modeling multimodal motion distributions, with conditioning mechanisms based on partial states \cite{Shi2024Interactive, huang2025diffuse}, pose or velocity commands \cite{liao2026beyondmimic}, or text prompts \cite{Tevet2023Human}.
A growing body of work has begun extending these architectures to physical control by mapping directly from robot states and commands to actions \cite{Truong2024PDP, Huang2025DiffuseLoco, huang2025diffuse}.
Although these approaches leverage expressive generative architectures to preserve motion diversity and eliminate separate tracking layers to streamline deployment, the underlying iterative sampling or integration procedures remain difficult to reconcile with the strict latency, computational, and power constraints of real-time onboard control.

The approaches discussed above show that motion data can provide rich priors for natural locomotion. However, adversarial or generative formulations introduce training instabilities, indirect behavioral control, or runtime inference overheads that are prohibitive for low-latency onboard humanoid control.
Instead, we cast human-motion-driven locomotion learning as a Partially Observable Markov Decision Process (POMDP), wherein full-body reference states are available as \emph{privileged} information during training but are not provided to the final deployment policy.
Rather than explicitly tracking a full-body reference trajectory at runtime, the deployment policy learns to reconstruct coordinated full-body behavior from proprioceptive observations and user \setwo~velocity commands, mapping these inputs directly to joint-level actions.
Generally, teacher--student architectures have proven effective in addressing POMDP formulations in legged robotics, particularly for rough-terrain locomotion where terrain geometry is only partially observable \cite{Miki2022WildLocomotion, Cheng2023ExtremeParkour} or unobservable on hardware \cite{RL_Joonho}. More closely aligned with our work, similar principles were used to learn diverse human-like behaviors conditioned on high-level user commands from human motion data in physically simulated characters \cite{Tessler2024MaskedMimic}.
Inspired by this approach, we distill full-body motion priors from human motion data into a single, compact, command-conditioned policy, avoiding both training instabilities and computationally intensive generative modules at deployment.

%
%
\ifdefined\isArXiv
\begin{figure*}[t!]
    \centering
    \includegraphics[width=\textwidth]{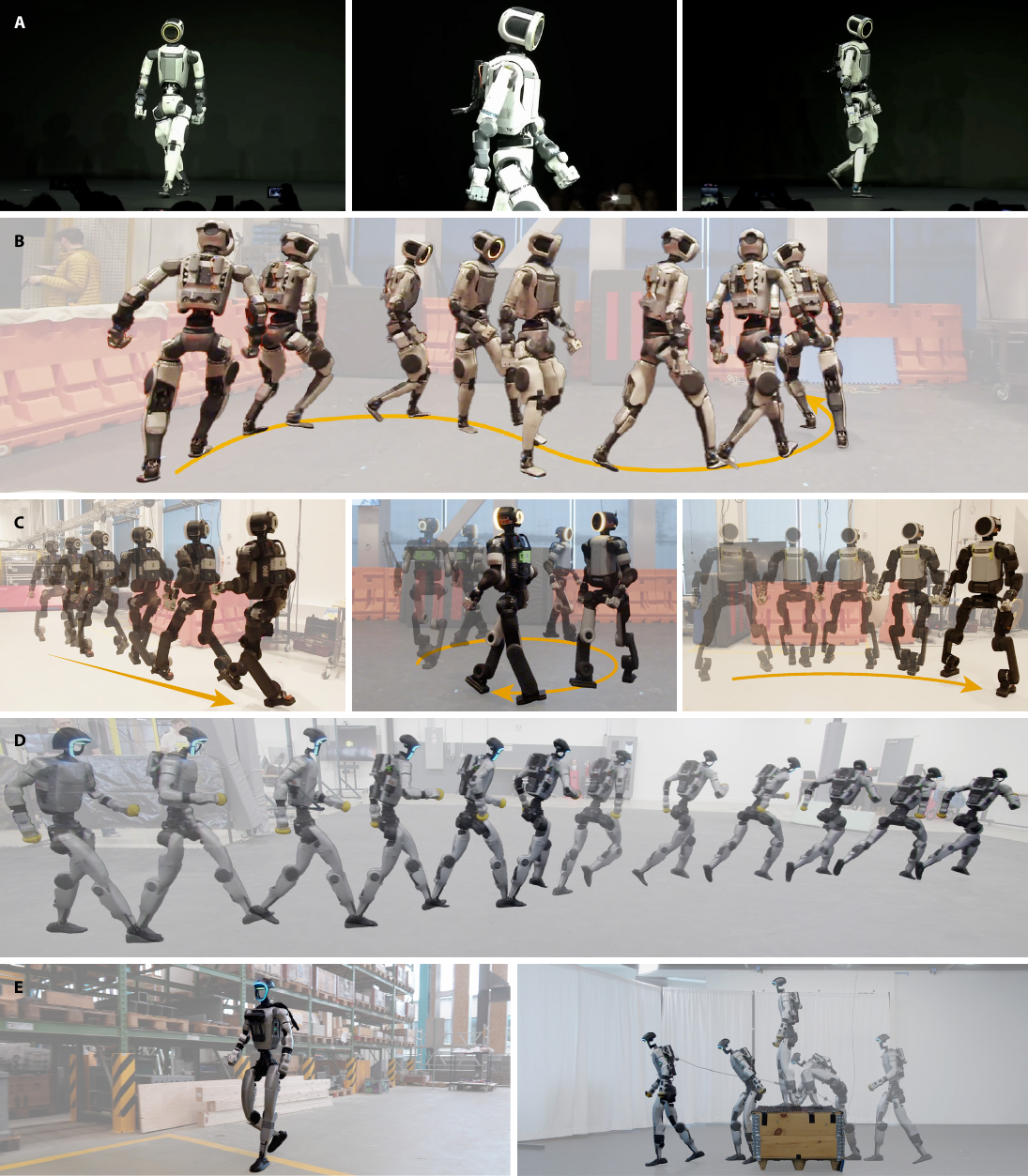}
    \caption{\teasercaption}
    \label{fig:teaser}
\end{figure*}
\fi

A primary technical challenge with relying solely on this teacher--student distillation is that training the policy strictly within the support of the reference dataset restricts its command coverage. At deployment time, steering commands that are absent or poorly represented in the original dataset can lead to out-of-distribution failures.
To address this limitation, we adopt an approach similar to the pre-training and post-training paradigm that has recently emerged in foundation models \cite{Long2022finetuningllm, guo2025deepseek}. 
Building on this core concept, applying RL fine-tuning as a post-training step has increasingly gained attention across the robotics community as a way to expose policies to broader distributions over tasks, commands, or environments.
For instance, \emph{Parkour in the Wild} \cite{Rudin2025ParkourWild} uses this strategy to distill multi-skill terrain-aware locomotion into a single policy and then fine-tunes it on broader terrain distributions.
\emph{InterPrior} \cite{xu2026interprior} applies a similar principle to whole-body loco-manipulation, distilling diverse skills into a policy conditioned on sparse, multimodal goals and fine-tuning it for generalization across large human-object interaction spaces.
\emph{PHP} \cite{wu2026perceptive} further combines privileged environmental information with human motion data to learn dynamic humanoid parkour behaviors, which are distilled into a single perceptive, velocity-commanded policy using \emph{DAgger} (Dataset Aggregation) \cite{Ross2011DAgger} and RL.
Extending this post-distillation fine-tuning paradigm, \methodtitle~resolves the style--command coverage trade-off through a structured multi-task RL formulation. Rather than fine-tuning the distilled policy solely for goal-conditioned command tracking, \methodtitle~couples short-horizon reference-guided imitation, which anchors the policy to the human motion manifold, with long-horizon goal-conditioned command tracking, which improves responsiveness and robustness beyond the dataset's support. 
A goal-conditioned state-initialization scheme further stabilizes this process by seeding rollouts from dynamically plausible, command-relevant states.

The key idea behind \methodtitle~is to separate style acquisition from command-space expansion.
The proposed framework implements this idea through a two-stage training process based on a newly curated human motion dataset encompassing diverse headings, velocities, and gait modes, including small stepping, canonical walking, sprinting, sidestepping, and turning.
In the first stage, a natural locomotion prior is distilled from a full-body reference-conditioned teacher policy into a student policy conditioned only on planar torso-velocity commands (\setwo~velocity commands) and the robot's proprioceptive state.
In the second stage, this prior is fine-tuned using the multi-task formulation described above to expand command coverage while preserving fidelity to the motion data. 
Together, these components yield a compact command-conditioned policy that produces naturalistic, diverse, and steerable locomotion without requiring full-body reference motions at runtime.

%
\ifdefined\isArXiv
\else
\begin{figure*}[t!]
    \centering
    \includegraphics[width=0.77\textwidth]{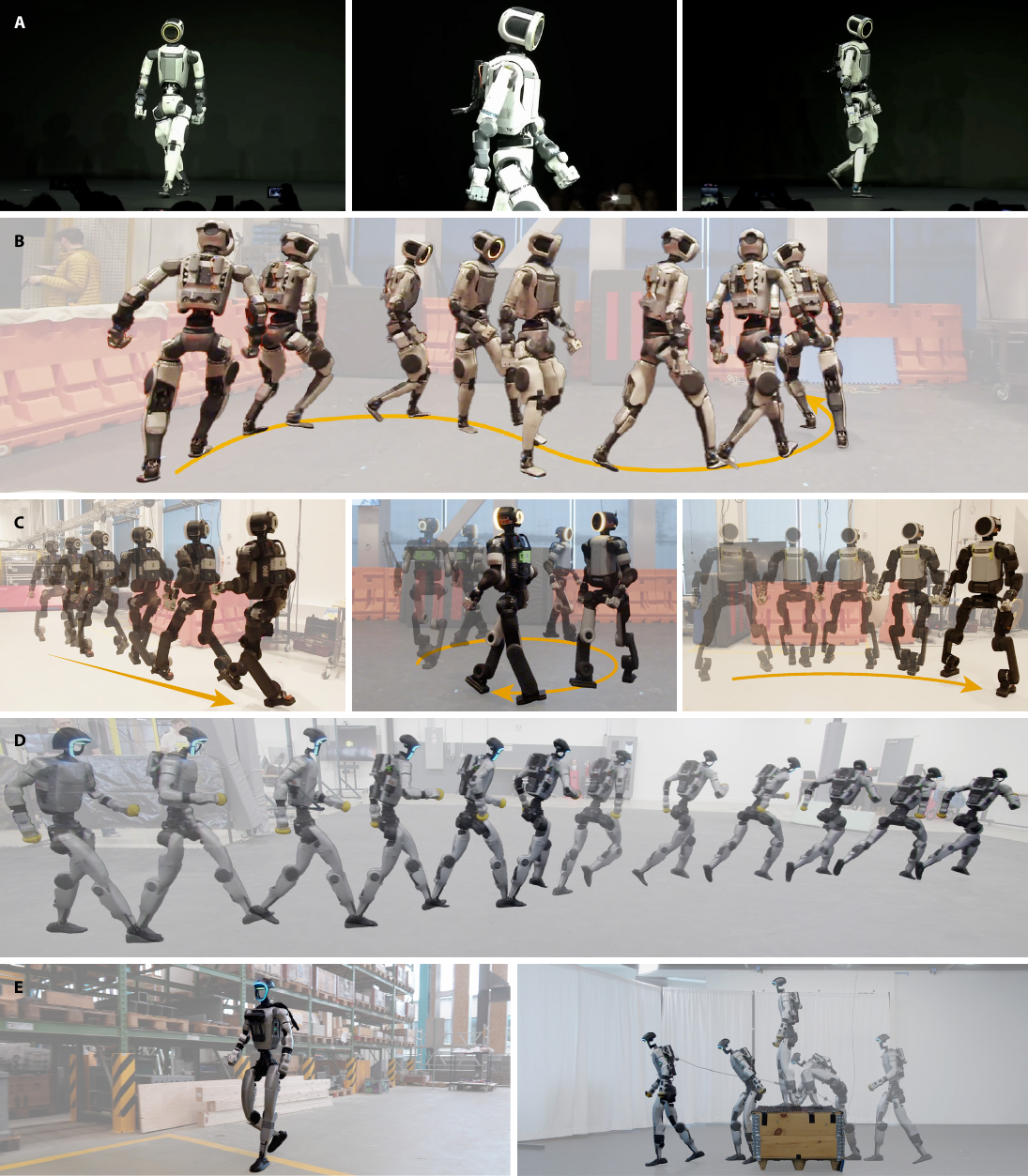}
    \caption{\teasercaption}
    \label{fig:teaser}
\end{figure*}
\fi

We evaluate how \methodtitle~balances the trade-off between human-like motion style, command responsiveness, robustness during real-world deployment through experiments on Boston Dynamics Atlas R1, Atlas D1, and Unitree G1 humanoids across both laboratory and outdoor locomotion scenarios as illustrated in \Cref{fig:teaser}.
In our quantitative evaluation, policy performance is analyzed using metrics aligned with the design criteria: fidelity to the human motion data, command-tracking accuracy over an expansive \setwo~velocity-command space, response latency, disturbance rejection, and sim-to-real transfer consistency.
Furthermore, we isolate the efficacy of our key architectural choices through simulation-based ablation studies.
These evaluations demonstrate that the learned policy preserves natural, full-body movement patterns by maintaining consistency with the original motion data while remaining highly responsive to user steering commands within a single, compact multi-layer perceptron (MLP) architecture.
Crucially, the resulting policy exhibits robust generalization to a wide variety of steering inputs and command transitions that are absent or only sparsely represented in the reference motion dataset. Our comprehensive evaluation confirms that \methodtitle{} bridges the gap between stylized motion and agile control, achieving robustness and command responsiveness comparable to a \tabularasa{} RL baseline optimized solely for command-tracking performance, while dramatically improving full-body behavioral naturalness.

Alongside our core results, we release the curated human locomotion dataset used in this study, which includes approximately three hours of motion capture data spanning diverse speeds, directions, and gait modalities. This dataset is provided to ensure research reproducibility and foster broader progress in humanoid locomotion.


\section*{RESULTS}

%
\ifdefined\isArXiv
\begin{figure*}[t!]
    \centering
    \includegraphics[width=\linewidth]{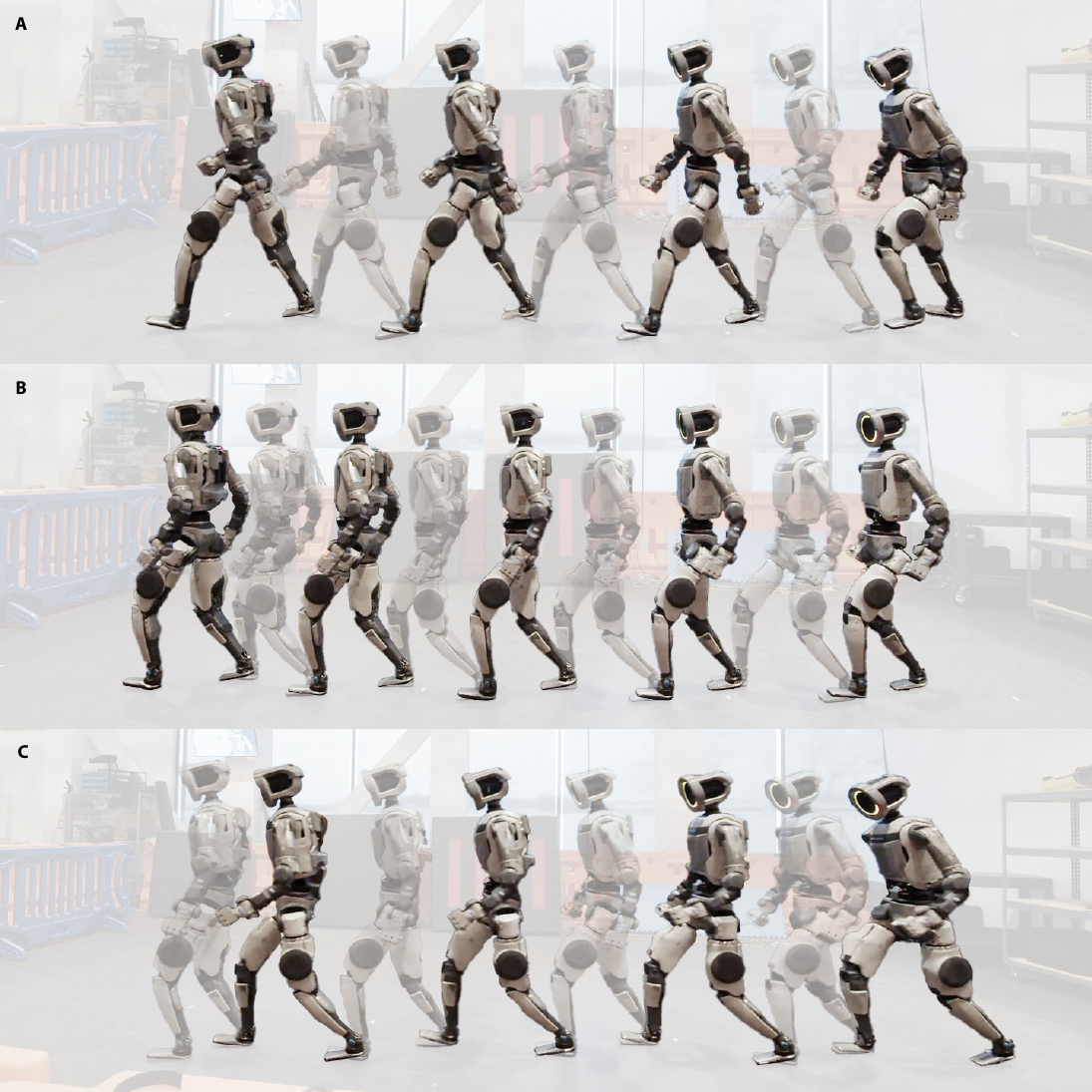}
    \caption{\stroboscopiccaption}
    \vspace{-1em}
    \label{fig:contrast}
\end{figure*}
\fi

\noindent This section presents the qualitative and quantitative evaluations of \methodtitle~and summarizes the empirical findings. The evaluation is conducted across three humanoid platforms: the Boston Dynamics Atlas R1, Atlas D1, and Unitree G1. 
We first qualitatively assess the stylistic quality of the final deployment policy trained with \methodtitle, focusing on visible features such as arm swing, heel strike, heel-toe rolling, knee extension, and upper-lower body coordination. 
To evaluate the preservation of human locomotion style quantitatively, we perform a data fidelity analysis that measures the reconstruction accuracy of full-body behaviors relative to the target reference dataset, using this metric as a proxy for stylistic consistency.
Subsequently, we assess the policy's command responsiveness, command coverage, and perturbation robustness through systematic, grid-search-style evaluation sweeps.
We then analyze sim-to-real transfer consistency by comparing command-tracking errors in simulation against physical hardware performance across all three humanoid platforms. 
Beyond these core benchmark evaluations, we showcase the versatility of the learned policy across diverse real-world deployment scenarios, including user joystick steering and integration within a hierarchical multi-policy framework, demonstrated both in the laboratory and outdoor environments.
Finally, we present simulation-based ablation studies to analyze the impact of key design choices within our framework.

All policies under evaluation were trained using the \emph{Isaac Lab} \cite{mittal2025isaac} physics simulator. While it served as our primary sim-based quantitative evaluation environment as well, we cross-validated responsiveness, command coverage and robustness evaluation using a simulation pipeline built on \emph{MuJoCo} \cite{todorov2012mujoco}, which provides a high-fidelity, hardware-validated proxy that captures physical deployment constraints.

%
\ifdefined\isArXiv
\begin{figure*}[t!]
\centering
\includegraphics[width=\textwidth]{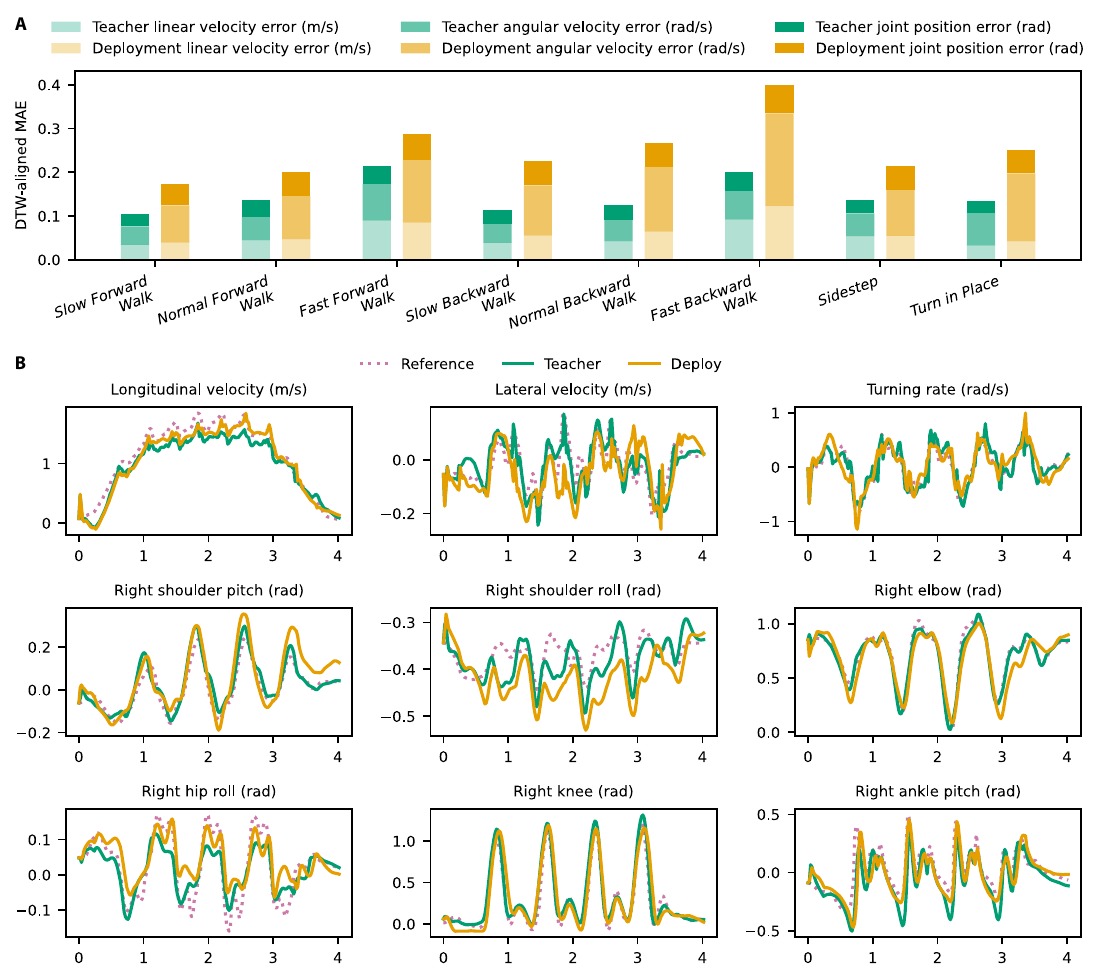}    
\caption{\datafidelitycaptionnew}
\label{fig:g1_data_fidelity}
\end{figure*}
\fi

%
\ifdefined\isArXiv
\else
\begin{figure*}[t!]
    \centering
    \includegraphics[width=\linewidth]{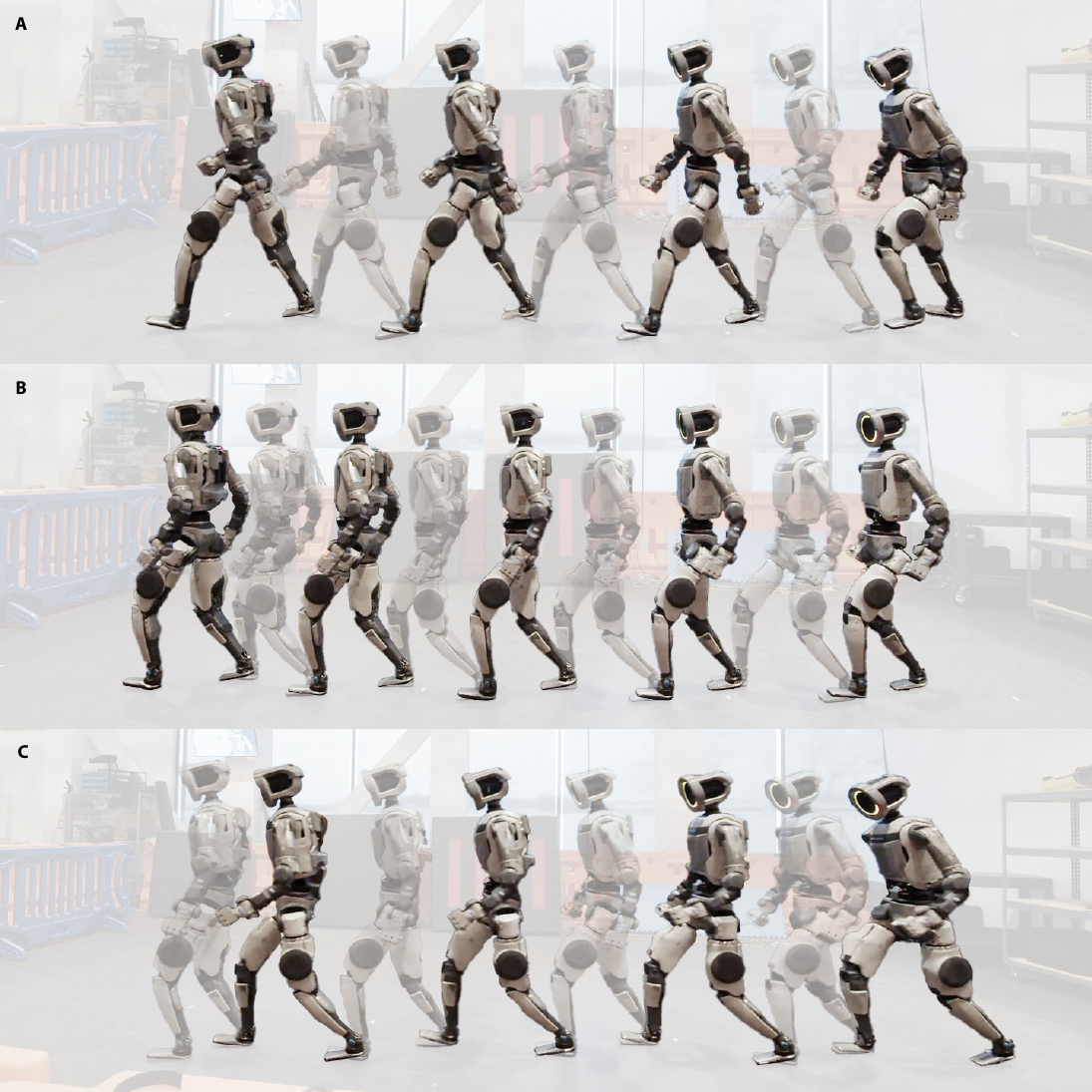}
    \caption{\stroboscopiccaption}
    \label{fig:contrast}
\end{figure*}
\fi

\subsection*{Qualitative Gait Characteristics}

We first highlight the stylistic characteristics exhibited by the policy trained with \methodtitle~during forward walking.
On the Atlas R1 robot, we contrast these characteristics against two baselines: the robot's default controller provided by Boston Dynamics, which uses footstep planning tracked by a Model Predictive Controller (MPC), and a \tabularasa~RL policy trained without human motion priors.
\Cref{fig:contrast} illustrates stroboscopic snapshots of the Atlas R1 robot executing forward walk in response to a constant forward velocity steering command under three different controllers: the \methodtitle~policy, the \tabularasa~RL policy, and the MPC.

Compared to the MPC, the \methodtitle~policy exhibits visible biomimetic gait features associated with the human-motion prior, including synchronized arm swing, knee extension near stance, hip motion used to regulate torso orientation, a heel-strike-like contact transition, and heel-to-toe rolling of the stance foot.
In contrast, the MPC produces stiff arm motion with the limbs remaining nearly fixed relative to the torso. While the MPC’s gait appears more strictly periodic, the \methodtitle~policy captures the natural fluidity and whole-body coordination characteristic of human movement.

Relative to the \tabularasa~policy, the \methodtitle~policy shows stronger spatiotemporal coordination between the upper and lower body, the clear presence of knee-locking, and a prominent heel-to-toe rolling during ground contact, leading to smoother momentum transfer and quieter steps. Similar gait features are observed for policies trained with \methodtitle~on the Atlas D1 and Unitree G1 robots. Notably, on Atlas D1, despite its non-human foot morphology, the policy retains a heel-to-toe rolling consistent with the human-motion prior. 

Beyond the constant-velocity comparison presented in \Cref{fig:contrast}, the \methodtitle~policy also seamlessly adapts its gait style across command speeds with smooth transitions as illustrated in the video. 
At lower speeds, the policy adopts a flatter, shuffling gait, which transitions toward a more pronounced heel-strike and toe-off pattern as the commanded speed increases. 
As evidenced by the \tabularasa~RL and MPC baselines, such human-like characteristics do not necessarily emerge even in highly capable locomotion controllers designed for utility and robustness. 
Although this analysis is qualitative, it highlights distinct gait features that differentiate the \methodtitle~policy from traditional controllers and motivates the rigorous quantitative fidelity analysis presented in the next section.

\subsection*{Fidelity to the Motion Dataset}

Beyond visual inspection, we quantitatively assess the stylistic characteristics of the policy trained by \methodtitle. 
Because stylistic quality is inherently subjective and difficult to measure directly, we use motion fidelity to the retargeted human motion data as a quantitative proxy. 
Since the final \methodtitle~policy is not conditioned on full-body reference data, it must reconstruct full-body behavior from proprioceptive observations and \setwo~velocity commands alone. 
Thus, its deviation from the reference trajectories measures how well the policy preserves the motion patterns encoded in the dataset. 
We also compare the deployment policy against a policy conditioned on the full-body reference, which we refer to as the teacher policy in the next section. The teacher policy is trained using a motion-mimic framework~\cite{sleiman2026zest} to track the reference motions while producing dynamically feasible behavior, thereby providing a meaningful upper bound on deployment performance. This baseline is critical, as the raw kinematic references often contain non-physical artifacts introduced during the motion retargeting process. 
Together, these comparisons quantify fidelity to the original motion data and a physically grounded instantiation of the learned motion prior.

The evaluation is designed as follows: In physical simulation, we evaluate the final \methodtitle~deployment policy trained for the Unitree G1 robot across eight canonical motion profiles extracted from the reference dataset. These canonical motions comprise three forward walking speeds (slow, normal, and fast), three backward walking speeds (slow, normal, and fast), a turn-in-place maneuver, and a sidestep motion\footnote[2]{For clarity, capitalized and italicized terms (e.g., \emph{Fast Forward Walk}) denote the specific named motion segments extracted from the reference dataset used throughout this evaluation.}. For each trajectory, we feed the annotated \setwo~velocity command from the reference data into the deployment policy as a steering command and collect the resulting robot trajectories. 

Because minor timing discrepancies between the reference trajectory and the simulation rollout can artificially inflate the tracking error metrics, we employ Dynamic Time Warping (DTW) \cite{giorgino2009dtw} as a time-series alignment method prior to the comparison. DTW mitigates the impact of these slight temporal offsets by non-linearly warping the time axes, allowing us to evaluate the structural similarity and underlying trends of the two trajectories independent of phase shifts. Specifically, temporal alignment is achieved by pairing sample points between the reference and rollout trajectories to minimize a distance metric computed over a concatenated vector of the torso linear velocity, torso angular velocity, and joint positions, subject to a strict temporal monotonicity constraint on the alignment indices.  
\Cref{fig:g1_data_fidelity} summarizes the evaluation results on the simulated Unitree G1 robot.

The bar chart in \customref{fig:g1_data_fidelity}{A} reports the DTW-aligned Mean Absolute Error (MAE) of the torso linear velocity, torso angular velocity, and joint positions relative to the reference motion data for both the teacher and deployment policies.
The mean absolute temporal displacement induced by DTW across evaluation scenarios is a modest \SI{68.7}{\milli\second}, indicating that the temporal alignment accounts for minor phase variations without significantly distorting the trajectory profiles.

%
\ifdefined\isArXiv
\else
\begin{figure*}[t!]
\centering
\includegraphics[width=0.98\textwidth]{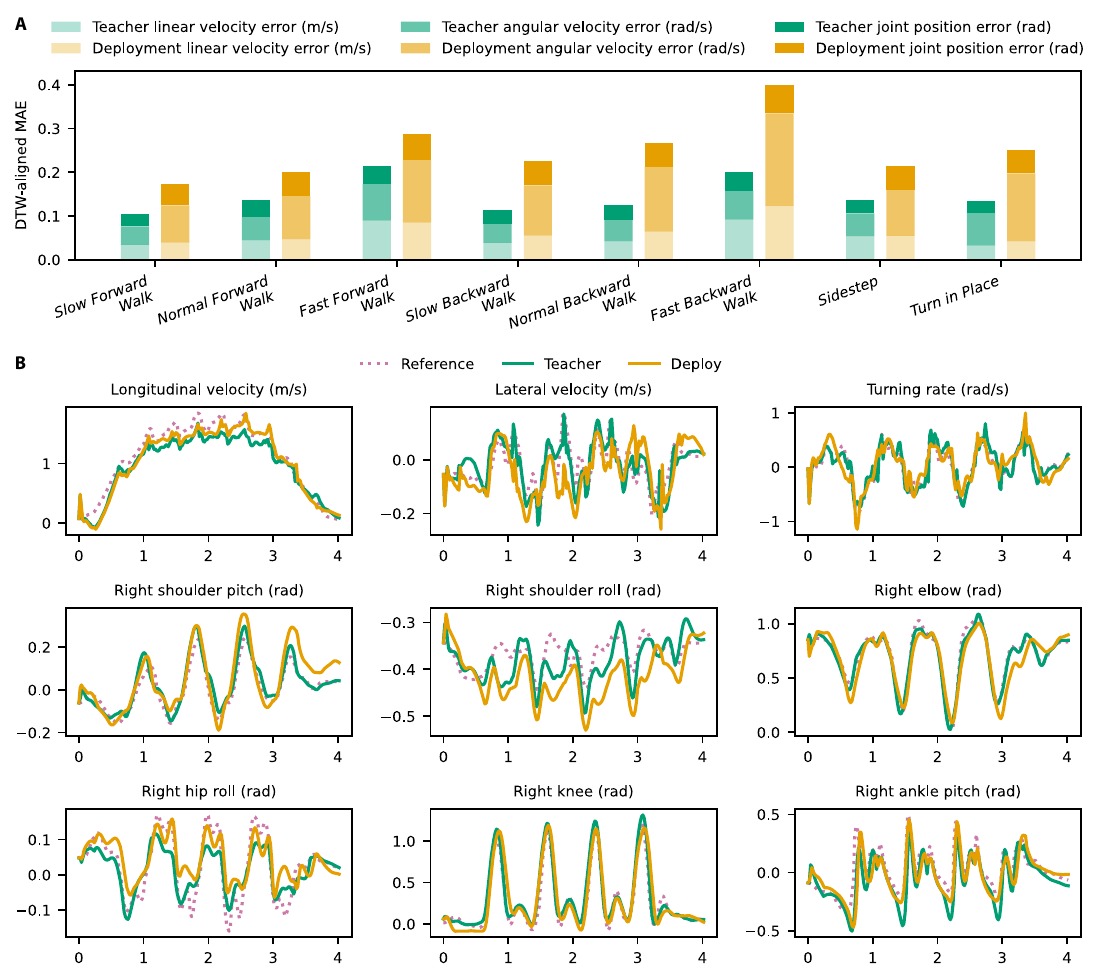}    
\caption{\datafidelitycaptionnew}
\label{fig:g1_data_fidelity}
\end{figure*}
\fi

Across the canonical motion trajectories, the deployment policy achieves a mean DTW-aligned MAE of \SI[per-mode=repeated-symbol]{0.058}{\metre\per\second} for torso linear velocity, \SI[per-mode=repeated-symbol]{0.127}{\radian\per\second} for torso angular velocity, and \SI{0.055}{\radian} for joint positions.
In comparison, the teacher policy achieves \SI[per-mode=repeated-symbol]{0.048}{\metre\per\second}, \SI[per-mode=repeated-symbol]{0.055}{\radian\per\second}, and \SI{0.033}{\radian}, respectively. 
Consequently, the deployment policy exhibits 20.83\%, 130.31\%, and 67.58\% greater tracking error than the teacher policy across these respective components. This performance gap is expected, as the teacher policy is optimized with direct access to full-body reference states, whereas the deployment policy must reconstruct the reference behaviors from partial observations. Furthermore, the deployment policy undergoes a command-conditioned fine-tuning stage to follow arbitrary steering inputs, thereby introducing some deviation from the references.

Nevertheless, relative to the standard deviations of the reference trajectories for linear velocity $\sigma_{\bm{v}}=0.390\,$\SI[per-mode=repeated-symbol]{}{\metre\per\second}, angular velocity $\sigma_{\bm{\omega}}=0.491\,$\SI[per-mode=repeated-symbol]{}{\radian\per\second}, and joint positions $\sigma_j=0.300\,$\SI{}{\radian}, the deployment policy's errors correspond to only $0.149\sigma_{\bm{v}}$, $0.257\sigma_{\bm{\omega}}$, and $0.184\sigma_j$, respectively. 
These normalized errors indicate that the deployment policy remains close to the retargeted motion distribution across the evaluated canonical motions, even in the absence of full-body reference inputs.

To examine the temporal structure of these errors, we compare the fast-forward walk reference with the corresponding simulation rollouts prior to DTW alignment, focusing on linear and angular velocities, as well as representative joint position trajectories, as illustrated in \customref{fig:g1_data_fidelity}{B}.
The illustrated joints were chosen to highlight the key visible components of human-like walking: shoulder and elbow motion for arm swing, hip roll for lateral balance and torso regulation, knee motion for stance-leg extension, and ankle pitch for foot placement and heel-to-toe rolling.
Across these dimensions, the deployment policy closely follows the phase and amplitude trends of the teacher and reference trajectories, despite receiving only proprioception and \setwo~velocity commands.
This close agreement indicates that the policy preserves not only the commanded torso motion but also highly coordinated upper- and lower-body patterns from the motion dataset.
For completeness, a comprehensive set of trajectory profiles and metrics for the remaining canonical motions is provided in the Supplementary Materials \highlight{(Figure~S1 and Table~S1)}.

%
\ifdefined\isArXiv
\begin{figure*}[t!]
    \centering
    \includegraphics[width=\textwidth]{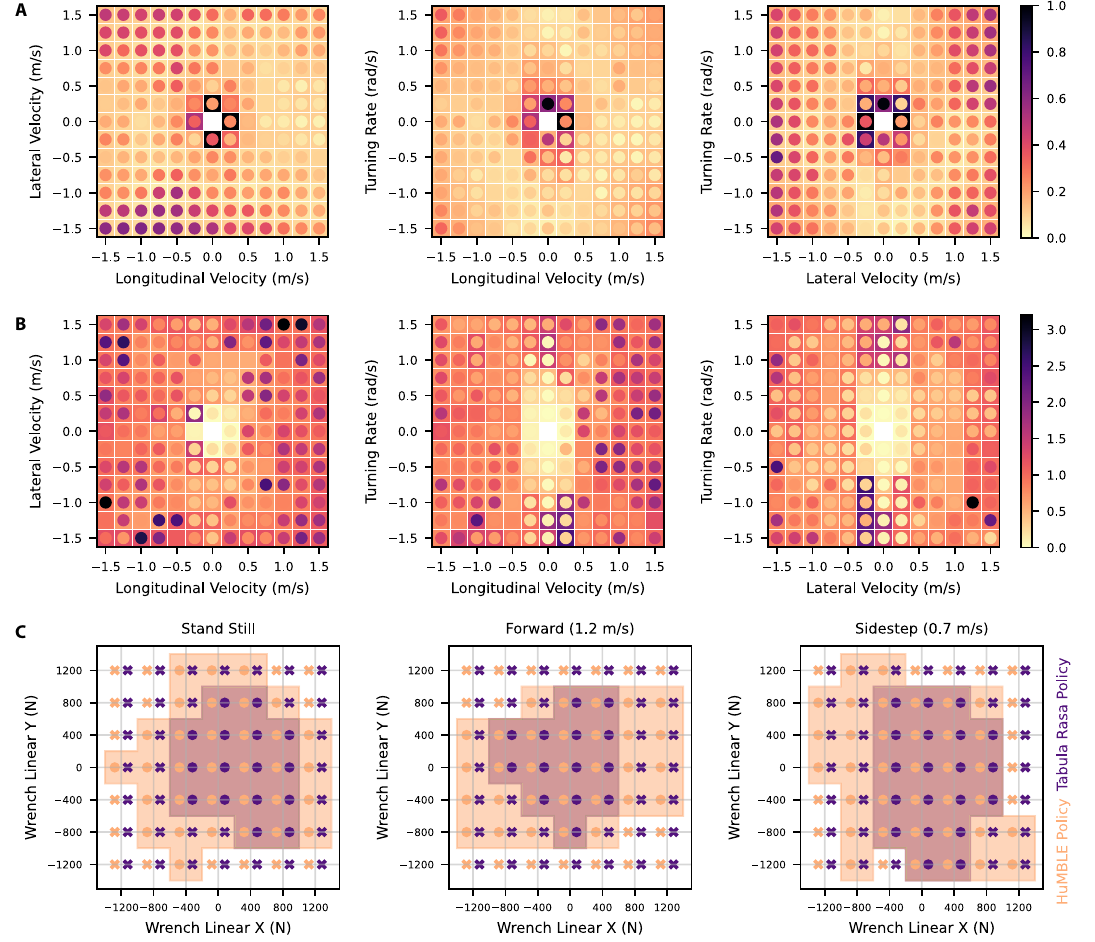}
    \vspace{-1em}
    \caption{\quantitativecaption}
    \label{fig:r1_quantitative_analysis}
\end{figure*}
\fi

\subsection*{Command Coverage, Responsiveness, and Robustness}
\noindent 
In addition to preserving stylistic consistency with the human data, the \methodtitle~policy must accurately track a broad range of steering commands with high responsiveness, while simultaneously maintaining robustness to external perturbations. To quantitatively evaluate these properties, we conducted a series of systematic, grid-search-style evaluation sweeps in simulation. To provide a benchmark, we compare the \methodtitle~policy against a \tabularasa~RL policy trained exclusively for velocity command tracking and robustness, without the stylistic guidance or constraints of the human motion data. While this baseline may not represent a globally optimal control design, as performance can vary with alternative RL formulations, it serves as a representative reference for a hardware-deployable locomotion policy with robust and reliable velocity command tracking under the identical observation, action, and domain randomization specifications as \methodtitle. A detailed implementation setup for our \tabularasa~RL policy is provided in the Supplementary Materials (\highlight{Section~S6}).

\Cref{fig:r1_quantitative_analysis} summarizes this evaluation for Atlas R1, with corresponding results for the other robots detailed in the Supplementary Materials \highlight{(Figure~S2 for Atlas D1 and Figure~S3 for Unitree G1)}.
In the first set of experiments, we perform a step-response analysis, inspired by classical control evaluation protocols, to characterize the robot's dynamic response across a broad command space. Evaluating a wide velocity envelope is motivated by the challenge of training a single policy to perform effectively over a diverse spectrum of speeds while maintaining high-resolution control. We evaluate the step response for each command by applying a step input to the policy. Steady state is reached when the robot's torso velocity has settled after the step input; in practice, we take this as \SI{3.0}{\second} after the step input. Rise time is measured as the duration from step initiation until 90\% of the steady-state torso velocity is first reached.

%
\ifdefined\isArXiv
\else
\begin{figure*}[t!]
    \centering
    \includegraphics[width=0.95\textwidth]{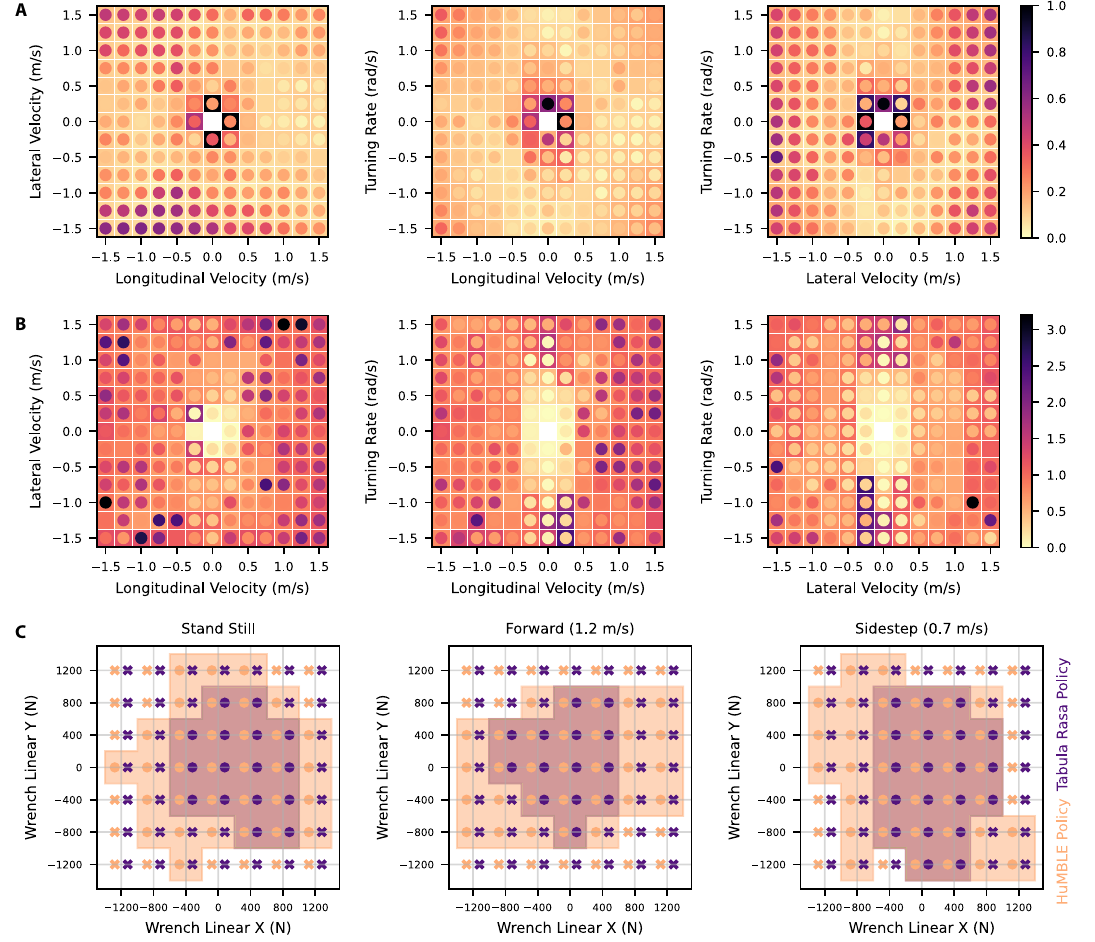}
    \caption{\quantitativecaption}
    \label{fig:r1_quantitative_analysis}
\end{figure*}
\fi

In \customref{fig:r1_quantitative_analysis}{A}, we report the relative command-tracking error, defined as the steady-state tracking error normalized by the command magnitude. 
The \methodtitle~policy demonstrates errors comparable to those of the \tabularasa~RL baseline across most of the evaluated \setwo~command space. The most significant errors are observed in command regions that are underrepresented or missing in the motion dataset, such as fast diagonal walking.
For near-zero velocity commands, the \methodtitle~policy performs favorably compared to the \tabularasa~RL baseline. This outcome likely results from the inclusion of slow-shuffling references in the motion dataset, which provide a motion prior for small but non-zero locomotion commands that are challenging to discover through task-reward optimization alone.

Regarding responsiveness, as illustrated in \customref{fig:r1_quantitative_analysis}{B}, both policies exhibit longer rise times at higher command magnitudes. The \tabularasa~RL baseline typically achieves large commands more rapidly, while the \methodtitle~policy yields smoother and more gradual responses in the high-speed regime. This pattern aligns with the acceleration profiles inherent to the human locomotion data and highlights the trade-off between naturalistic transitions and aggressive command tracking.

In the final set of experiments, we evaluate robustness by perturbing the robot's torso with planar force pushes of varying directions and magnitudes, and 
report whether the robot can withstand each disturbance without falling.
A fall is defined as the robot's torso tilting excessively or the torso height falling below a threshold (\SI{30}{\degree} along the pitch or roll axes, and 40\% of the nominal standing torso height respectively, for Atlas R1). This evaluation is conducted across three scenarios: standing still, walking forward, and sidestepping. Forces are applied as continuous \SI{0.2}{\second} pulses once the policy reaches steady-state velocity, while the target command remains constant throughout each test case.

As illustrated in \customref{fig:r1_quantitative_analysis}{C}, the \methodtitle~policy withstands push magnitudes of up to \SI{1400}{\newton} along major principal directions on Atlas R1. 
Notably, the push robustness of the \methodtitle~policy exhibits a directional dependence closely aligned with the gait kinematics. For example, during forward walking, our policy withstands stronger pushes along the longitudinal direction compared to the lateral direction, primarily due to the sagittal leg swing and the effective support polygon established during forward locomotion. Conversely, during sidestepping, a wider stance increases lateral stability at the expense of longitudinal stability. 

Although the exact evaluation metrics vary across the three robotic platforms, the overall trends remain consistent: \methodtitle{} policies deliver command-tracking accuracy, responsiveness, and disturbance robustness comparable to the \tabularasa{} RL baselines. The minor trade-off in high-speed responsiveness—manifested as slightly longer rise times during command transitions—is a direct result of the naturalistic acceleration profiles learned from human reference data. Overall, we conclude that \methodtitle{} policies are well-suited for real-world deployment as robust locomotion controllers, offering a compelling balance between biomimetic agility, trajectory tracking precision, and push recovery.

\ifdefined\isArXiv
%
%
\begin{table}[t]
\centering
\caption{\simtorealcaption}
\label{tab:sim2real_analysis}
\vspace{0.5em}

\begingroup
\small
\setlength{\tabcolsep}{4pt}
\renewcommand{\arraystretch}{1.0}

\sisetup{
  round-mode          = places,
  round-precision     = 2,
  table-number-alignment = center,
}

\definecolor{realgray}{gray}{0.92}
\begin{tabular}{lcccc}
\toprule
\multicolumn{1}{c}{} &
\multicolumn{4}{c}{\textbf{Boston Dynamics Atlas R1}} \\
\cmidrule(lr){2-5}

\multicolumn{1}{c}{} &
\multicolumn{2}{c}{\textbf{Steady-State Error}} & \multicolumn{2}{c}{\textbf{Mean Absolute Error}} \\
\cmidrule(lr){2-3}\cmidrule(lr){4-5}

\textbf{Command} &
\multicolumn{1}{c}{\textbf{Sim}} & \multicolumn{1}{c}{\textbf{Real}} &
\multicolumn{1}{c}{\textbf{Sim}} & \multicolumn{1}{c}{\textbf{Real}} \\
\midrule
\midrule
\multicolumn{5}{@{}l}{\textbf{Forward Walk}} \\
\addlinespace[0.3em]
(0.4, 0.0, 0.0)      & (0.08, 0.02) & (0.12, 0.03) & (0.10, 0.10) & (0.11, 0.10) \\
(1.2, 0.0, 0.0)      & (0.03, 0.02) & (0.06, 0.07) & (0.10, 0.15) & (0.10, 0.18) \\
(2.0, 0.0, 0.0)      & (0.13, 0.11) & (0.20, 0.06) & (0.15, 0.13) & (0.18, 0.16) \\
\addlinespace[0.3em]
\midrule
\multicolumn{5}{@{}l}{\textbf{Backward Walk}} \\
\addlinespace[0.3em]
(-0.3, 0.0, 0.0)      & (0.04, 0.02) & (0.03, 0.02) & (0.09, 0.10) & (0.09, 0.10) \\
(-0.9, 0.0, 0.0)      & (0.10, 0.02) & (0.20, 0.10) & (0.11, 0.12) & (0.14, 0.16) \\
(-1.5, 0.0, 0.0)      & (0.23, 0.01) & (0.21, 0.07) & (0.17, 0.14) & (0.18, 0.16) \\
\addlinespace[0.3em]
\midrule
\multicolumn{5}{@{}l}{\textbf{Sidestepping}} \\
\addlinespace[0.3em]
(0.0, -0.7, 0.0)     & (0.09, 0.13) & (0.07, 0.04) & (0.13, 0.09) & (0.14, 0.11) \\
(0.0, 0.7, 0.0)      & (0.08, 0.02) & (0.10, 0.05) & (0.13, 0.10) & (0.14, 0.07) \\

(0.0, -1.0, 0.0)     & (0.27, 0.04) & (0.25, 0.09) & (0.19, 0.08) & (0.21, 0.11) \\
(0.0, 1.0, 0.0)      & (0.14, 0.09) & (0.24, 0.04) & (0.19, 0.08) & (0.21, 0.11) \\
\addlinespace[0.3em]
\midrule
\multicolumn{5}{@{}l}{\textbf{Turn in Place}} \\
\addlinespace[0.3em]
(0.0, 0.0, -1.0)    & (0.05, 0.04) & (0.07, 0.07) & (0.08, 0.07) & (0.08, 0.10) \\
(0.0, 0.0, 1.0)     & (0.04, 0.08) & (0.05, 0.03) & (0.09, 0.08) & (0.09, 0.07) \\

(0.0, 0.0, -2.0)    & (0.04, 0.15) & (0.09, 0.22) & (0.10, 0.15) & (0.10, 0.17) \\
(0.0, 0.0, 2.0)     & (0.02, 0.27) & (0.06, 0.34) & (0.10, 0.22) & (0.10, 0.25) \\

(0.0, 0.0, -3.0)    & (0.07, 0.53) & (0.05, 0.60) & (0.11, 0.34) & (0.13, 0.39) \\
(0.0, 0.0, 3.0)     & (0.05, 0.47) & (0.06, 0.42) & (0.14, 0.37) & (0.13, 0.39) \\
\addlinespace[0.3em]
\midrule
\multicolumn{5}{@{}l}{\textbf{Forward Circle}} \\
\addlinespace[0.3em]
(1.0, 0.0, -1.0)    & (0.07, 0.03) & (0.09, 0.19) & (0.10, 0.17) & (0.14, 0.21) \\
(1.0, 0.0, 1.0)     & (0.06, 0.12) & (0.12, 0.07) & (0.11, 0.14) & (0.11, 0.14) \\

(1.0, 0.0, -1.5)    & (0.08, 0.25) & (0.11, 0.33) & (0.10, 0.18) & (0.13, 0.24) \\
(1.0, 0.0, 1.5)     & (0.09, 0.26) & (0.05, 0.28) & (0.12, 0.21) & (0.13, 0.24) \\
\bottomrule
\end{tabular}
\endgroup
\end{table}
\fi

\subsection*{Sim-to-Real Analysis}

\ifdefined\isArXiv
\else
%
%
\begin{table}[t!]
\centering
\caption{\simtorealcaption}
\label{tab:sim2real_analysis}

\begingroup
\scriptsize
\setlength{\tabcolsep}{4pt}
\renewcommand{\arraystretch}{1.0}

\sisetup{
  round-mode          = places,
  round-precision     = 2,
  table-number-alignment = center,
}

\definecolor{realgray}{gray}{0.92}
\begin{tabular}{lcccc}
\toprule
\multicolumn{1}{c}{} &
\multicolumn{4}{c}{\textbf{Boston Dynamics Atlas R1}} \\
\cmidrule(lr){2-5}

\multicolumn{1}{c}{} &
\multicolumn{2}{c}{\textbf{Steady-State Error}} & \multicolumn{2}{c}{\textbf{Mean Absolute Error}} \\
\cmidrule(lr){2-3}\cmidrule(lr){4-5}

\textbf{Command} &
\multicolumn{1}{c}{\textbf{Sim}} & \multicolumn{1}{c}{\textbf{Real}} &
\multicolumn{1}{c}{\textbf{Sim}} & \multicolumn{1}{c}{\textbf{Real}} \\
\midrule
\midrule
\multicolumn{5}{@{}l}{\textbf{Forward Walk}} \\
\addlinespace[0.3em]
(0.4, 0.0, 0.0)      & (0.08, 0.02) & (0.12, 0.03) & (0.10, 0.10) & (0.11, 0.10) \\
(1.2, 0.0, 0.0)      & (0.03, 0.02) & (0.06, 0.07) & (0.10, 0.15) & (0.10, 0.18) \\
(2.0, 0.0, 0.0)      & (0.13, 0.11) & (0.20, 0.06) & (0.15, 0.13) & (0.18, 0.16) \\
\addlinespace[0.3em]
\midrule
\multicolumn{5}{@{}l}{\textbf{Backward Walk}} \\
\addlinespace[0.3em]
(-0.3, 0.0, 0.0)      & (0.04, 0.02) & (0.03, 0.02) & (0.09, 0.10) & (0.09, 0.10) \\
(-0.9, 0.0, 0.0)      & (0.10, 0.02) & (0.20, 0.10) & (0.11, 0.12) & (0.14, 0.16) \\
(-1.5, 0.0, 0.0)      & (0.23, 0.01) & (0.21, 0.07) & (0.17, 0.14) & (0.18, 0.16) \\
\addlinespace[0.3em]
\midrule
\multicolumn{5}{@{}l}{\textbf{Sidestepping}} \\
\addlinespace[0.3em]
(0.0, -0.7, 0.0)     & (0.09, 0.13) & (0.07, 0.04) & (0.13, 0.09) & (0.14, 0.11) \\
(0.0, 0.7, 0.0)      & (0.08, 0.02) & (0.10, 0.05) & (0.13, 0.10) & (0.14, 0.07) \\

(0.0, -1.0, 0.0)     & (0.27, 0.04) & (0.25, 0.09) & (0.19, 0.08) & (0.21, 0.11) \\
(0.0, 1.0, 0.0)      & (0.14, 0.09) & (0.24, 0.04) & (0.19, 0.08) & (0.21, 0.11) \\
\addlinespace[0.3em]
\midrule
\multicolumn{5}{@{}l}{\textbf{Turn in Place}} \\
\addlinespace[0.3em]
(0.0, 0.0, -1.0)    & (0.05, 0.04) & (0.07, 0.07) & (0.08, 0.07) & (0.08, 0.10) \\
(0.0, 0.0, 1.0)     & (0.04, 0.08) & (0.05, 0.03) & (0.09, 0.08) & (0.09, 0.07) \\

(0.0, 0.0, -2.0)    & (0.04, 0.15) & (0.09, 0.22) & (0.10, 0.15) & (0.10, 0.17) \\
(0.0, 0.0, 2.0)     & (0.02, 0.27) & (0.06, 0.34) & (0.10, 0.22) & (0.10, 0.25) \\

(0.0, 0.0, -3.0)    & (0.07, 0.53) & (0.05, 0.60) & (0.11, 0.34) & (0.13, 0.39) \\
(0.0, 0.0, 3.0)     & (0.05, 0.47) & (0.06, 0.42) & (0.14, 0.37) & (0.13, 0.39) \\
\addlinespace[0.3em]
\midrule
\multicolumn{5}{@{}l}{\textbf{Forward Circle}} \\
\addlinespace[0.3em]
(1.0, 0.0, -1.0)    & (0.07, 0.03) & (0.09, 0.19) & (0.10, 0.17) & (0.14, 0.21) \\
(1.0, 0.0, 1.0)     & (0.06, 0.12) & (0.12, 0.07) & (0.11, 0.14) & (0.11, 0.14) \\

(1.0, 0.0, -1.5)    & (0.08, 0.25) & (0.11, 0.33) & (0.10, 0.18) & (0.13, 0.24) \\
(1.0, 0.0, 1.5)     & (0.09, 0.26) & (0.05, 0.28) & (0.12, 0.21) & (0.13, 0.24) \\
\bottomrule
\end{tabular}
\endgroup
\end{table}
\fi

\noindent To demonstrate the policy's sim-to-real transfer, its performance in simulation and on hardware is compared using step responses over a set of representative canonical commands at various speeds (e.g., slow, normal, fast). These include forward and backward walking, sidestepping, turning in place, and circular forward walking. Each command rollout starts with the robot standing still; a rectangular pulse command is then applied and held for a duration before returning to zero. For each command, we report the steady-state tracking error and the mean absolute error over the entire rollout to capture differences in the step transients. Results for Atlas R1 are provided in \Cref{tab:sim2real_analysis}. Similar sim-to-real comparison results for Atlas D1 and Unitree G1 are provided in \highlight{Table~S2} of the Supplementary Materials.

Overall, the tracking errors between simulation and hardware rollouts are closely matched, demonstrating good transfer of the policy to hardware. The differences in tracking errors are quite narrow across the range of commands, generally bounded for most commands within \SI[per-mode=repeated-symbol]{0.05}{\metre\per\second} and \SI[per-mode=repeated-symbol]{0.10}{\radian\per\second} for the linear and angular errors, respectively. As expected, tracking errors increase for more aggressive commands. For example, during the high-speed turning in place command at \SI[per-mode=repeated-symbol]{3.0}{\radian\per\second}, both the steady-state and MAE increase. However, this trend is consistent across both the simulation and hardware rollouts and is not due to poor sim-to-real transfer.
These results highlight that \methodtitle~develops a policy that deploys seamlessly to physical hardware with consistent performance across standard locomotion modalities. Nevertheless, the subtle divergence observed under high turn-rate commands indicates that the sim-to-real gap widens slightly in the most aggressive regions of the command space.

\subsection*{Application of Learned Locomotion Behaviors}

We demonstrate the utility of the \methodtitle~policy across three representative scenarios: direct operator control via a joystick to command \setwo~velocity targets, autonomous navigation with obstacle and hazard avoidance, and a box parkour task where \methodtitle~is integrated alongside separately learned climbing behaviors.
These demonstrations are intended to show that the learned policy can serve as a low-level locomotion module driven by either direct user joystick inputs or higher-level controllers.

During joystick-controlled deployments on physical hardware, the policy is stable and responsive to a broad range of commands, both in controlled lab environments and real-world settings. Notably, the policy exhibits smooth emergent transitions between distinct gait patterns: slow shuffling steps for low-velocity commands, standard-pace walking with nominal strides for mid-range commands, and high-velocity sprinting characterized by large strides and distinct flight phases. 

As the policy uses a lightweight MLP architecture, it can run sufficiently fast onboard for real-time control across all robot platforms considered. For the Unitree G1, we measured the average policy inference time to be \SI{1.1}{\milli\second} on the robot's NVIDIA Jetson AGX Orin's CPU using ONNX Runtime\footnote[3]{Policy inference timings on Boston Dynamics platforms are omitted due to confidentiality agreements.}\cite{onnxruntime}. 

We demonstrate the policy's modularity and applicability to downstream tasks by integrating it into hierarchical control stacks where a higher-level controller provides \setwo~steering commands to the \methodtitle~policy, as illustrated in \customref{fig:teaser}{E}. In the first demonstration, \methodtitle~is deployed as the locomotion layer within a perceptive waypoint navigation stack, which plans around obstacles or hazardous terrain to safely reach waypoints within a warehouse environment. In the second, a high-level box parkour policy is trained to orchestrate three sub-policies: the \methodtitle~policy, a box-climb-up policy, and a box-climb-down policy. Given a goal waypoint, which may be on the ground or on top of a box, the high-level policy sequences and generates steering commands for the sub-policies to move the robot to the waypoint. These demonstrations are intended to illustrate how the \methodtitle~policy can be driven by higher-level controllers. While the detailed design and evaluation of the higher-level controllers themselves are outside the scope of this work, an implementation summary of the high-level box parkour policy is outlined in the Supplementary Materials  (\highlight{Section S7}) to illustrate how the hierarchical control framework can be structured. 

\ifdefined\isArXiv
\begin{figure*}[t!]
\centering
\includegraphics[width=\textwidth]{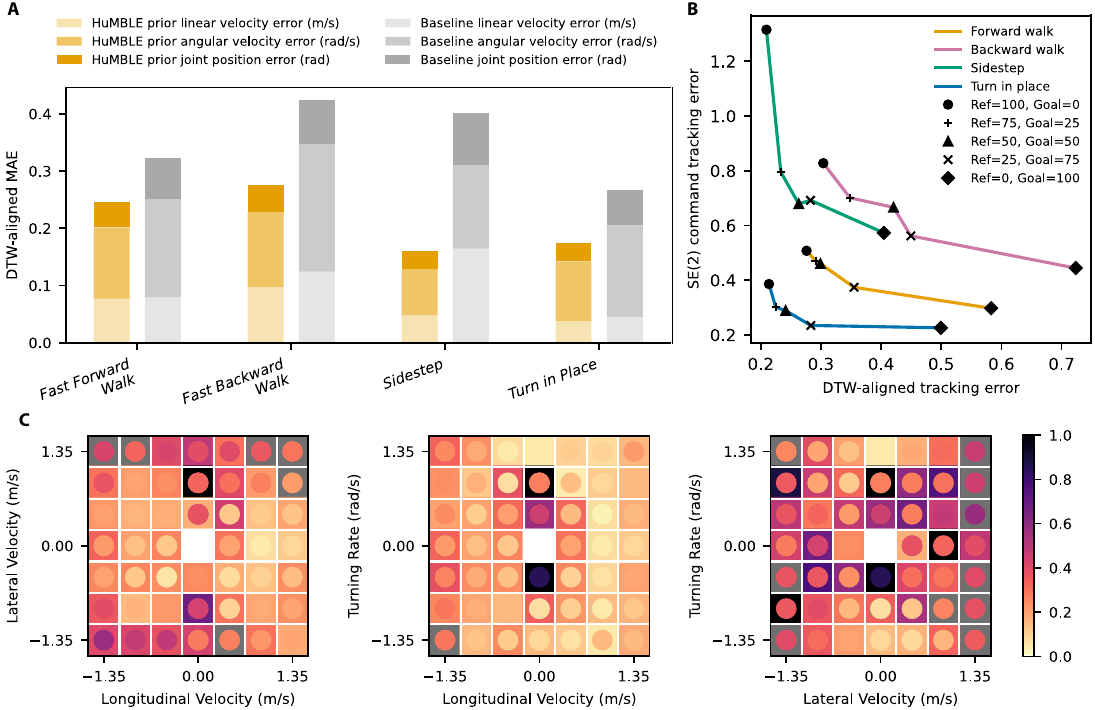}    
\caption{\ablationstudycaption}
\label{fig:g1_ablation}
\end{figure*}
\fi

\subsection*{Ablation Studies}

\noindent To evaluate the core design choices within \methodtitle, we conduct systematic simulation-based ablations addressing three central questions:
\begin{itemize}
\item \textbf{Role of teacher--student distillation in Stage~I:} Is teacher--student distillation necessary for learning a reliable command-conditioned motion prior, or can such a prior be trained directly from the reference motions using imitation rewards under the deployment observation space?
\item \textbf{Efficacy of multi-task RL in Stage~II:} Given a learned motion prior, does the proposed multi-task RL formulation successfully balance the competing objectives of expanding command coverage and preserving stylistic consistency with the motion dataset?
\item \textbf{Impact of RL fine-tuning in Stage~II:} Relative to the Stage~I prior, what quantitative gains in command responsiveness and perturbation robustness are obtained through the Stage~II fine-tuning process?
\end{itemize}

\subsubsection*{Role of Teacher--Student Distillation}
The objective of Stage~I is to learn a prior policy that reconstructs full-body locomotion from proprioceptive observations and \setwo~steering commands in in-distribution settings. 
Here, \emph{in-distribution} refers to rollouts in which the steering commands and reference trajectories are drawn directly from the motion dataset, meaning the commanded velocities perfectly match the intentions of the motion data.
This setting poses two coupled challenges: the policy must realize kinematic reference motions on a physical system subject to dynamics and physical constraints, while also inferring full-body coordination from \setwo~commands rather than full-body reference states. 

\methodtitle~addresses these challenges through teacher--student distillation. To evaluate this design choice, we compare our approach against a single-stage RL baseline that directly trains a command-conditioned policy, $\pibaselineone(\odep)$, without a separate teacher. 
This baseline uses the deployment observation space and is optimized with the same tracking reward, $r_{\text{teacher}}$, forcing it to reconstruct the reference trajectories from proprioception and steering commands alone. For a controlled comparison, $\pibaselineone$ uses the same asymmetric critic architecture and hyperparameters as the teacher-training phase.

We evaluate the data-tracking fidelity of the \methodtitle~prior policy and this single-stage baseline on the simulated Unitree G1 robot, using the same metric as in the data-fidelity analysis. The results are summarized in \customref{fig:g1_ablation}{A}.

For several canonical motion segments extracted from the reference motion dataset, including \emph{Fast Forward Walk}, \emph{Fast Backward Walk}, and \emph{Turn in Place}, the single-stage baseline reproduces the reference motion but with consistently higher error than the \methodtitle~prior. Across these three motions, the baseline increases the tracking error by 16.36\% for linear velocity, 54.85\% for angular velocity, and 70.87\% for joint positions relative to the \methodtitle~prior policy. The gap becomes substantially larger for \emph{Sidestep} motion, a coordination-heavy behavior that requires lateral balance and precise full-body timing. In this case, the baseline fails to reconstruct the reference behavior: after taking a few initial steps, the robot settles into a near-standing posture rather than responding to the steering command. This failure leads to increases in errors of 240.87\%, 80.54\%, and 186.34\% for linear velocity, angular velocity, and joint positions, respectively. 

These results indicate that single-stage partial-observation imitation can recover some locomotion modes, but is less reliable at preserving the full diversity of behaviors in the motion dataset. Teacher--student distillation improves this process by separating the two sources of difficulty: the teacher first learns a dynamically feasible realization of the full-body references in simulation, and the student then distills this behavior into the deployment observation space. This staged formulation yields a more reliable command-conditioned prior for the subsequent fine-tuning stage.

\subsubsection*{Efficacy of the Multi-Task RL}

Stage~II of \methodtitle~uses a multi-task RL framework to balance two competing objectives: improving command responsiveness and coverage while preserving the locomotion style learned in Stage~I. The long-horizon goal-conditioned task improves responsiveness and robustness across a broad range of \setwo~steering commands, including those that are sparse or absent in the motion dataset. In parallel, the short-horizon reference-guided imitation task regularizes the policy toward the human-motion prior, preventing the fine-tuning process from sacrificing natural gait patterns to satisfy aggressive velocity targets. In the nominal \methodtitle~training configuration, we allocate the parallel simulation environments equally between these two task groups, using a 50:50 split to balance command generalization and stylistic consistency.

To analyze how the two task objectives trade off command responsiveness against motion fidelity, we sweep the allocation ratio between reference-guided and goal-conditioned environments. Starting from the same Stage~I prior, we fine-tune policies with five environment splits: 0:100, 25:75, 50:50, 75:25, and 100:0, where the first value denotes the percentage of reference-guided environments and the second denotes the percentage of goal-conditioned environments. We then evaluate each policy for both reference-tracking fidelity and command tracking.

The evaluation is performed across four canonical motion modes: forward walk, backward walk, sidestep, and turn in place. For reference-tracking fidelity, the policy is driven by the time-varying \setwo~commands annotated from the corresponding reference motion, and we report the DTW-aligned full-body tracking error relative to that reference trajectory. Specifically, we pair each motion modes with the same benchmark segments used in our previous experiments: \emph{Fast Forward Walk}, \emph{Fast Backward Walk}, \emph{Sidestep}, and \emph{Turn in Place}, respectively. Meanwhile, for velocity command tracking, the policy is driven by a constant command associated with each mode: $(1.35, 0, 0)$ for forward walk, $(-1.35, 0, 0)$ for backward walk, $(0, 1.35, 0)$ for sidestep, and $(0, 0, 1.7)$ for turn in place. In this regime, we report the mean velocity-tracking error.

\customref{fig:g1_ablation}{B} shows how motion fidelity and command-tracking performance vary with the allocation ratio between reference-guided and goal-conditioned environments.
Although the trends differ somewhat across motion modes, the results reveal a consistent trade-off: increasing the fraction of reference-guided environments improves fidelity to the motion data, whereas increasing the fraction of goal-conditioned environments improves velocity command tracking.
The resulting curves therefore characterize the Pareto trade-off induced by the multi-task fine-tuning objective.

A detailed inspection of the extreme ratios further highlights this fundamental balance. The 0:100 policy, trained exclusively on goal-conditioned environments, rapidly loses stylistic consistency with the motion data. During forward and backward walking, it exhibits wiggly torso movement and unnatural arm swings that are inconsistent with the reference motions. Conversely, the 100:0 policy, trained solely on reference-guided environments, fails to execute the lateral command for sidestepping and falls. Because lateral steering commands are sparse in the dataset, this failure underlines the vulnerability of reference-guided policies under out-of-distribution conditions. Together, these extreme cases validate our multi-task RL design in balancing command generalization with motion quality. 

While simultaneously optimizing both objectives remains challenging, the proposed multi-task formulation provides a controllable mechanism for balancing stylistic consistency and command responsiveness.
In Stage~II, the environment allocation ratio serves as an interpretable hyperparameter that can be adjusted to match the desired deployment behavior.
The nominal 50:50 allocation provides a practical middle ground, preserving motion fidelity while improving command tracking for general-purpose steerable locomotion.

\ifdefined\isArXiv
\else
\begin{figure*}[t!]
\centering
\includegraphics[width=\textwidth]{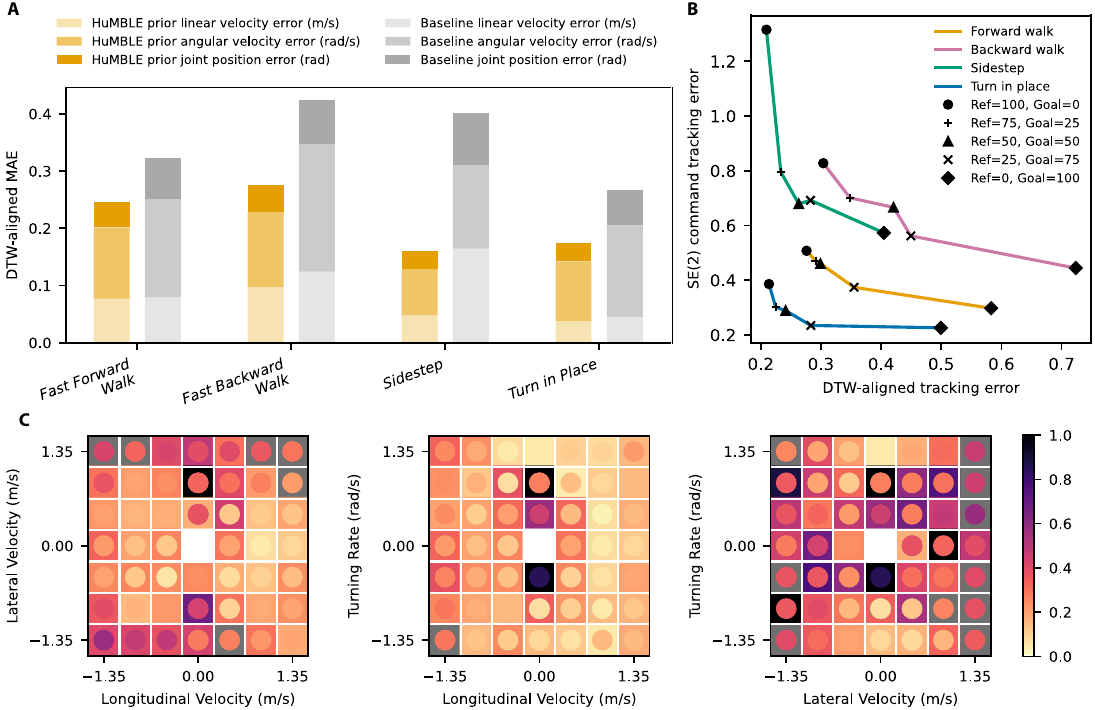}    
\caption{\ablationstudycaption}
\label{fig:g1_ablation}
\end{figure*}
\fi

\subsubsection*{Impact of RL Fine-Tuning}

The Stage~I prior policy can respond to the in-distribution steering command derived from a motion dataset, but the dataset may not cover the full command space necessary for real-world deployment under arbitrary user commands. Stage~II of \methodtitle~aims to address this by improving the command coverage of the final policy for robust execution under arbitrary steering commands.

In this final ablation study, we evaluate the command coverage of the final deployment policy after Stage~II RL fine-tuning and directly compare its performance with that of the Stage~I prior policy. 
Following the step-response evaluation methodology established in our primary command coverage analysis, we apply a series of target \setwo~velocity step inputs to both policies on a simulated Unitree G1 robot and measure the steady-state command tracking error.

\customref{fig:g1_ablation}{C} illustrates the comparative results of this evaluation. The Stage~I prior policy maintains a reasonable level of command-tracking proficiency in regions where the underlying motion-data distribution is dense. However, its performance degrades rapidly in regions where the reference data is sparse or absent. Specifically, across several high-velocity command regimes, the prior policy fails to respond robustly to the steering command, resulting in dynamic instability and frequent falls. Most notably, this degradation becomes particularly pronounced during lateral motion paired with yawing---the sparsest region in our dataset, as visualized in \Cref{fig:r1_data_distribution}---which inevitably leads to falls. Conversely, the fine-tuned deployment policy exhibits consistently lower tracking errors and vastly expanded command coverage across the entire operational space. Notably, across all test cases, the deployment policy experiences no falls and reliably generates locomotion behavior associated with the target velocity commands without collapsing into unnatural gaits.

These results demonstrate that teacher--student distillation alone is insufficient to achieve the high level of responsiveness and broad command-space coverage required for real-world deployment. Instead, executing the multi-task RL fine-tuning phase in Stage~II is critical to ensuring that the humanoid locomotion policy functions robustly across general-purpose locomotion tasks.


\section*{DISCUSSION}

\noindent 
The goal of this work is to develop a humanoid locomotion controller that simultaneously achieves two distinct criteria: first, preserving the stylistic quality and natural transitions of human movement; second, maintaining responsiveness to user steering commands across a wide operational envelope.
The \methodtitle\ framework targets these objectives by combining a newly curated human motion dataset with a two-stage learning pipeline. Our motion dataset captures a highly diverse repertoire of human locomotion, providing the data density necessary to span wide variations in steering speeds and omnidirectional headings. 
Employing this dataset as a foundation, our two-stage training pipeline decouples style acquisition from command-space expansion. Initially, it focuses on internalizing and reconstructing the complex full-body behaviors solely from proprioception and the velocity command. Subsequently, the framework expands the command coverage beyond the original dataset's support. Notably, it achieves this while maintaining the stylistic quality established in the prior policy. Because style is preserved by an explicit reference-tracking reward rather than a learned or latent style objective, the balance between the style and command-tracking objectives remains directly interpretable.

Extensive empirical evaluations across three distinct humanoid platforms validate that \methodtitle~successfully bridges the gap between stylistic quality and responsive command tracking across a wide spectrum of steering inputs. The final deployment policies consistently reproduce distinct biological features such as synchronized arm swinging, knee-locking, and heel-strike transitions, which remain difficult to elicit through reward engineering alone.
Furthermore, our quantitative analysis and ablation study demonstrate that \methodtitle~effectively mitigates the trade-off between stylistic fidelity to motion data and high responsiveness across a broad command space, providing a reliable option for direct command control in general-purpose locomotion tasks. 
\methodtitle\ embeds these capabilities into a single compact MLP network with minimal memory overhead and low computational complexity. This lightweight architecture streamlines the deployment of natural, robust behaviors directly onto physical robotic hardware across diverse real-world scenarios.

In summary, the key contributions of \methodtitle\ span from the compilation and release of a large-scale human motion dataset curated for learning naturalistic locomotion behaviors, through a structured two-stage learning framework designed to decouple and optimize style preservation alongside command expansion, to a rigorous evaluation methodology devised to benchmark the data fidelity, command coverage, responsiveness, and robustness of data-driven humanoid policies, as demonstrated via stable, steerable locomotion across diverse physical robotic platforms.

Despite these contributions, several limitations remain, which conversely present promising opportunities for future research. First, our results are restricted to standard humanoid locomotion on flat terrain environments. Although an immediate extension to mild terrain roughness is feasible through more extensive domain randomization, scaling to more general locomotion scenarios, such as staircase climbing \cite{zhang2026learningwholebodyhumanoidlocomotion} or executing whole-body movements like climbing \cite{wang2026generalizingre} or vaulting over obstacles \cite{zhuang2026deepwholebodyparkour, wu2026perceptive} presents an open challenge. These advanced locomotion modalities require a new, dedicated set of human motion references that capture these agile behaviors. They also necessitate training a policy conditioned on terrain observations, which requires the integration of exteroceptive sensory inputs at deployment. Integrating these diverse behavioral and sensory modalities to explore the boundaries of our two-stage learning framework represents a compelling direction for future work.

Second, \methodtitle~requires a command annotation process to train a steerable policy conditioned on velocity commands. In this work, we addressed this via post-processing after dataset collection, estimating the performer's movement intentions and annotating each motion frame with the estimated \setwo~velocity command. While this approach is relatively straightforward for \setwo~commands, which can be directly computed from the reference trajectory itself, it may not scale easily to more complicated steering modes. For instance, incorporating high-level semantic annotations, such as natural language command inputs \cite{petrovich2021action, Tevet2023Human, Tessler2024MaskedMimic}, may require manual labeling or more advanced captioning techniques.

Third, in our experiments, we deliberately chose a simple MLP with a modest number of learnable parameters to minimize computational and memory overhead at deployment. Our results demonstrate that this compact architecture is sufficient to internalize and reconstruct full-body behaviors from velocity command inputs, capturing the underlying distribution of a motion dataset containing $N=\numsamples$ frames, which is equivalent to approximately three hours of human locomotion data. However, we leave a rigorous scalability analysis of our framework using this simple architecture to future work. Determining how the number of parameters affects policy performance, identifying the capacity limits of the chosen architecture, and mapping the exact boundary conditions under which the policy fails to capture the data distribution all remain open questions. We are particularly interested in evaluating the capacity of the \methodtitle~framework, when constrained to a simple network architecture, to capture more diverse human motion behaviors beyond locomotion tasks. Specifically, we intend to explore more complex scenarios, such as whole-body loco-manipulation or humanoid teleoperation, leveraging natural skill sets learned from larger-scale human motion datasets across diverse behavioral modalities. Whether a simple MLP can accommodate these extensions, and whether our two-stage learning pipeline can effectively capture data distributions of this complexity to develop steerable policies with extended command modalities, will serve as central research questions in our future work.

Beyond locomotion, we view \methodtitle\ as a general recipe for turning human motion data into command-driven policies. The framework makes few assumptions specific to locomotion and it can be applied to broader applications whenever the following ingredients are available: a motion dataset, a command signal that can be annotated onto each frame, and a reward that measures tracking of that command. In practice, this means that once the comparatively simpler task of training a motion imitation policy has been addressed for a given robot, the setup can be extended to a command-conditioned policy using the proposed recipe. Other under-constrained command modalities are a natural next step, such as teleoperation from sparse hand and base inputs, where the policy must reconstruct full-body behavior, much as \methodtitle~does for \setwo\ locomotion commands.




\section*{MATERIALS AND METHODS}

%
%
\begin{figure*}[t!]
    \centering
    \includegraphics[width=\textwidth]{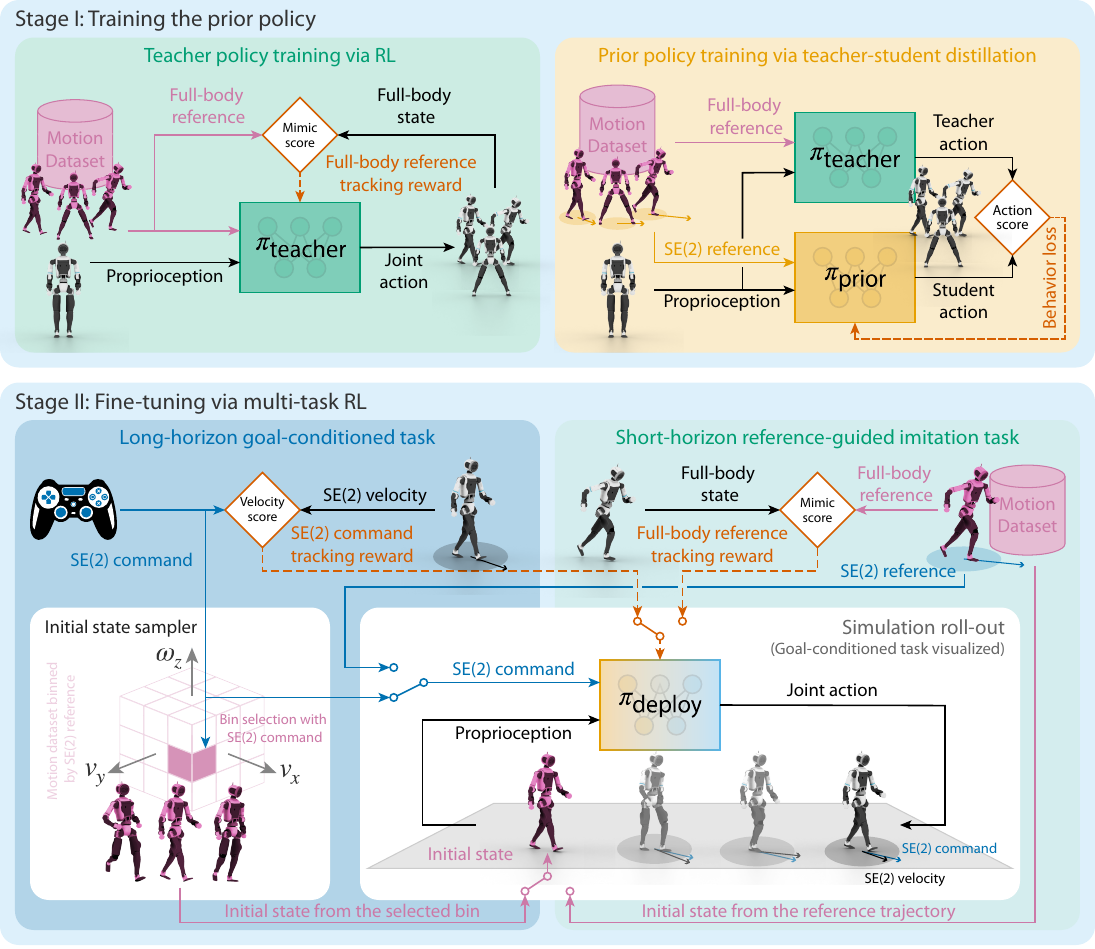}
    \caption{\textbf{Overview of the \methodtitle~framework.} Our approach trains the final control policy $\pideploy$ through a two-stage process. Stage I employs a teacher--student distillation pipeline to extract a natural locomotion prior $\pistudent$. Specifically, a teacher policy $\piteacher$ trained to track full-body reference motions is distilled into the student  policy $\pistudent$, which maps only proprioceptive observations and a \setwo~command to joint-level actions. In Stage II, the prior policy is fine-tuned via multi-task RL to enhance responsiveness to arbitrary user steering commands while preserving the stylistic qualities of the reference motion data, ultimately forming the final deployment policy $\pideploy$.}
    \label{fig:overview}
\end{figure*}   

\noindent
\Cref{fig:overview} summarizes the \methodtitle~training pipeline, which leverages retargeted human locomotion references to train a hardware-deployable, command-conditioned locomotion policy.
The pipeline is structured into two sequential training stages and comprises three policy instances.
In Stage~I, a full-body reference-conditioned teacher policy, $\piteacher$, is trained to track kinematic references from the dataset.
Teacher--student distillation is then used to train a prior policy, $\pistudent$, that observes only the deployment inputs: proprioception and \setwo~velocity commands.
In Stage II, the prior policy is fine-tuned into the final deployment policy, $\pideploy$, using a multi-task RL formulation split into two concurrent task groups: a short-horizon reference-guided imitation task and a long-horizon goal-conditioned command-tracking task.
The resulting policy maps deployment observations onto joint-position targets, which are tracked by joint-level PD controllers.
In the following sections, we describe the motion dataset and command annotations, the deployment policy architecture, the technical details of each training stage, and the hardware deployment procedure.

%
%
\begin{figure*}[t!]
\centering
\includegraphics[width=\textwidth]{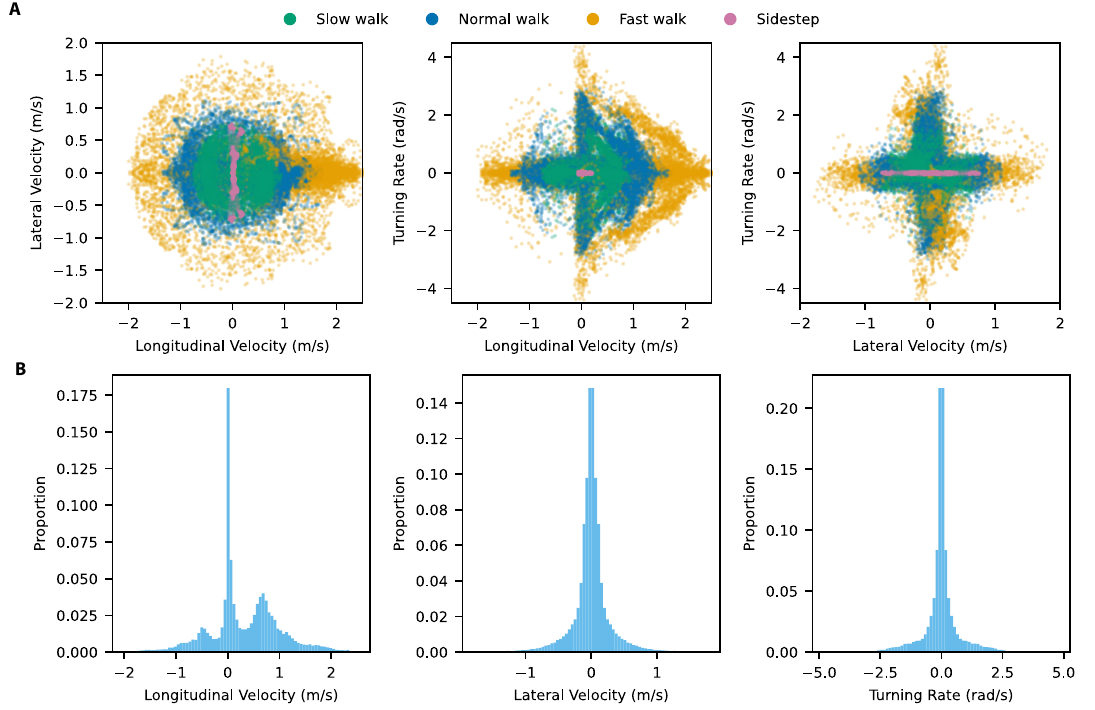}    
\caption{\textbf{Distribution of the human reference motion dataset.} We illustrate the distribution of the annotated commands for our in-house curated human motion dataset, retargeted to Atlas R1. While the dataset also contains high-speed sprinting maneuvers, this visualization focuses exclusively on the standard walking command distribution to clearly illustrate the most densely populated region. \textbf{(A)} Scatter plots illustrating the distribution of annotated velocity commands across four locomotion regimes (fast, normal, slow walk and sidestep). For visual clarity, the number of data points is decimated by a factor of 10. \textbf{(B)} Histograms displaying the marginal proportions of the dataset frames binned by longitudinal velocity, lateral velocity, and turning rate. The y-axis denotes the proportion relative to the dataset size $N=\numsamples$.}
\label{fig:r1_data_distribution}
\end{figure*}

\subsection*{Dataset Collection and Command Annotation}
\label{subsec:dataset}

\noindent We collect high-quality human motion references using a Vicon motion capture system~\cite{vicon_mocap}. Motion data is captured at a frame rate of \SI{120}{\hertz} utilizing a marker set and a character rig calibrated to the actor’s specific anthropometry and range of motion. The raw marker trajectories are post-processed by correcting missing marker data via interpolation and skeletal kinematic constraints, followed by filtering to eliminate high-frequency noise. Finally, the character rig is mapped to the refined marker data and exported in the Biovision Hierarchy (BVH) format, providing a standardized skeletal animation foundation. 

Subsequently, we convert the BVH animations into whole-body robot motion trajectories tailored for the target robotic system through a two-stage motion retargeting process.
In the first stage, the BVH animations are mapped to a \emph{kinematic proxy skeleton} featuring a human-like joint topology with spherical joints, but with link proportions matched to those of the target robot. Alignment frames are defined on source skeleton segments at the head, torso, upper arms, lower arms, hands, pelvis, knees, and feet. We formulate this retargeting as a spacetime optimization problem that minimizes tracking errors and constraint violations between these source frames and their counterparts on the proxy skeleton by solving for the proxy joint poses across all timesteps simultaneously~\cite{Gle98}. Additionally, we temporally warp the trajectory by scaling its time duration by a factor equal to the square root of the spatial scaling factor, preserving dynamic consistency under a nominal gravitational acceleration of \SI[per-mode=repeated-symbol]{9.81}{\metre\per\square\second}.

The second stage applies kinematic retargeting to map the resulting proxy motions onto the target robot's morphology. To do so, we minimize constraint violations and regularization costs across all timesteps. Similar to the first stage, alignment frames are placed on the target robot to specify spacetime constraints that guide retargeting. To regularize under-constrained joints and links, small penalties are imposed to minimize joint velocity and maintain a neutral posture. Finally, collision primitives are specified on the robot and the environment, with non-penetration constraints imposed to prevent self-collisions and ground interpenetration. The resulting reference motions encompass diverse locomotion behaviors ranging from slow-paced walking to high-speed sprinting with varied headings. 

Once the robot motion trajectories are prepared, we annotate each motion frame with an estimated \setwo~velocity command representing the performer's movement intentions, which are defined by longitudinal (forward-backward) and lateral (sideways) velocities, along with the turning rate. 
To compute this annotation, we estimate the torso linear and angular velocities at each frame via finite differencing, express them in the torso frame, and apply a low-pass filter. We then shift the resulting velocity signals backward in time by an amount proportional to the torso acceleration, under the assumption that future torso velocity serves as a proxy for current movement intention. The longitudinal and lateral command velocities are extracted as the $x$ and $y$ components of the processed linear velocity, respectively, while the turning rate is defined as the yaw component of the processed angular velocity.

Finally, we augment the motion dataset by mirroring all trajectories across the robot's sagittal plane to achieve broader and more balanced motion coverage. This mirroring also enforces bilateral symmetry in the learned locomotion patterns, ensuring the policy exhibits consistent behavior when executing left or right steering. 

\Cref{fig:r1_data_distribution} shows the distribution of the annotated \setwo~velocity commands for Atlas R1. The dataset contains \numsamples~samples, visualized using scatter plots and density histograms over the \setwo~velocity command space. The distribution shows that the data collection scenarios were designed to cover a broad operational envelope within the locomotion space, including variations in longitudinal and lateral velocities and turning rates. Additional details on the dataset collection pipeline, including the motion choreography protocol, kinematic post-processing, and the temporal adjustment used for command annotation, are provided in the Supplementary Materials (\highlight{Section S2}).

\subsection*{Deployment Policy}
\label{subsec:training-overview}

\noindent The final outcome of our training pipeline is a control policy, $\pi_{\text{deploy}}$, that exhibits locomotion styles consistent with the human motion prior while remaining highly responsive to user steering commands. This policy is parameterized as a single MLP neural network that maps the deployment observations directly to the joint actions. For all robot platforms, the policy is inferred at a frequency of \SI{50}{\hertz}. 

At time step $t$, the deployment action $\bm a_t$ is element-wise multiplied by pre-defined scaling factors and offset by the nominal joint positions to yield the desired joint position targets, which are subsequently tracked by low-level, joint-level PD controllers. The PD gains are designed based on the projected joint armature value for each joint\footnote[4]{While the joint-level parameters for the Boston Dynamics platforms are omitted due to confidentiality agreements, we provide the corresponding details for the Unitree G1 platform in the Supplementary Materials (\highlight{Table~S7}).}, following the procedure detailed in \emph{ZEST} \cite{sleiman2026zest}.

The deployment observation, $\odep$, consists of proprioceptive observation, $\oprop$, and the steering command, $\ocmd$:
\begin{equation}
\label{eq:observation_policy_nominal}
    \odep=(\oprop,\ \ocmd).
\end{equation}

The proprioceptive component, $\oprop$, is restricted to onboard measurements such that the policy is deployable on hardware:
\begin{equation}
{\oprop=\big({}_\mathcal{T} \linvel_{\mathcal{IT}},\ {}_\mathcal{T} \angvel_{\mathcal{IT}},\ {}_\mathcal{T} \gravity,\ \jointpos,\ \jointvel,\ \bm a_{t-1}\big)},
\end{equation}
here, ${}_\mathcal{T} \linvel_{\mathcal{IT}}$ and ${}_\mathcal{T} \angvel_{\mathcal{IT}}$ denote the linear and angular velocities measured within the IMU sensor frame $\{\mathcal{T}\}$. The term ${}_{\mathcal{T}} \gravity$ is the projected gravity vector expressed in $\{\mathcal{T}\}$. We note that the linear-velocity observation ${}_\mathcal{T} \linvel_{\mathcal{IT}}$ relies on the robot's state estimate during deployment. 
$\jointpos,$ and $\jointvel$ represent instantaneous joint positions and velocities and $\bm a_{t-1}$ is the previous action. 

The command observation, ${\ocmd=\big(\hat{v}_{\text{long}},\ \hat{v}_{\text{lat}},\ \hat{\omega}_{\text{turn}}\big)}$, defines the \setwo~steering velocity command. It consists of the target longitudinal and lateral linear velocities, together with the turning rate.

\subsection*{Stage I: Training the Prior Policy}
\label{subsec:stage_1}
\noindent 
Stage~I focuses on training the prior policy, $\pistudent$, which internalizes and reconstructs the human motion patterns present in the dataset. Because the final deployment policy, $\pideploy$, is obtained by fine-tuning $\pistudent$ (as detailed in Stage II), both $\pideploy$ and $\pistudent$ have identical neural network architectures and share the same observation (i.e., $\odep$) and action spaces. 
The deployment observation does not include the full-body reference states. Therefore, $\pistudent$ must be capable of reconstructing full-body motions consistent with the reference dataset from only proprioception and \setwo~velocity commands. To address this partial observability, we employ a teacher--student distillation approach. 
A teacher policy, $\piteacher$, is trained using an RL-based motion imitation framework \cite{sleiman2026zest}, with access to privileged full-body reference states from the motion dataset, to generate actions that drive the robot to match the reference trajectory as closely as possible. Subsequently, the prior policy, $\pistudent$, is trained via imitation learning to match the actions of $\piteacher$. The resulting $\pistudent$ learns to map the non-privileged deployment observations to actions that produce reference-consistent motions for states in-distribution of the dataset.

\subsubsection*{Teacher Policy Training via RL}
\label{subsec:teacher_training}

\noindent To train the teacher policy, $\piteacher$, we adopt the setup from \emph{ZEST}~\cite{sleiman2026zest}, utilizing a similar observation space, reward structure, termination criteria, and the assistive wrench curriculum. This training setup enables a single-teacher policy to reliably replicate the diverse locomotion behaviors in our dataset. 

%
In contrast to the deployment observation, the teacher policy's observation space incorporates full-body reference states in addition to the deployment observation. Formally, the teacher policy observation is defined as 
\begin{equation}
    \oteacher=(\odep,\ \oref),
\end{equation}
where ${\oref=\big({}_{\mathcal{I}} \hat{\bm r}^z_{\mathcal{IB}},\ {}_{\mathcal{B}} \hat{\bm v}_{\mathcal{IB}},\ {}_{\mathcal{B}} \hat{\bm \omega}_{\mathcal{IB}},\ {}_{\mathcal{B}} \hat{\bm g}_{\mathcal{I}},\ \hat{\bm q}_j\big)}$ contains the target states of the imitation: ${}_{\mathcal{I}} \hat{\bm r}^z_{\mathcal{IB}}$ denotes the reference torso height; ${}_{\mathcal{B}} \hat{\bm v}_{\mathcal{IB}}$ and ${}_{\mathcal{B}} \hat{\bm \omega}_{\mathcal{IB}}$ represent the  linear and angular torso velocities references expressed in the torso frame $\{\mathcal{B}\}$; ${}_\mathcal{B} \hat{\bm g}_{I}$ is the reference projected gravity in $\{\mathcal{B}\}$; and $\hat{\bm q}_j$ are the reference joint positions.

The teacher reward function is composed of three main terms: reference tracking, regularization, and a survival term:
\begin{equation}
\label{eq:teacher-reward}
    \rteacher \;=\; \rtrack + \rreg + \rsurvival.
\end{equation}
The reference tracking term, $\rtrack$, encourages the policy to match the robot's state to the reference by penalizing deviation across the state variables. The regularization term, $\rreg$, promotes smooth and physically feasible behaviors. Finally, the survival reward, $\rsurvival = c_{\text{survival}}$, is a constant positive reward granted at each time step discouraging premature termination.

Upon termination, which is triggered either by completing the reference trajectory or early due to excessive deviation from the reference, the environment is reset using Reference State Initialization (RSI) \cite{Peng2018DeepMimic}, where the initial state for a new episode is generated by sampling a state from the reference dataset and perturbing the robot state around that sample.

We use Proximal Policy Optimization (PPO)~\cite{PPO} as the RL algorithm to train the teacher policy, which utilizes a separate critic network to estimate the on-policy value function. To facilitate fast convergence, we adopt an Asymmetric Actor-Critic design~\cite{AsymmetricActorCritic}, where the critic network is granted explicit access to privileged environmental observations and a binary data support indicator:
\begin{equation}
\label{eq:observation_critic}
\bm o_{\text{critic}} = (\odep,\ \oref,\ \opriv,\ \osup),
\end{equation}
where the privileged observation, $\opriv$, includes the robot's torso height, the end-effector positions and velocities relative to the torso frame, the assistive wrench level, and the generalized end-effector forces. 

The binary data support indicator observation, $\osup \in \{0, 1\}$, is used to distinguish the task context. A value of zero indicates that the steering command lies within the dataset support (in-distribution), while a value of one corresponds to out-of-distribution commands. Throughout Stage~I teacher training, this value remains strictly zero, as all steering commands are derived directly from the reference motion data. However, this indicator is dynamically modulated during the subsequent multi-task fine-tuning phase in Stage~II.

Further details on the teacher policy training process, including comprehensive definitions of rewards (\highlight{Section~S3}) and hyperparameters (\highlight{Table~S9}), are provided in the Supplementary Materials.

\subsubsection*{Prior Policy Training via Teacher--Student Distillation}
\label{subsec:student_distillation}

\noindent Once the teacher policy is trained to reliably produce reference motions conditioned on the full-body reference input, we train the prior policy, $\pistudent$, using teacher--student distillation \cite{RL_Joonho, Miki2022WildLocomotion, Tessler2024MaskedMimic}. This process is a variant of the DAgger algorithm \cite{Ross2011DAgger}, where the student policy (in our framework, the prior policy) generates rollout data that are subsequently labeled with the actions the teacher policy would have taken at the same state. 
We therefore collect rollouts using $\pistudent(\odep)$, and for each state encountered during these rollouts, $\piteacher(\oteacher)$ acts as an expert to provide the ground-truth action labels. 
We then optimize the prior policy in a supervised learning fashion, minimizing the difference between the prior policy's actions and the teacher's labels:
\begin{equation}
    \pistudent = \arg\min_{\bm{\pi}} E_{\pi} \| \bm{\pi}(\odep) - \piteacher(\oteacher) \|.
\end{equation}

While the teacher uses a full-body reference observation, the prior policy operates under partial observability, strictly restricted to the deployment observation space $\odep$. This constraint forces the prior policy to internalize the motion priors and recovery strategies originally guided by the teacher's privileged reference observation. By distilling the teacher's expertise, the prior policy learns to exhibit natural, high-fidelity locomotion solely from proprioceptive feedback and \setwo~velocity commands.

\subsection*{Stage II: Fine-tuning via Multi-Task RL}
\label{subsec:student_finetuning}
\noindent The prior policy trained in Stage~I maintains high fidelity to the motion dataset. However, it struggles with out-of-distribution velocity commands that are underrepresented in the motion dataset. For example, diagonal strafing combined with a fast turning rate is within the space of commands reachable via a joystick but uncommon in human locomotion. 
To address this, Stage II focuses on expanding command coverage and enhancing the policy's robustness by fine-tuning the prior policy to track velocity commands drawn from a distribution that matches the operational range of interest.

However, naively fine-tuning the policy with a reward that incentivizes tracking arbitrary velocity commands typically leads to style collapse, in which the policy compromises the natural movement patterns learned in Stage I to better track those commands. To mitigate this more effectively, we formulate Stage II as a multi-task RL problem, sharing conceptual similarities with concurrent work \cite{wang2026generalizingre}. 
We take advantage of parallel training environments by partitioning the environments into two concurrent groups.
The first subset of environments, $\envref$, is dedicated to a \emph{short-horizon reference-guided imitation task}. Here, the policy continues to track the reference dataset, acting as a regularizer to preserve stylistic consistency acquired in the previous stage. The remaining environments, $\envgoal$, focus on a \emph{long-horizon goal-conditioned task}. In this task, the policy is optimized for robust tracking of randomly sampled velocity commands, independently of any motion-reference tracking objective. Formally, the fine-tuning reward, $\rfinetune$, is structured as
\begin{equation}
  \rfinetunestructure
\end{equation}

We design this multi-task RL setup to carefully balance stylistic consistency against command coverage, responsiveness, and robustness. 
Conceptually, the reference-guided environment acts as style-regularizing experience replay, preventing the policy from diverging into unnatural gaits while the goal-conditioned environment expands the policy's operational envelope beyond the reference motion dataset's distribution. In the following sections, we describe the specific design choices and reward formulations for these two environments in detail.

\subsubsection*{Short-horizon Reference-Guided Imitation Task}
\noindent The reference-guided environments follow a setup similar to that of the teacher policy, using the same reward function $\rteacher$ presented in \Cref{eq:teacher-reward} and the same initialization scheme (RSI) upon termination. The primary distinction is that the policy observes only the deployment observation space, $\odep$, and has no access to the full-body reference, $\oref$.

In this partially observable setting, the policy must reconstruct the reference motion solely from the command and proprioception. Although the policy was pre-trained via distillation to perform this reconstruction, minor tracking errors are inevitable: because the policy lacks direct observation of the reference, it cannot as effectively produce corrective actions once it deviates too far from the reference trajectory. For long episodes, these errors compound, causing the policy to drift from the reference and eventually reaching states where the reference-tracking reward is no longer informative. Therefore, we deliberately set a short episode length for the reference-guided task to focus the reinforcement signal on local style preservation and prevent the optimization from being derailed by long-term cumulative drift.

\subsubsection*{Long-horizon Goal-Conditioned Task}
\noindent The goal-conditioned environments train the agent to track randomly sampled commands. Unlike the reference-guided task, these environments do not use a reference motion trajectory to compute reward. Instead, the reward function is defined as: 
\begin{equation}
    \rgoal = \rcommand + \rreg + \rsurvival,
\end{equation}
where the command tracking term, $\rcommand$, incentivizes the agent to match the \setwo~velocity of the robot torso to the sampled command.
Both the reference-guided imitation and goal-conditioned tasks share a common set of reward components, specifically the regularization rewards, $\rreg$, and the survival rewards, $\rsurvival$, alongside a common set of termination criteria such as terminating upon excessively large contact forces. These elements are identical to those utilized during teacher training. 
For a comprehensive breakdown of the reward formulations and termination criteria, we refer readers to \highlight{Section~S4} and \highlight{Table~S4} of the Supplementary Materials, respectively.

We set a longer episode length for the goal-conditioned environments, allowing commands to be resampled within an episode to expose the policy to these transitions. In addition to the standard termination criteria, an episode is terminated if the robot falls into an unrecoverable state and cannot continue tracking the command. For initializing goal-conditioned rollouts, we introduce a scheme named Goal-Conditioned Reference-State Initialization (GC-RSI), which is detailed in the subsequent section.

\subsubsection*{Goal-Conditioned Reference State Initialization}
\noindent In GC-RSI, the target velocity command is used to retrieve a motion frame from the dataset with a similar annotated command. To facilitate efficient retrieval, we precompute a reference-state lookup table by binning reference states according to their annotated commands. Given a query command, we sample a reference frame from its corresponding bin and randomly pick a candidate state from a temporal window centered on that frame. Finally, to encourage broader exploration, we apply a small perturbation to this state to produce the final initial state.

This command-conditioned initialization mechanism serves two critical purposes. 
First, it anchors fine-tuning near states that remain within the learned behavioral prior's support. Since $\pistudent$ has already learned to control motions drawn from the dataset distribution, initializing from a reference state whose annotated command is close to the sampled command allows the policy to rely on its existing prior at the beginning of each episode. This prevents the policy from immediately drifting into out-of-distribution states or developing an isolated behavior mode unrelated to the reference motions. Instead, GC-RSI provides a gradual pathway for expanding the policy's command coverage while preserving the stylistic and dynamical structure acquired during prior training. 
Second, GC-RSI prevents severe mismatches between the commanded velocity and the initial state. This is particularly crucial for high-magnitude commands; forcing the robot to accelerate aggressively from a mismatched pose can cause the episode to terminate prematurely before meaningful learning can occur.

\subsubsection*{Training Details for Multi-Task Fine-Tuning}

We once again employ PPO~\cite{PPO} for the RL-based fine-tuning stage. However, initializing PPO directly from the teacher--student distilled prior policy, $\pistudent$, introduces a cold-start problem: randomly initializing the PPO critic network can rapidly erode the stylistic nuances learned during prior training as the actor is updated using noisy, unreliable value estimates from an untrained critic.
To address this, we initialize the fine-tuning process by loading both the policy network of $\pistudent$ and the critic network from the teacher training stage, $\vteacher$. Because $\vteacher$ was trained to approximate the value function associated with the teacher policy $\piteacher$, rather than $\pistudent$, it must be adapted before actual multi-task fine-tuning. 
We therefore perform an initial value-function warm-start phase of 5000 iterations on the reference-guided task ($\vteacher \rightarrow \vstudent$), training only the critic while keeping the policy fixed.
Once this adaptation phase is complete, we transition to full multi-task RL to jointly train the deployment policy and critic networks ($\pistudent \rightarrow \pideploy$ and $\vstudent \rightarrow \vdeploy$). 

Because the reward functions differ between the goal-conditioned and reference-guided tasks, the critic must predict task-appropriate returns. To achieve this, we leverage the data support indicator observation $\osup \in \{0, 1\}$ introduced in \Cref{eq:observation_critic}, which is provided exclusively to the critic network as a context signal to distinguish the active task domain. While this indicator is kept strictly at zero throughout Stage~I teacher training, during Stage~II fine-tuning, it is specified by the environment type: it remains zero in short-horizon reference-guided environments and is set to one in long-horizon goal-conditioned environments. This contextual modulation allows the critic network to effectively learn a dual value function conditioned on the task support, yielding more accurate on-policy value estimations. Moreover, we separately normalize the critic's estimated advantages for the reference-guided and goal-conditioned environments, an approach similarly employed in \emph{RobotKeyframing} \cite{Zargarbashi2024Robot} to address potential scale differences between the two reward functions.

Finally, to further safeguard the internalized stylistic characteristics against large policy updates, we lower the policy entropy coefficient, and utilize a small learning rate compared to the previous stage.

\subsection*{Hardware Deployment}
\label{subsec:hw_deployment}
\noindent Following the two-stage training process, the policy is deployed to the robot hardware zero-shot, serving as a steerable locomotion controller that guides the robot to exhibit human-like behaviors in response to real-time joystick commands. For deployment with a joystick, the steering inputs are processed to bound their rate of change, as we found this resulted in a smoother user driving experience at some expense of the policy's responsiveness.

The linear velocity observation for the policy relies on the platform’s onboard state estimator. The Boston Dynamics Atlas robots are equipped with a robust state estimator that provides reliable linear-velocity feedback in both indoor and outdoor environments. 
However, deploying the G1 robot in unstructured outdoor environments can introduce significant noise and drift into the velocity estimation. To mitigate potential system failure in these scenarios, we trained an alternative policy variant that excludes linear velocity observations. This linear-velocity-free policy ensures stable operation, albeit at the cost of reduced velocity-tracking precision relative to the target command.

\section*{Acknowledgments}
We would like to thank Ashley Dodge and Matthew Tran for their assistance with the Unitree G1 hardware experiments, as well as Maria Krinner Anegón and Alexander Liniger for their help with the Unitree G1 autonomous navigation demonstration. Additionally, we extend our gratitude to Jakob Genhart for developing visualization tools used in the supplementary videos. We note that while all ideas, methods, and results are original to the authors, AI-based tools were used solely to refine the manuscript's clarity and readability. 
\textbf{Author contributions:} The \textit{Core Contributors} designed and implemented the \methodtitle~pipeline; trained the policies, and oversaw the hardware experiments on robots. They also contributed to the writing of the manuscript. 
The \textit{Additional Contributors} assisted with engineering tasks to support the project, including deployment code on hardware, data collection, and supporting hardware experiments. 
The \textit{Project Leads} set the overarching vision by defining the core scientific goals. They provided technical guidance on significant decisions and coordinated the team’s efforts. Finally, they drive the narrative of the paper. 
\highlight{The work on Boston Dynamics Atlas robots involved all authors, while authors affiliated with the RAI Institute completed the work on the Unitree G1 robot.}
\textbf{Competing interests:} The authors declare that they have no competing interests. 
\textbf{Data and materials availability:} All data needed to evaluate the conclusions in the paper are present in the main manuscript or the Supplementary Materials. \highlight{While data for the Boston Dynamics Atlas platforms are proprietary and cannot be publicly released, the data underlying the figures and tables for the Unitree G1, together with the G1 reference motion dataset will be released.} 

\section*{Supplementary materials}
\vspace{1ex}
{
\renewcommand{\arraystretch}{0.9}
\noindent
\begin{tabular}{@{}p{1.4cm}p{\dimexpr\linewidth-1.4cm\relax}@{}}
Section S1. & Nomenclature \\
Section S2. & Motion Dataset Details \\
Section S3. & Reward Function for Teacher Policy Training \\
Section S4. & Reward Function for Multi-Task RL Fine-tuning \\
Section S5. & Other Training Details \\
Section S6. & Tabula Rasa Policy \\
Section S7. & Autonomous Box Parkour\\
Figure S1.  & Extended Data Fidelity Analysis Results \\
Figure S2.  & Atlas D1 Quantitative Performance Analysis \\
Figure S3.  & Unitree G1 Quantitative Performance Analysis \\
Table S1.   & Extended Data Fidelity Analysis Metrics  \\
Table S2.   & Atlas D1 and Unitree G1 Sim-to-Real Analysis Results. \\
Table S3.   & Summary of Observation Terms. \\
Table S4.   & Termination Conditions. \\
Table S5.   & Domain Randomization Terms. \\
Table S6.   & Summary of Rewards. \\
Table S7.   & Unitree G1 Joint Gains and Action Specifications. \\
Table S8.   & Assistive Wrench Curriculum Parameters. \\
Table S9.   & Training Hyperparameters. \\
\end{tabular}
}




\bibliography{references}

\clearpage
\newpage

\setcounter{table}{0}
\makeatletter 
\renewcommand{\thetable}{S\@arabic\c@table}
\makeatother

\setcounter{figure}{0}
\makeatletter 
\renewcommand{\thefigure}{S\@arabic\c@figure}
\makeatother

\setcounter{algorithm}{0}
\makeatletter 
\renewcommand{\thealgorithm}{S\@arabic\c@algorithm}
\makeatother

\renewcommand{\thefigure}{S\arabic{figure}}
\renewcommand{\thetable}{S\arabic{table}}
\newcommand{\btau}{\boldsymbol{\tau}}
\newcommand{\bGamma}{\boldsymbol{\Gamma}}
\renewcommand{\v}{\mathbf{v}}
\renewcommand{\S}{\mathbf{S}}
\newcommand{\hli}[1]{{\color{blue}#1}}

\clearpage
\onecolumn
{\Huge \bfseries Supplementary Material\par}
\subsection*{Section S1. Nomenclature}
\label{sec:supplementary_nomenclature}
\begin{table}[H]
\centering
\small
\setlength{\tabcolsep}{6pt}

\begin{tabular}{ c l }
 $\mathcal{I}$                  & Inertial (world) frame \\
 $\mathcal{B}$                  & Torso (floating base) frame \\
 $\mathcal{T}$                  & IMU frame \\
 $\mathcal{E}_{k}$              & $k$-th end-effector frame \\
 $\numimu$                      & Number of IMUs \\
 $\numjoint$                    & Number of joints \\
 $\numee$                       & Number of end effectors \\
 $m$                            & Mass  \\
 $\bm I$                        & Inertia matrix \\
 $\piteacher$                   & Teacher policy \\
 $\vteacher$                    & Value function associated with the teacher policy \\
 $\pistudent$                   & Prior policy \\
 $\pideploy$                    & Deployment policy \\
 $\vdeploy$                     & Value function associated with the deployment policy \\
 ${}_{\mathcal{F}}{(\cdot)}$    & Vector expressed in coordinate frame $\mathcal{F}$  \\ 
 ${(\cdot)}^x, {(\cdot)}^y, {(\cdot)}^z$                        & $x$, $y$, $z$ Cartesian components of a 3D vector \\ 
 $\dot{(\cdot)}$                & First derivative  \\ 
 $\hat{(\cdot)}$                & Reference (desired) value  \\
 $\position_{\mathcal{FG}}$     & Position of coordinate frame $\mathcal{G}$ with respect to $\mathcal{F}$ \\
 $\Phi_{\mathcal{FG}}$          & Orientation of coordinate frame $\mathcal{G}$ with respect to $\mathcal{F}$ \\
 $(\phi,\ \theta,\ \psi)$       & Roll, pitch, yaw angles  \\
 $\gravity$                     & Gravity vector  \\
 $\linvel_{\mathcal{FG}}$       & Linear velocity of coordinate frame $\mathcal{G}$ with respect to $\mathcal{F}$ \\
 $\angvel_{\mathcal{FG}}$       & Angular velocity of coordinate frame $\mathcal{G}$ with respect to $\mathcal{F}$ \\
 $(v_{\text{long}},\ v_{\text{lat}},\ \omega_{\text{turn}})$    & SE(2) body velocity (longitudinal velocity, lateral velocity, and turning rate) \\
 $\jerk_{\mathcal{FG}}$         & Jerk (time-derivative of acceleration) of coordinate frame $\mathcal{G}$ with respect to $\mathcal{F}$ \\
 $\jointpos$                    & Joint positions  \\  
 $\jointtorque$                 & Joint torque     \\
 ${\bm f}$                      & Force   \\
 $\bm \tau$                     & Torque   \\
 ${\bm w}_e$                    & End-effector wrench (force and torque) \\
 $\action_t$                    & Action at time $t$ \\
 $\similarity$                  & Similarity score \\
 $\bar{L}_{\mathrm{episode}}$   & Episode length limit \\
 $\envref$                      & Set of reference-guided environments \\
 $\envgoal$                     & Set of goal-conditioned environments \\
 $\mathcal{U}(a, b)$            & Uniform distribution between $a$ and $b$ \\
 $\mathcal{N}(\mu, \sigma)$     & Gaussian distribution with mean $\mu$ and standard deviation $\sigma$ \\
 $\boxminus$                    & Lie-group difference on $SO(3)$ \\
 $\times$                       & Cross product operator \\
 $\oslash$                      & Element-wise division \\
\end{tabular}
\end{table}

\clearpage
\subsection*{Section S2. Motion Dataset Details}
\label{sec:supplementary_motion_dataset}

\subsubsection*{Dataset Composition and Choreography}

We collected an in-house motion dataset with choreography designed to cover the broad envelope of \setwo~torso velocity space that humans experience during natural locomotion on flat ground. 
Data was collected at five different speed categories: slow walking, normal walking, fast walking, jogging, and sprinting. Within each of these speed categories, various \emph{dance cards} were specified and reused, ensuring consistent coverage of \setwo~trajectories across speeds. These dance cards were selected to capture an overlapping set of turning radii and angles, as well as distinct human heading-control strategies, such as heading-aligned walking and locked heading, where an actor maintains a constant body orientation.

The primary structured tracking scripts include:
\begin{itemize}
    \item Discrete directional turns: Changes in heading executed across an eight-pointed star configuration ($0^{\circ}$, $\pm45^{\circ}$, $\pm90^{\circ}$, $\pm135^{\circ}$, and $180^{\circ}$).
    \item Continuous spline tracking: Spiral trajectories designed to sweep through a continuous range of curvature radii during constant-velocity locomotion.
    \item Linear maneuvers: Straight-line initialization, abrupt terminations, directional reversals, and lateral strafing.
    \item Zero-velocity pivots: Turning in place to execute heading changes at zero linear velocity, initiating and terminating at the same eight cardinal orientations.
\end{itemize}


Throughout the development of the \methodtitle~framework, additional data were collected iteratively to address observed gaps in the policy's capabilities due to data sparsity. For example, if sidestepping performance was sub-optimal, additional sidestepping data were collected to provide more examples.

In addition to these structured takes, unstructured data were collected to capture natural transitions between speeds, turns, stopping, and starting that are not inherently present in the structured scripts. These transitions were elicited by generating a randomized sequence of prompt commands (e.g. start, stop, turn, walk fast, slow, etc.) that were vocalized to the actor at randomized intervals. This protocol ensured high data entropy and prevented bias from the actor's subconscious movement preferences during transition. A variety of these takes were gathered under different constraints, including restricted headings, speed limits, and directional boundaries (such as straight-line locomotion versus unconstrained turning). While not exhaustive, this dataset provided broad coverage of trajectories representative of standard joystick-based steering.

The captured behaviors were partially constrained by the dimensions of the motion capture volume. To maximize spatial efficiency within these boundaries, choreography was optimized by separating left and right turns into distinct takes to enable larger turning radii, and by clearly marking the volume perimeter so the actor could fully utilize the available capture area.

Beyond standard locomotion, additional data were captured to cover tasks such as stepping onto boxes of various sizes, box manipulation, reaching for goals while constrained to a small footprint, push recoveries, and exaggerated walk style variations. Although these motions are not utilized for \methodtitle, \highlight{we plan to release the dataset to the public} to support research on more diverse behaviors.

\subsubsection*{Post-Process}


All trajectories are mirrored across the sagittal plane to enforce bilateral symmetry, providing isotropic coverage of left and right movements while simultaneously doubling the effective size of the dataset. Additionally, due to the inherent sensor noise and tracking artifacts occasionally present in optical motion capture systems, foot bones can exhibit slight misalignments, making the precise timing of ground contact events ambiguous.
To address this issue, the ground plane and the corresponding sole positions are estimated based on the kinematic configuration and foot height relative to the coordinate origin during periods when the actor is empirically determined to be stationary. These estimated contact-aligned bone segments are appended directly to the dataset's BVH files.

\subsubsection*{Temporal Adjustment}


The average height of the motion capture actor was \SI{178}{\centi\metre}. While Boston Dynamics Atlas robots roughly match these dimensions, many modern commercial humanoid platforms, including Unitree G1, are designed at approximately a $0.8\times$ geometric scale relative to the actor. To facilitate immediate deployment on smaller humanoids, a pre-scaled variant of the dataset is provided alongside the raw sequences. Under this scaling, all linear metrics (including positions, offsets, and segment dimensions) are uniformly scaled such that \SI{1}{\metre} in the original data maps to \SI{0.8}{\metre} in the scaled variant.
However, spatial scaling alone can compromise the physical realism of the motions, as it implicitly alters the effective gravitational acceleration governing the system dynamics. Naively retargeting this spatially scaled data degrades tracking fidelity during the subsequent policy training phase in physical simulation; this drawback is particularly pronounced in motions featuring frequent flight phases such as sprinting. As a straightforward countermeasure, we temporally warp the motion trajectories by scaling the original frame rate by a factor of $1/\sqrt{0.8}$. This temporal scaling preserves the nominal value of gravitational acceleration $g$ across the scaled kinematics. To illustrate this rationale, consider a simplified free-fall scenario where the vertical trajectory is governed by the equation of motion, $z = \frac{1}{2} g t^{2}$. When spatial scaling is applied, the modified vertical trajectory becomes $(0.8)z  = \frac{1}{2} g t'^{2}$ where $t'$ is the scaled motion duration. Substituting the original expression yields $(0.8)g t^{2}= g t'^{2}$, which results in the required temporal warp, $t' = \sqrt{0.8}t$.
\clearpage
\subsection*{Section S3. Reward Function for Teacher Policy Training}
\label{sec:supplementary_teacher_reward}

The total reward function for teacher training is defined as $r_{\text{teacher}} \;=\; \rtrack + \rreg + \rsurvival$, where the components represent the reference tracking reward, regularization reward, and survival reward, respectively. The following subsections describe each reward term in more detail and outline their mathematical definitions. For the hyperparameters including the weighting coefficients associated with each term, please refer to the summary provided in \highlight{\Cref{tab:supplementary_reward_summary}}. 

\subsubsection*{Reference Tracking Term}
The reference tracking term, $\rtrack$, is a weighted sum of sub-terms corresponding to state variable $i$, including torso linear velocity, torso angular velocity, torso orientation, joint positions. Each sub-term utilizes a structure that combines an exponential kernel with a quadratic penalty to penalize large deviations:
\begin{equation}
    \rtrack 
    \;=\; 
    \sum_{i} c_{t,i} \biggl( \kappa_t \cdot \exp\!\Bigl(-\, \frac{\|\einormalized\|}{\lambda_t} \Bigr) - \|\einormalized\|^2 \biggr).
\end{equation}
In this formulation, $c_{t_i}$ denotes the weight for each state variable $i$, and $\einormalized$ represents the normalized error between the current state and its reference value.
We apply a uniform coefficient, $\kappa_t = 5.0$ and $\lambda_t=10.0$, across all sub-terms to regulate the contribution of the exponential component relative to the quadratic penalty. 
To ensure the reward remains robust across the diverse motion repertoire, we adapt the sensitivity of the exponential term and the scale of the error based on the specific motion trajectory, $k$, currently being tracked. The normalized error is computed as:
\begin{equation}
\label{eq:error_normalized}
    \einormalized = \ei \oslash \bm{\sigma}^k_i ,
\end{equation}
where $\ei$ is the raw state error, $\bm{\sigma}^k_i$ is the pre-calculated standard deviation of state variable $i$ over trajectory $k$, and $\oslash$ denotes element-wise division. To prevent numerical instability and excessively high rewards for quasi-static motions, these standard deviations are clipped by a user-specified lower limit, $\sigma_{i,\text{min}}$. This normalization strategy ensures that the reward scale remains uniform across various physical units and motion styles, facilitating stable learning across the entire dataset.

The subterms are computed based on the following tracking error definitions:
\begin{itemize}
    \item Reference torso position tracking $r_{t,\position}$ : $\bm e_{\position} = {}_{\mathcal{I}} \hat{\position}_{\mathcal{IB}} - {}_{\mathcal{I}} {\position}_{\mathcal{IB}}$
    \item Reference torso orientation tracking $r_{t,\Phi}$ : $\bm e_{\Phi} = {}_\mathcal{I}\hat{\bm{\Phi}}_{\mathcal{IB}}\boxminus {}_\mathcal{I}\bm{\Phi}_{\mathcal{IB}}$
    \item Reference torso linear velocity tracking $r_{t,\linvel}$ : $\bm e_{\linvel} = {}_{\mathcal{B}} \hat{\linvel}_{\mathcal{IB}} - {}_{\mathcal{B}} {\linvel}_{\mathcal{IB}}$
    \item Reference torso angular velocity tracking $r_{t,\angvel}$: $\bm e_{\angvel} = {}_{\mathcal{B}} \hat{\angvel}_{\mathcal{IB}} - {}_{\mathcal{B}} {\angvel}_{\mathcal{IB}}$
    \item Reference torso projected gravity tracking $r_{t,{\bm g}}$: $\bm e_{\bm g} = {}_{\mathcal{B}} \gravityref - {}_{\mathcal{B}} \gravity$
    \item Reference joint position tracking $r_{t,j}$: $\bm e_j = {\jointposref} - \jointpos$ 
\end{itemize}

\subsubsection*{Regularization Term} 
The regularization term, $r_{\text{reg}}$, promotes smooth and physically feasible behaviors by penalizing high-frequency oscillations, excessive physical effort, and regularizing motion to alleviate artifacts in motion data. The term is a weighted sum of the following subterms, with the specified weight coefficients:
\begin{equation}
    \rreg 
    \;=\; 
    \sum_{i} c_{\text{reg},i} \cdot r_{\text{reg},i}.
\end{equation}

The subterms are listed as follows:
\begin{itemize}
    \item Action smoothness penalty $r_{\text{reg},{\Delta \bm a}}$: this term penalizes large action rates to promote smooth motion, defined as $r_{\text{reg},{\Delta \bm a}} = \|{\bm a_{t}} - {\bm a_{t-1}}\|_2$
    \item Joint torque penalty $r_{\text{reg},{\bm{\tau}_j}}$: this term penalizes excessive torque exerted by the joint-level PD controller, defined as $r_{\text{reg},{\bm{\tau}_j}} = \|\jointtorque\|_2$
    \item Feet slide penalty $r_{\text{reg},{\text{slide}}}$: this term penalizes feet slide in feet contacted with ground. Given that ${}_{\mathcal{I}} \bm{v}_{\mathcal{IE}_k}$ represents the velocity of the $k$-th contact foot with respect to the inertial frame (expressed in the inertial frame), the penalty is defined as $r_{\text{reg},{\text{slide}}} = \sum_{k\, \in \,\text{contact}} \|{}_{\mathcal{I}} \bm{v}_{\mathcal{IE}_k}\|_2.$
    \item Feet jerk penalty $r_{\text{reg},{\jerk}}$: this term penalizes foot jerk---the time derivative of foot acceleration. Given the jerk of the $k$-th foot with respect to the inertial frame (expressed in the inertial frame) as ${}_{\mathcal{I}} \jerk_{\mathcal{IE}_k}$, the penalty is defined as $r_{\text{reg},{\jerk}} = \sum_{k\, \in \,\text{feet}} \|{}_{\mathcal{I}} \jerk_{\mathcal{IE}_k}\|_2$.
    \item Feet clearance reward $r_{\text{reg},{\text{clr}}}$: this term promotes clean leg swings, preventing foot scuffing that could compromise overall stability. For $k$-feet, we compute the foot clearance error $e_{\text{clr},k} = \max(0, {}_\mathcal{I}\hat\position^z_{\text{clr}} - {}_\mathcal{I}\position^z_{\mathcal{IE}_k}) \in \mathbb{R}^{+}$, where ${}_\mathcal{I}\hat\position^z_{\text{clr}}$ is the nominal feet swing height set as a hyperparameter. By clipping the error at zero, we ensure that the policy is only penalized for insufficient clearance and not for higher foot swings. 
    The squared error is then weighted by the hyperbolic tangent of the norm of the corresponding reference $xy$-velocity of the foot. This scaling mechanism amplifies the penalty when the reference foot velocity is high---typically at the peak of the swing trajectory---while minimizing its influence when the velocity is low, such as near liftoff or touchdown. The final reward is formulated as:
    \begin{equation}
        r_{\text{reg},{\text{clr}}} = \sum_{k\in\text{feet}} 
        \exp\Big(- 
        \tanh \big(\lambda_{\text{clr},v} \cdot \| {}_\mathcal{I}\hat\linvel^{xy}_{\mathcal{IE}_k} \| \big)
        \cdot
        \frac{e^2_{\text{clr},k}}{\lambda_{\text{clr},e}} 
        \Big).
    \end{equation}
    Unlike the other regularization terms that carry over to Stage II, this term is excluded during the goal-conditioned learning phase.
\end{itemize}

\subsubsection*{Survival term}
The survival reward, $\rsurvival = c_{\text{survival}}$, is a constant positive term granted at each time step discouraging premature termination by incentivizing longer episodes and ensures a positive total reward offset. 

\clearpage
\subsection*{Section S4. Reward Function for Multi-Task RL Fine-tuning}
\label{sec:supplementary_multirl_reward}

The reward function for the multi-task RL fine-tuning is defined as 
\begin{equation}
    \rfinetunestructure
\end{equation}
where the reward for the reference-guided environment is $\rref = \rteacher$. Meantime, for the goal-conditioned task, the function is defined as $\rgoal = \rcommand + \rreg + \rsurvival$, where the components represent the command tracking reward, regularization reward, and survival reward, respectively. The regularization term $\rreg$ and survival bonus $\rsurvival$ are identical to those used in the teacher reward function. 

\subsubsection*{Command Tracking Term}

The command tracking term aligns the robot's $\setwo$ velocity with the given steering command. It is computed as a weighted sum of the linear velocity and turn rate tracking components using an exponential kernel:
\begin{equation}
    \rcommand = \sum_i c_{\text{cmd},i} \cdot \exp\Bigl(- \frac{\|\bm e_i\|^2}{\lambda_{\text{cmd},i}} \Bigr)
\end{equation}

The command tracking errors are defined as follows:
\begin{itemize}
    \item Linear velocity command tracking $r_{\text{cmd}, v}$: $\bm e_{v} = \big(\hat{v}_{\text{long}},\ \hat{v}_{\text{lat}}\big) - \big({v}_{\text{long}},\ {v}_{\text{lat}}\big) \in \mathbb{R}^2$
    \item Turn rate command tracking $r_{\text{cmd}, \omega}$: $\bm e_{\text{turn}} = \hat{\omega}_{\text{turn}} - {\omega}_{\text{turn}} \in \mathbb{R}$
\end{itemize}

The values of the weighting coefficients $c_{\text{cmd},i}$ and the exponential scale factor $\lambda_{\text{cmd},i}$ for each subterms are listed in \highlight{\Cref{tab:supplementary_reward_summary}}. 
\subsection*{Section S5. Other Training Details}

This section provides additional training details and implementation specifics.

\subsubsection*{Early Termination}

To improve sample efficiency, we employ early termination across all training stages within the pipeline. An episode is early terminated if ground reaction forces exceed a predefined threshold, a self-collision occurs, or the agent deviates significantly from the reference torso height and orientation. A comprehensive summary of the termination criteria is provided in \highlight{\Cref{tab:supplementary_termination}}. 

\subsubsection*{Assistive Wrench Curriculum in Teacher Training}

To accelerate teacher policy training and ensure the policy internalizes the diverse motion repertoire without skill-specific reward tuning, we employ an assistive wrench curriculum. Drawing inspiration from \emph{ZEST}, this curriculum applies an external assistive force and torque to the robot's torso. These are computed via a model-based PD control law combined with a feedforward component that compensates for nominal torso dynamics:
\begin{subequations}
\label{eq:assist}
\begin{align}
\bm{f}_{\text{assist}} \;&=\; 
m\!\left({}_{\mathcal{I}}\hat{\dot{\linvel}}_{\mathcal{IB}} 
\;+\; k_p^v\,({}_{\mathcal{I}}\hat{\position}_{\mathcal{IB}} - {}_{\mathcal{I}}\position_{\mathcal{IB}})  
\;+\; k_d^v\,({}_{\mathcal{I}}\hat{\linvel}_{\mathcal{IB}} - {}_{\mathcal{I}}\linvel_{\mathcal{IB}}) 
\;-\; {}_{\mathcal{I}}\gravity\right), \\
\bm{\tau}_{\text{assist}} \;&=\; 
{}_{\mathcal{I}}\bm{I}\,{}_{\mathcal{I}}\hat{\dot{\angvel}}_{\mathcal{IB}}
\;+\; k_p^\omega\, {}_{\mathcal{I}}\bm{I}\,\bigl({}_\mathcal{I}\hat{\bm{\Phi}}_{\mathcal{IB}}\boxminus {}_\mathcal{I}\bm{\Phi}_{\mathcal{IB}}\bigr)
\;+\; k_d^\omega\, {}_{\mathcal{I}}\bm{I}\,({}_{\mathcal{I}}\hat{\angvel}_{\mathcal{IB}}-{}_{\mathcal{I}}\angvel_{\mathcal{IB}}) \nonumber \\
& \quad\qquad\qquad \;+\; {}_{\mathcal{I}}\angvel_{\mathcal{IB}} \times({}_{\mathcal{I}}\bm{I}{}_{\mathcal{I}}\angvel_{\mathcal{IB}})
\;-\; {}_{\mathcal{I}}\position_{\mathcal{B},\mathrm{com}}\times m{}_{\mathcal{I}}\gravity.
\end{align}
\end{subequations}
In this formulation, we approximate the system as a single rigid body with total mass $m$ and a Composite Rigid Body (CRB) inertia matrix ${}_{\mathcal{B}}\bm{I}$, calculated in a nominal configuration relative to the torso frame $\{\mathcal{B}\}$. The matrix ${}_{\mathcal{I}}\bm{I}$ denotes the inertia matrix expressed in the inertial frame $\{\mathcal{I}\}$, and ${}_{\mathcal{I}}\bm{r}_{\mathcal{B},\text{com}}$ denotes the position of the center of mass (CoM) relative to the torso frame. The operator $\boxminus$ denotes the Lie-group difference on $SO(3)$, implemented as the rotation-log map for quaternions to produce an error vector in the tangent space.

We scale the assistive force and torque by a curriculum gain ${\beta\!\in\!(0,\beta_{\text{max}})}$ with ${\beta_{\text{max}} < 1.0}$. 
The curriculum is adaptively modulated based on the agent's performance, measured by an episodic similarity metric:
\begin{equation}
     \similarity = \frac{1}{\bar{L}_{\mathrm{episode}}} \sum_{t=0}^{\bar{L}_{\mathrm{episode}}} \exp\!\Bigl(-\, \frac{\|\tilde{\bm e}_{j, t}\|}{\lambda_s} \Bigr)
\end{equation}
where $L_{\text{episode}}$ is the maximum episode length, and the coefficients are set to $\lambda_s = 10.0$. The term $\tilde{\bm e}_{j, t}$ represents the normalized joint position error at time $t$, defined in \Cref{eq:error_normalized}. The resulting similarity metric $\similarity \in [0, 1]$ serves as a proxy for how accurately the agent tracks the reference motion manifold.
Upon episode termination, we update the curriculum gain $\beta$ stochastically. The gain is increased by $\Delta\beta_\uparrow$ with a probability equal to the similarity score $\similarity$, and decreased by $\Delta\beta_\downarrow$ with a probability of $1 - \similarity$. 
This probabilistic update mechanism ensures that the assistive wrench is phased out only as the policy demonstrates consistent tracking proficiency. By capping the gain at $\beta_{\text{max}}$, we maintain the assistance as only partial, encouraging meaningful exploration of the robot's self-sustained dynamics and preventing the policy from overfitting to the modified physics of the assisted environment.
The parameter values utilized for the assistive wrench curriculum in our experiments are summarized in \highlight{\Cref{tab:curriculum}}.

\subsubsection*{Domain Randomization}

To facilitate zero-shot sim-to-real transfer and foster robust recovery behaviors, we employ domain randomization. This includes injecting Gaussian noise into the actor's proprioceptive observations, applying random impulsive perturbations to the torso, and varying environmental parameters such as surface friction. Furthermore, we randomize the robot's physical properties, such as link masses, center-of-mass (CoM) offsets, joint friction coefficients, and joint-zero position offsets. The domain randomizations used can differ for various robots, for example, high stiction in the parallel linkage mechanism on the ankle of Atlas R1 resulted in the use of the joint friction randomization. 
Beyond bridging the reality gap, these randomizations expose the agent to a broader distribution of states during training. This prevents the policy from overfitting to the narrow state-space of the reference trajectories, thereby enhancing its capability to recover from significant deviations and environmental uncertainties. 
A comprehensive list of the domain randomization parameters and their respective ranges used across our experiments is summarized in \highlight{\Cref{tab:supplementary_domain_randomization}}.

We note that during initial hardware deployment of the final policy on the Unitree G1 robot, the emergent ankle behaviors exhibited excessive stiffness. While these behaviors were not problematic in simulation, they proved brittle in real-world environments; specifically, the lack of ankle compliance led to frequent stumbling when the foot failed to strike the ground at a nominal landing angle. To mitigate this, we introduced additional domain randomization in terrain roughness. By forcing the agent to traverse uneven surfaces, we successfully increased the diversity of the roll-out samples, exposing the policy to a much wider range of ankle angles and impact conditions.

Terrain roughness randomization is implemented by uniformly sampling random height offsets for each grid point at a resolution of \SI{0.1}{\metre}. The continuous terrain mesh is subsequently generated by interpolating these grid points. The specific ranges for these terrain offsets used in our G1 policy train are detailed in \highlight{\Cref{tab:supplementary_domain_randomization}}. 

To prevent domain randomization from compromising the stylistic fidelity of the learned behaviors, training begins on a flat surface, allowing the teacher policy to prioritize the acquisition of high-fidelity stylistic nuances from the motion dataset without the interference of environmental noise. Then we introduce terrain randomization only after the midpoint of the teacher policy's training phase, maintaining it through the remainder of the optimization process.
Additionally, to streamline the policy training, we provide the terrain height observation $\oterrain$ to the critic network as privileged information. This observation samples terrain heights from a $17 \times 11$ grid centered at the robot with \SI{0.1}{\metre} spacing between sample points. The resulting heightmap is then flattened into a vector before being provided to the critic.




\subsection*{Section S6. Tabula Rasa Policy}
\label{sec:tabula_rasa}

The \tabularasa~RL policy used for benchmarking shares the same observation and action spaces as the final deployable \methodtitle~policy. It maps proprioceptive observations and \setwo~velocity commands directly to joint actions, generating locomotion behaviors that track the target velocity. Unlike our proposed method, these baseline policies are trained without prior human motion data, relying on emergent locomotion behaviors shaped purely by hand-crafted reward functions. 
In this section, we provide a high-level overview of the training and implementation details for the \tabularasa~RL policy as implemented on the Unitree G1 robot.

Similar to \methodtitle~training setup, the G1 \tabularasa~training also employs an Asymmetric Actor-Critic architecture, where the critic network is provided with additional privileged observation to accelerate the training. These privileged observations include the robot's base height, the positions and velocities of the feet relative to the base frame, and the feet wrenches resulting from joint torques. Additionally, we introduce a binary command activation indicator, which is set to zero when the velocity commands are zero, and set to one otherwise. This indicator helps train the policy to stand still rather than step in place when the velocity command is zero.

At a high-level, the reward function of \tabularasa~policy is structured as $r_{\text{tabula rasa}} = \rcommand + \rreg + \rsurvival$ where the components represent the command tracking reward, regularization reward and survival reward respectively. Among these terms, survival term is defined as a constant positive value, similar to the \methodtitle~RL training rewards, and the command tracking reward is designed to be identical to that of the goal-conditioned task during \methodtitle~fine-tuning stage. Both the magnitude of the survival term and the tracking subterm weights are adjusted to align with the landscape of the overall \tabularasa~reward function.

The regularization reward of \tabularasa, by contrast, is considerably more complex than its \methodtitle~counterpart, as it is solely responsible for shaping the behavior of the policy.
These regularization terms include penalties on torso vertical, roll, and pitch movements, as well as deviations from an upright posture. At the joint level, motion is regularized by penalizing deviations from a nominal configuration, alongside soft constraints on joint positions to prevent action-induced joint limit violations. To encourage smooth stepping patterns while preventing stomping or stance-foot slippage, penalties are imposed on foot acceleration and jerk, coupled with a foot-slide penalty. Furthermore, the yaw of the feet relative to the torso frame is restricted to prevent excessive twisting, which is complemented by a term encouraging the feet to remain flat relative to the ground. Joint torques and joint accelerations are also penalized to prevent excessive actuator stress, mitigate hardware damage, and reduce energy consumption. Lastly, to encourage a natural leg swing, two additional sub-terms incentivize longer swing foot air time and a foot clearance profile targeting a maximum swing height of \SI{0.15}{\metre}.


Because the \tabularasa~policy does not utilize a motion dataset, we omit the assistive wrench curriculum, which depends on tracking deviations from a reference trajectory. Instead, we implement a curriculum that progressively expands the velocity command sampling range. This adaptive curriculum broadens the command space when an environment exhibits high tracking fidelity to previously sampled velocity commands, and contracts the range otherwise. For the same reason, episode initialization is designed to reset the robot to its nominal joint configuration with randomized joint position deviations sampled from $\mathcal{U}(-0.2, 0.2)\;\SI{}{\radian}$ and joint velocity deviations from $\mathcal{U}(-0.2, 0.2)\;\SI[per-mode=repeated-symbol]{}{\radian\per\second}$.

The domain randomization settings for the \tabularasa~policy training are identical to those maintained across the entire \methodtitle~training pipeline. The termination conditions match the goal-conditioned task during \methodtitle~fine-tuning stage, while the PPO hyperparameters are set to the same values used during the \methodtitle~teacher policy training.

\subsection*{Section S7. Autonomous Box Parkour}
\label{sec:supplementary_box_parkour}

In this section, we present the implementation details of the autonomous box parkour pipeline, showcased as an extended use case of the \methodtitle~policy. The objective of this pipeline is to enable the robot to navigate box parkour environments, including unseen configurations, toward a defined \setwo~goal pose while producing natural motions.
The task requires the robot to perform three behaviors: walking, climbing up, and climbing down. To achieve this, we adopt a perceptive multi-task hierarchical RL framework utilizing a mixture-of-experts technique. 
In this setup, a high-level policy orchestrates separately learned motor behaviors to guide the robot through the environment toward its goal. 

The perceptive high-level policy outputs, at \SI{10}{\hertz}, \setwo~velocity commands $(v_{\text{long}},\ v_{\text{lat}},\ \omega_{\text{turn}})$ and a set of gating weights over three frozen, pretrained low-level expert policies: a \methodtitle~policy, a box climb-up policy, and a box climb-down policy. The experts each run at \SI{50}{\hertz} and output joint position targets. The \setwo~velocity command is low-pass filtered to prevent abrupt changes and is provided to each expert in its required representation: the \methodtitle~policy receives the velocity command directly, whereas the climb-up and climb-down experts receive an \setwo~pose waypoint obtained by integrating the velocity command. Each expert's outputs are then combined as a weighted average, using gating weights normalized to sum to one, to yield the final action. Note that only the high-level policy is perceptive, while the low-level experts are non-perceptive.

The Unitree G1 robot successfully performs a variety of long-horizon navigation tasks in environments containing one and two boxes, deployed zero-shot on hardware. The robot transitions smoothly between walking, climbing up, and climbing down, chaining these behaviors autonomously, with the high-level policy selecting and blending experts on the fly.

\subsubsection*{Low-level Policies Training Details}

The \methodtitle~policy is trained as described in the main manuscript. The box climb-up and climb-down policies (collectively, the box-climbing policies) are trained via reference guided imitation with PPO. While both the box-climbing policies and the \methodtitle~teacher policy take inspiration from \emph{ZEST}, their implementations differ as follows.
First, in the box-climbing policies the command is defined as an \setwo~pose waypoint rather than an \setwo~velocity. At every timestep, the reference motions and their mirrored counterparts define a waypoint at a \SI{0.5}{\second} look-ahead horizon, representing the desired future pose expressed in the robot's torso heading frame. Gaussian observation noise is added to the waypoint commands to account for inaccuracies in the high-level planner. 
Second, the full-body reference states are not provided as inputs to the box-climbing policies. 
This omission allows these policies to be directly deployed without distillation, and it also enables them to generalize to unseen conditions. Third, compared to the \methodtitle~teacher policy, the critic observations include different privileged information such as tracking scores and a task phase variable that increments from zero to one along the reference motion trajectory.
Otherwise, the reward functions, training curriculum, domain randomization schemes, and PPO hyperparameters used in the box-climbing policies closely follow the \methodtitle~teacher training, with only minor differences.

\subsubsection*{High-level Policy Training Details}

The high-level policy is trained using PPO in two stages. The first stage focuses exclusively on primitive tasks, walking, climb-up and climb-down, with each task sampled with equal probability. The second stage additionally introduces three compound tasks: walking to climb-up, climb-up to climb-down, and climb-down to walking. These compound tasks are sampled with half the probability of the primitive tasks.

The policy observes proprioceptive information, the \setwo~goal pose command, and a local height scan. The critic receives additional privileged information, including the torso height and---specifically for box-climbing motions---the torso height error relative to the reference motion. 
All tasks incorporate regularization and goal tracking rewards, whereas the box-climbing tasks additionally include rewards based on reference torso height that encourage stylistic fidelity to the reference motions. 

%
\begin{figure*}[t!]
\centering
\ifdefined\isArXiv
%
%
\begin{subfigure}[b]{0.48\textwidth}
    \centering
    \includegraphics[width=\textwidth]{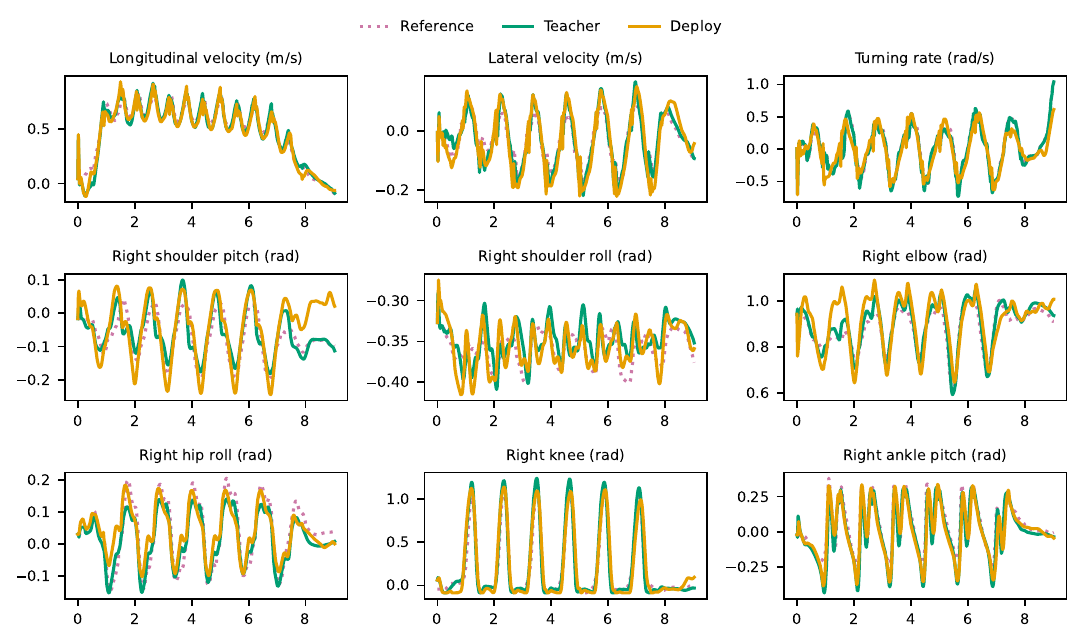}    
    \caption{\emph{Slow Forward Walk.}}
\end{subfigure}
\begin{subfigure}[b]{0.48\textwidth}
    \centering
    \includegraphics[width=\textwidth]{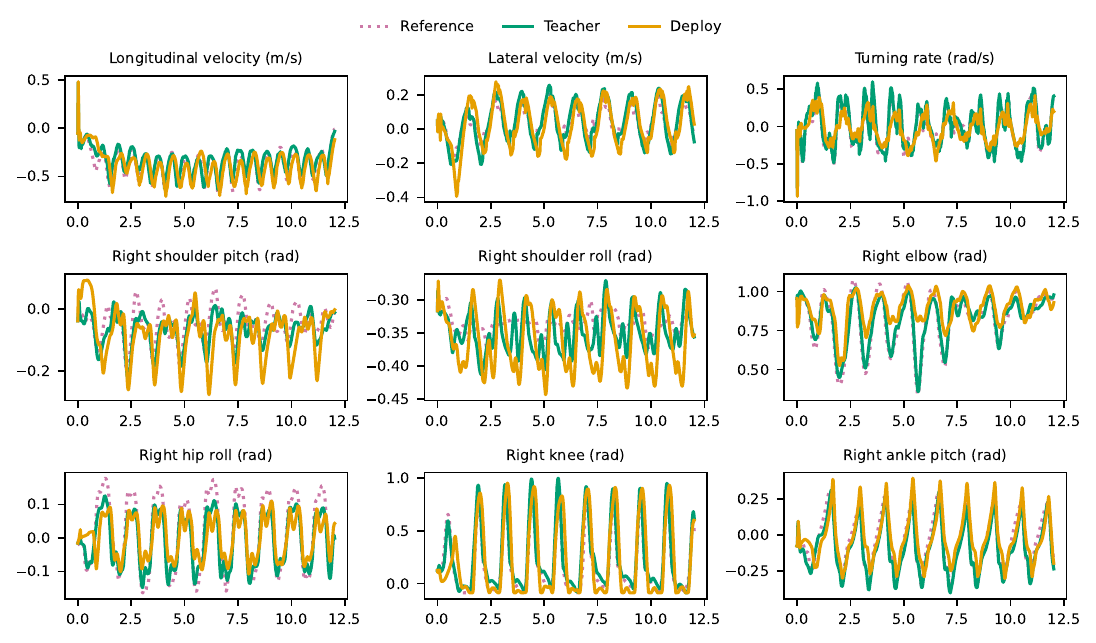}    
    \caption{\emph{Slow Backward Walk.}}
\end{subfigure}
\begin{subfigure}[b]{0.48\textwidth}
    \centering
    \includegraphics[width=\textwidth]{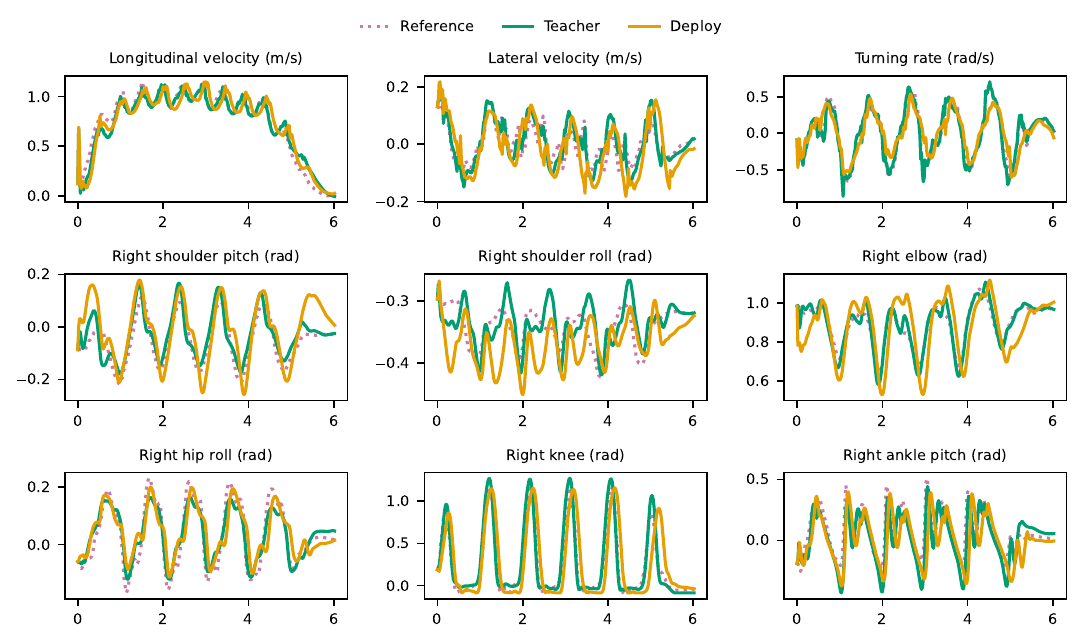}    
    \caption{\emph{Normal Forward Walk.}}
\end{subfigure}
\begin{subfigure}[b]{0.48\textwidth}
    \centering
    \includegraphics[width=\textwidth]{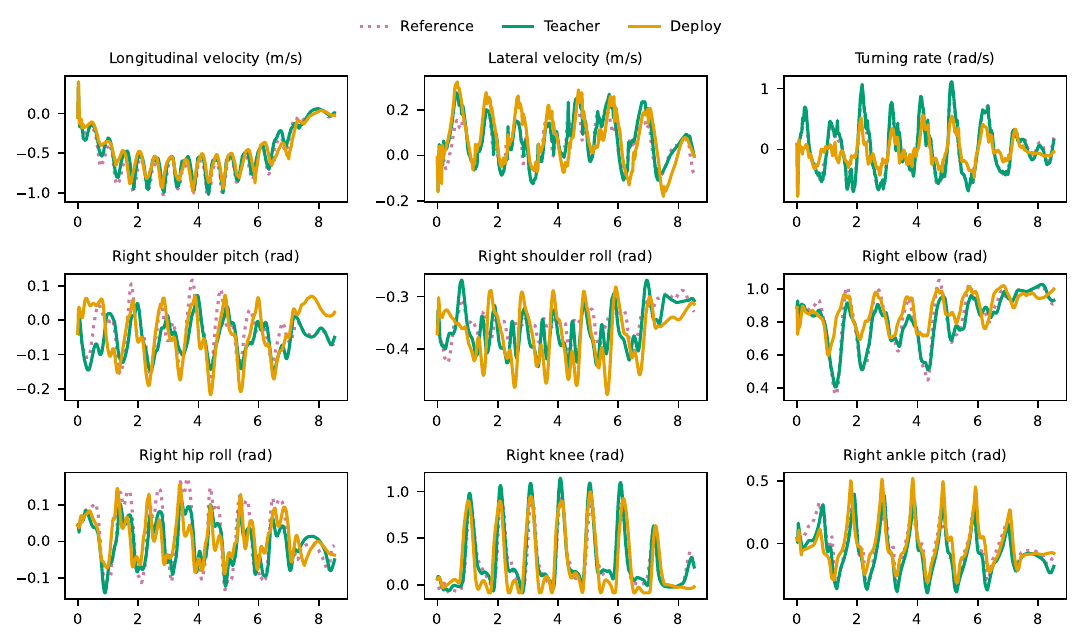}    
    \caption{\emph{Normal Backward Walk.}}
\end{subfigure}
\begin{subfigure}[b]{0.48\textwidth}
    \centering
    \includegraphics[width=\textwidth]{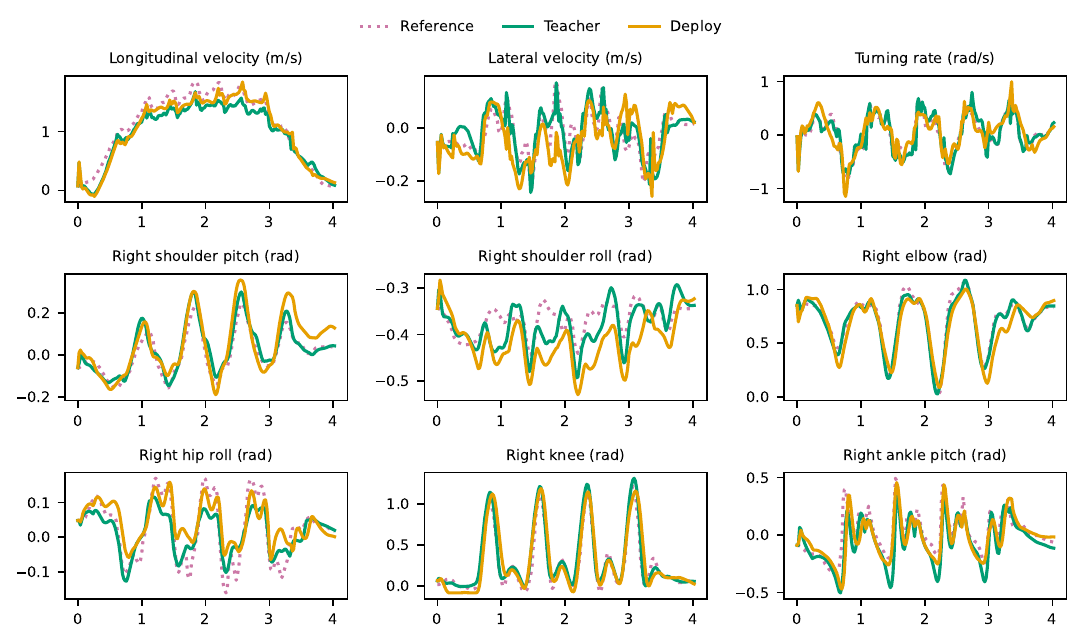}    
    \caption{\emph{Fast Forward Walk.}}
\end{subfigure}
\begin{subfigure}[b]{0.48\textwidth}
    \centering
    \includegraphics[width=\textwidth]{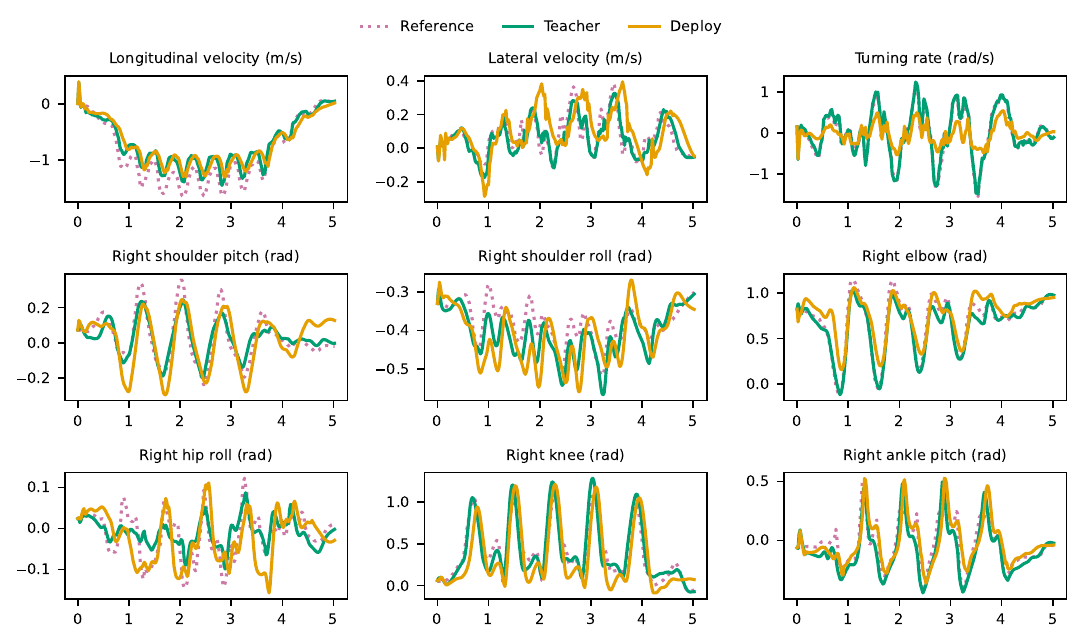}    
    \caption{\emph{Fast Backward Walk.}}
\end{subfigure}
\begin{subfigure}[b]{0.48\textwidth}
    \centering
    \includegraphics[width=\textwidth]{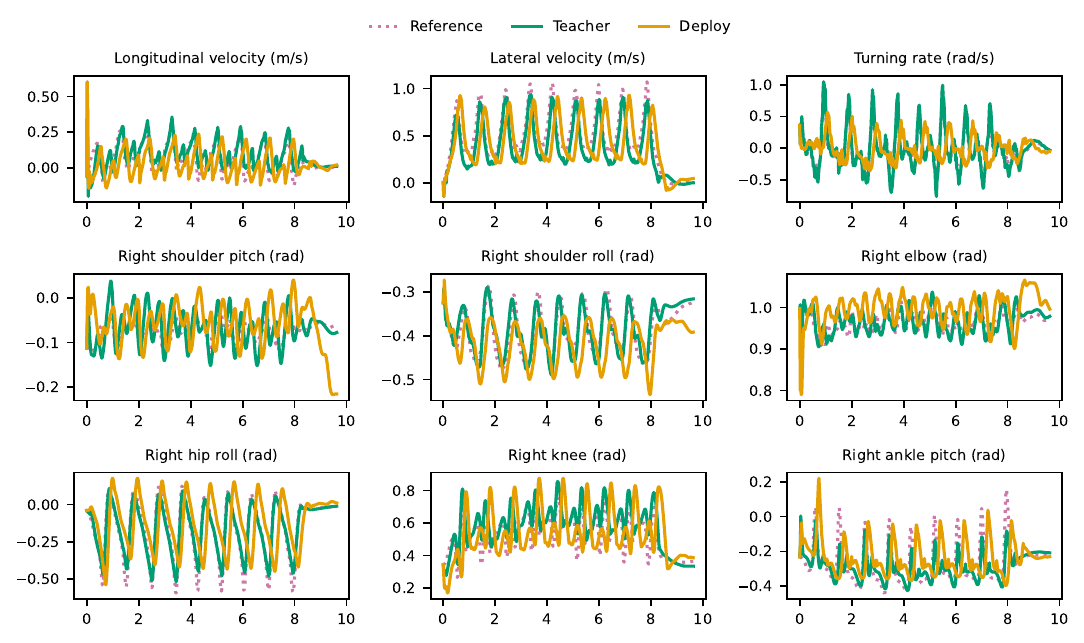}    
    \caption{\emph{Sidestep.}}
\end{subfigure}
\begin{subfigure}[b]{0.48\textwidth}
    \centering
    \includegraphics[width=\textwidth]{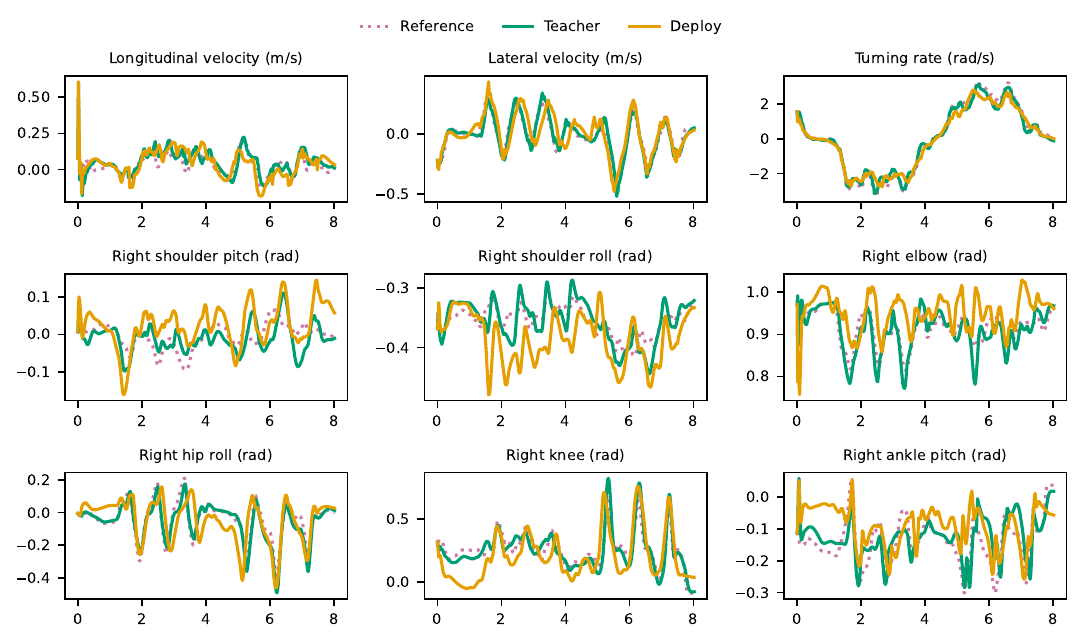}    
    \caption{\emph{Turn in Place.}}
\end{subfigure}
\else
%
%
\begin{subfigure}[b]{0.45\textwidth}
    \centering
    \includegraphics[width=\textwidth]{Images/Supplementary/Fidelity/Slow_Forward_Walk_dof.pdf}    
    \caption{\emph{Slow Forward Walk.}}
\end{subfigure}
\hspace{1em}
\begin{subfigure}[b]{0.45\textwidth}
    \centering
    \includegraphics[width=\textwidth]{Images/Supplementary/Fidelity/Slow_Backward_Walk_dof.pdf}    
    \caption{\emph{Slow Backward Walk.}}
\end{subfigure}
\begin{subfigure}[b]{0.45\textwidth}
    \centering
    \includegraphics[width=\textwidth]{Images/Supplementary/Fidelity/Normal_Forward_Walk_dof.pdf}    
    \caption{\emph{Normal Forward Walk.}}
\end{subfigure}
\hspace{1em}
\begin{subfigure}[b]{0.45\textwidth}
    \centering
    \includegraphics[width=\textwidth]{Images/Supplementary/Fidelity/Normal_Backward_Walk_dof.pdf}    
    \caption{\emph{Normal Backward Walk.}}
\end{subfigure}
\begin{subfigure}[b]{0.45\textwidth}
    \centering
    \includegraphics[width=\textwidth]{Images/Supplementary/Fidelity/Fast_Forward_Walk_dof.pdf}    
    \caption{\emph{Fast Forward Walk.}}
\end{subfigure}
\hspace{1em}
\begin{subfigure}[b]{0.45\textwidth}
    \centering
    \includegraphics[width=\textwidth]{Images/Supplementary/Fidelity/Fast_Backward_Walk_dof.pdf}    
    \caption{\emph{Fast Backward Walk.}}
\end{subfigure}
\begin{subfigure}[b]{0.45\textwidth}
    \centering
    \includegraphics[width=\textwidth]{Images/Supplementary/Fidelity/Side_Step_dof.pdf}    
    \caption{\emph{Sidestep.}}
\end{subfigure}
\hspace{1em}
\begin{subfigure}[b]{0.45\textwidth}
    \centering
    \includegraphics[width=\textwidth]{Images/Supplementary/Fidelity/Turn_in_Place_dof.pdf}    
    \caption{\emph{Turn in Place.}}
\end{subfigure}
\fi
\caption{
    \textbf{Extended Data Fidelity Analysis Results.} Comprehensive trajectory comparison of the raw $\text{SE}(2)$ torso velocities and joint angle trajectories for the simulated Unitree G1 robot across all eight canonical behaviors. 
}
\label{fig:supplementary_g1_data_fidelity}
\end{figure*}

\clearpage
%
%
\begin{figure*}[t!]
    \centering
    
    \begin{subfigure}[b]{\textwidth}
        \centering
        \includegraphics[width=\textwidth]{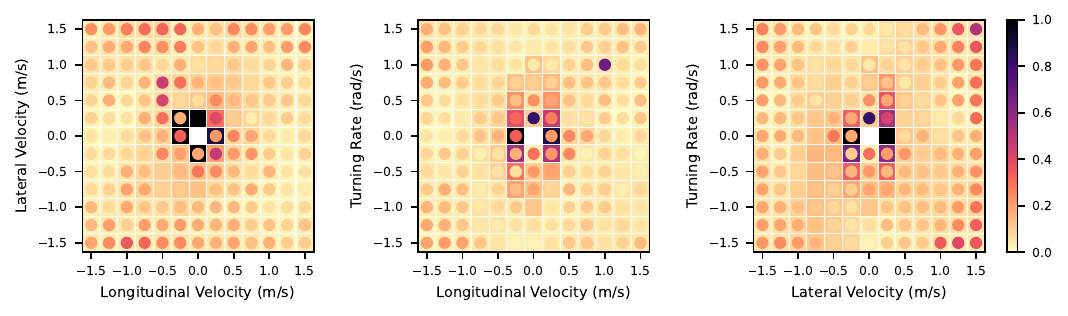}
        \vspace{-2em}
        \caption{Relative command tracking error in steady-state. From left to right: $\omega_{\text{turn}}=0$, $v_{\text{lat}}=0$, $v_{\text{long}}=0$.}
        \label{fig:AtlasD1_SteadyStateRelativeError}
    \end{subfigure}
    
    \vspace{1mm}

    \begin{subfigure}[b]{\textwidth}
        \centering
        \includegraphics[width=\textwidth]{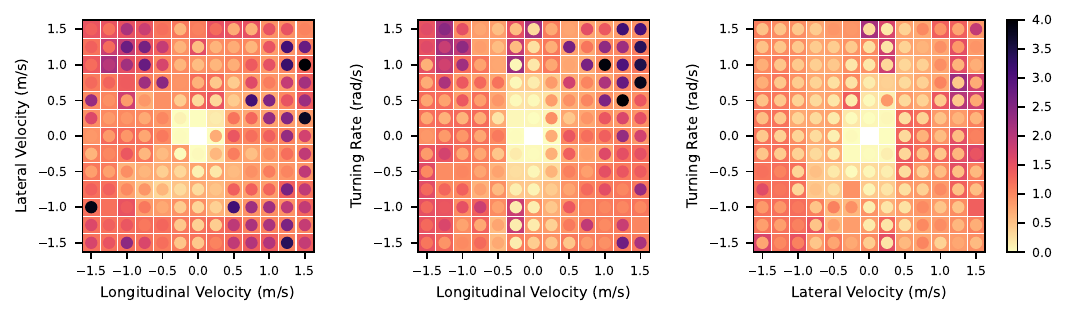}
        \vspace{-2em}
        \caption{Rise time. From left to right: $\omega_{\text{turn}}=0$, $v_{\text{lat}}=0$, $v_{\text{long}}=0$.}
        \label{fig:AtlasD1_RiseTime}
    \end{subfigure}   


    \vspace{1mm}

    \begin{subfigure}[b]{\textwidth}
        \centering
        \includegraphics[width=\textwidth]{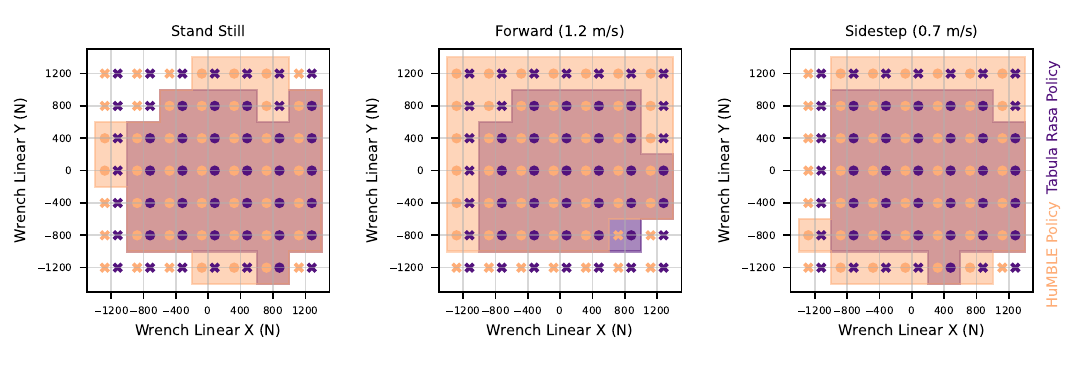}
        \vspace{-3em}
        \caption{Push recovery robustness.}
        \label{fig:AtlasD1_PushRecovery}
    \end{subfigure}

    \caption{
        \textbf{Atlas D1 quantitative performance analysis.} 
        We provide the same performance evaluations for the simulated Atlas D1 robot as presented for the Atlas R1 robot in \highlight{Figure~4}.
    }
    \label{fig:d1_quantitative_analysis}
\end{figure*}

%
%
\begin{figure*}[t!]
    \centering

    \begin{subfigure}[b]{\textwidth}
        \centering
        \includegraphics[width=\textwidth]{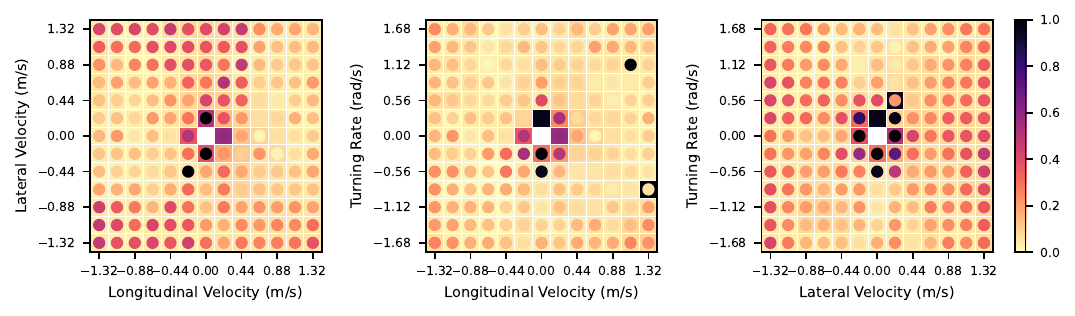}
        \vspace{-2em}
        \caption{Relative command tracking error in steady-state. From left to right: $\omega_{\text{turn}}=0$, $v_{\text{lat}}=0$, $v_{\text{long}}=0$.}
        \label{fig:G1_SteadyStateRelativeError}
    \end{subfigure}
    
    \vspace{1mm}

    \begin{subfigure}[b]{\textwidth}
        \centering
        \includegraphics[width=\textwidth]{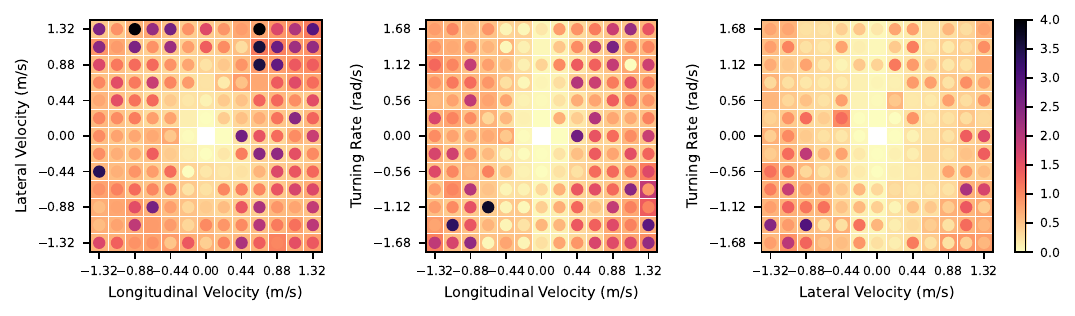}
        \vspace{-2em}
        \caption{Rise time. From left to right: $\omega_{\text{turn}}=0$, $v_{\text{lat}}=0$, $v_{\text{long}}=0$.}
        \label{fig:G1_RiseTime}
    \end{subfigure}   


    \vspace{1mm}

    \begin{subfigure}[b]{\textwidth}
        \centering
        \includegraphics[width=\textwidth]{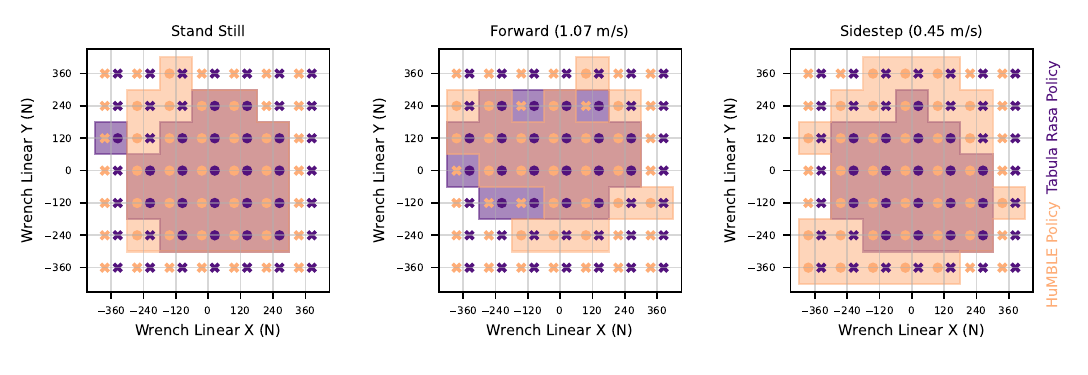}
        \vspace{-3em}
        \caption{Push recovery robustness. }
        \label{fig:G1_PushRecovery}
    \end{subfigure}

    \caption{
        \textbf{Unitree G1 quantitative performance analysis.}
        We provide the same performance evaluations for the simulated Unitree G1 robot as presented for the Atlas R1 robot in \highlight{Figure~4}. 
        Note that the G1 policy is evaluated under a different command range than the Atlas robots, as spatial and temporal scaling factors are applied during motion retargeting to accommodate the G1's smaller kinematic scale.
    }
    \label{fig:g1_quantitative_analysis}
\end{figure*}
\clearpage

%
\begin{table*}[t]
\centering
\caption{\textbf{Extended data fidelity analysis metrics.} Comprehensive tracking performance comparison between the teacher and deployment policies across eight canonical motion trajectories for the Unitree G1 robot. Reported values indicate the Mean Absolute Error (MAE) for torso linear velocity, torso angular velocity, and joint positions, alongside the normalized mean time warping metrics converted to physical time units ($\text{ms}$). We note that the mean time warping metric based on DTW statistics can be computed as the total warp area divided by the length of the trajectory. Specifically, the warp area is defined by summing the absolute index differences between the aligned time pairs along the optimal warping path. To provide an intuitive temporal scale, we scale this discrete frame offset by the simulation control loop timestep, thereby converting the mean time warping metric into physical time units of \SI{}{\milli\second}.}
\label{tab:supplementary_g1_data_fidelity}
\vspace{0.1em}
%
%
\ifdefined\isArXiv
\small
\begin{tabular}{lcccccccc}
\toprule
                                                & \multicolumn{2}{c}{\textbf{Torso linear velocity} \; (\SI[per-mode=repeated-symbol]{}{\metre\per\second})}  
                                                & \multicolumn{2}{c}{\textbf{Torso angular velocity} \; (\SI[per-mode=repeated-symbol]{}{\radian\per\second})}   
                                                & \multicolumn{2}{c}{\textbf{Joint positions} \; (\SI{}{\radian})}   
                                                & \multicolumn{2}{c}{\textbf{Mean time warping} \; (\SI{}{\milli\second})}  \\
                                                & \textbf{Teacher}  & \textbf{Deploy.}       
                                                & \textbf{Teacher}  & \textbf{Deploy.}       
                                                & \textbf{Teacher}  & \textbf{Deploy.}       
                                                & \textbf{Teacher}  & \textbf{Deploy.} \\
\midrule
                        Slow Forward Walk       & 0.0341    & 0.0388        & 0.0421    & 0.0852        & 0.0294    & 0.0491    & 11.5929   & 27.3009   \\
\rowcolor[HTML]{EFEFEF} Normal Forward Walk     & 0.0441    & 0.0472        & 0.0541    & 0.0979        & 0.0376    & 0.0547    & 8.6093    & 58.7417   \\
                        Fast Forward Walk       & 0.0895    & 0.0847        & 0.0841    & 0.1438        & 0.0402    & 0.0598    & 4.0594    & 20.9901   \\
\rowcolor[HTML]{EFEFEF} Slow Backward Walk      & 0.0383    & 0.0550        & 0.0435    & 0.1149        & 0.0326    & 0.0549    & 8.0731    & 55.8140   \\
                        Normal Backward Walk    & 0.0418    & 0.0643        & 0.0496    & 0.1478        & 0.0328    & 0.0555    & 8.1499    & 62.8103   \\
\rowcolor[HTML]{EFEFEF} Fast Backward Walk      & 0.0916    & 0.1233        & 0.0665    & 0.2112        & 0.0427    & 0.0657    & 5.0000    & 46.5079   \\
                        Sidestep                & 0.0537    & 0.0546        & 0.0523    & 0.1054        & 0.0304    & 0.0549    & 13.7759   & 169.3776  \\
\rowcolor[HTML]{EFEFEF} Turn in Place           & 0.0322    & 0.0418        & 0.0753    & 0.1555        & 0.0275    & 0.0544    & 16.0697   & 65.7214   \\
\midrule
                        \textbf{Mean}           & 0.0480    & 0.0588        & 0.0551    & 0.1269        & 0.0330    & 0.0553    & 10.0481   & 68.7344   \\
\bottomrule
\end{tabular}
\else
%
%
\scriptsize
\begin{tabular}{lcccccccc}
\toprule
                                                & \multicolumn{2}{c}{\makecell{\textbf{Torso linear velocity}\\(\SI[per-mode=repeated-symbol]{}{\metre\per\second})}}  
                                                & \multicolumn{2}{c}{\makecell{\textbf{Torso angular velocity}\\(\SI[per-mode=repeated-symbol]{}{\radian\per\second})}}  
                                                & \multicolumn{2}{c}{\makecell{\textbf{Joint positions}\\(\SI{}{\radian})}}   
                                                & \multicolumn{2}{c}{\makecell{\textbf{Mean time warping}\\(\SI{}{\milli\second})}}  \\
                                                & \textbf{Teacher}  & \textbf{Deploy.}       
                                                & \textbf{Teacher}  & \textbf{Deploy.}       
                                                & \textbf{Teacher}  & \textbf{Deploy.}       
                                                & \textbf{Teacher}  & \textbf{Deploy.} \\
\midrule
                        Slow Forward Walk       & 0.0341    & 0.0388        & 0.0421    & 0.0852        & 0.0294    & 0.0491    & 11.5929   & 27.3009   \\
\rowcolor[HTML]{EFEFEF} Normal Forward Walk     & 0.0441    & 0.0472        & 0.0541    & 0.0979        & 0.0376    & 0.0547    & 8.6093    & 58.7417   \\
                        Fast Forward Walk       & 0.0895    & 0.0847        & 0.0841    & 0.1438        & 0.0402    & 0.0598    & 4.0594    & 20.9901   \\
\rowcolor[HTML]{EFEFEF} Slow Backward Walk      & 0.0383    & 0.0550        & 0.0435    & 0.1149        & 0.0326    & 0.0549    & 8.0731    & 55.8140   \\
                        Normal Backward Walk    & 0.0418    & 0.0643        & 0.0496    & 0.1478        & 0.0328    & 0.0555    & 8.1499    & 62.8103   \\
\rowcolor[HTML]{EFEFEF} Fast Backward Walk      & 0.0916    & 0.1233        & 0.0665    & 0.2112        & 0.0427    & 0.0657    & 5.0000    & 46.5079   \\
                        Sidestep                & 0.0537    & 0.0546        & 0.0523    & 0.1054        & 0.0304    & 0.0549    & 13.7759   & 169.3776  \\
\rowcolor[HTML]{EFEFEF} Turn in Place           & 0.0322    & 0.0418        & 0.0753    & 0.1555        & 0.0275    & 0.0544    & 16.0697   & 65.7214   \\
\midrule
                        \textbf{Mean}           & 0.0480    & 0.0588        & 0.0551    & 0.1269        & 0.0330    & 0.0553    & 10.0481   & 68.7344   \\
\bottomrule
\end{tabular}
\fi
\end{table*}

%
\begin{table*}[t]
\centering
\caption{\textbf{Atlas D1 and Unitree G1 Sim-to-real analysis results.} We provide the same sim-to-real analysis results for the Boston Dynamics D1 and Unitree G1 robot as presented for the Atlas R1 robot in \highlight{Table~1}.
We note that the G1 policy is evaluated under a different command than the Atlas robots as spatial and temporal scaling factors are applied during motion retargeting to accommodate the G1's smaller kinematic scale.}
\label{tab:sim2real_analysis_others}
\ifdefined\isArXiv
%
%
\begingroup
\small
\setlength{\tabcolsep}{3.5pt}
\renewcommand{\arraystretch}{1.05}

\sisetup{
  round-mode          = places,
  round-precision     = 2,
  table-number-alignment = center,
}

\begin{minipage}[t]{0.49\textwidth}
\centering
\begin{tabular}{lcccc}
\toprule
\multicolumn{1}{c}{} & \multicolumn{4}{c}{\textbf{Boston Dynamics Atlas D1}} \\
\cmidrule(lr){2-5}
& \multicolumn{2}{c}{\textbf{Steady-State Error}} & \multicolumn{2}{c}{\textbf{Mean Absolute Error}} \\
\cmidrule(lr){2-3}\cmidrule(lr){4-5}
\textbf{Command} & \textbf{Sim} & \textbf{Real} & \textbf{Sim} & \textbf{Real} \\
\midrule
\midrule
\multicolumn{5}{@{}l}{\textbf{Forward Walk}} \\
\addlinespace[0.2em]
(0.4, 0.0, 0.0)      & (0.03, 0.01) & (0.20, 0.02) & (0.08, 0.12) & (0.14, 0.08) \\
(1.2, 0.0, 0.0)      & (0.08, 0.08) & (0.11, 0.00) & (0.10, 0.18) & (0.16, 0.21) \\
(2.0, 0.0, 0.0)      & (0.31, 0.01) & (0.28, 0.17) & (0.23, 0.17) & (0.23, 0.29) \\
\midrule
\multicolumn{5}{@{}l}{\textbf{Backward Walk}} \\
\addlinespace[0.2em]
(-0.3, 0.0, 0.0)     & (0.09, 0.02) & (0.12, 0.07) & (0.09, 0.06) & (0.11, 0.08) \\
(-0.9, 0.0, 0.0)     & (0.07, 0.03) & (0.02, 0.02) & (0.10, 0.10) & (0.14, 0.15) \\
(-1.5, 0.0, 0.0)     & (0.14, 0.04) & (0.14, 0.03) & (0.15, 0.11) & (0.17, 0.16) \\
\midrule
\multicolumn{5}{@{}l}{\textbf{Sidestepping}} \\
\addlinespace[0.2em]
(0.0, -0.7, 0.0)     & (0.08, 0.00) & (0.07, 0.03) & (0.13, 0.05) & (0.16, 0.08) \\
(0.0, 0.7, 0.0)      & (0.05, 0.05) & (0.07, 0.05) & (0.16, 0.07) & (0.18, 0.07) \\
(0.0, -1.0, 0.0)     & (0.10, 0.03) & (0.13, 0.01) & (0.18, 0.06) & (0.19, 0.09) \\
(0.0, 1.0, 0.0)      & (0.05, 0.00) & (0.24, 0.09) & (0.19, 0.05) & (0.25, 0.08) \\
\midrule
\multicolumn{5}{@{}l}{\textbf{Turn in Place}} \\
\addlinespace[0.2em]
(0.0, 0.0, -1.0)     & (0.09, 0.10) & (0.07, 0.01) & (0.07, 0.10) & (0.06, 0.11) \\
(0.0, 0.0, 1.0)      & (0.04, 0.01) & (0.01, 0.08) & (0.06, 0.07) & (0.06, 0.11) \\
(0.0, 0.0, -2.0)     & (0.04, 0.08) & (0.06, 0.10) & (0.08, 0.13) & (0.08, 0.14) \\
(0.0, 0.0, 2.0)      & (0.06, 0.20) & (0.10, 0.12) & (0.08, 0.19) & (0.10, 0.20) \\
(0.0, 0.0, -3.0)     & (0.03, 0.37) & (0.02, 0.31) & (0.09, 0.29) & (0.10, 0.29) \\
(0.0, 0.0, 3.0)      & (0.06, 0.45) & (0.11, 0.66) & (0.10, 0.39) & (0.09, 0.42) \\
\midrule
\multicolumn{5}{@{}l}{\textbf{Forward Circle}} \\
\addlinespace[0.2em]
(1.0, 0.0, -1.0)     & (0.05, 0.11) & (0.09, 0.16) & (0.11, 0.16) & (0.09, 0.20) \\
(1.0, 0.0, 1.0)      & (0.10, 0.01) & (0.18, 0.17) & (0.09, 0.19) & (0.12, 0.16) \\
(1.0, 0.0, -1.5)     & (0.02, 0.13) & (0.19, 0.39) & (0.10, 0.16) & (0.14, 0.26) \\
(1.0, 0.0, 1.5)      & (0.17, 0.11) & (0.25, 0.19) & (0.12, 0.20) & (0.18, 0.23) \\
\bottomrule
\end{tabular}
\end{minipage}
\hfill
\begin{minipage}[t]{0.49\textwidth}
\centering
\begin{tabular}{lcccc}
\toprule
\multicolumn{1}{c}{} & \multicolumn{4}{c}{\textbf{Unitree G1}} \\
\cmidrule(lr){2-5}
 & \multicolumn{2}{c}{\textbf{Steady-State Error}} & \multicolumn{2}{c}{\textbf{Mean Absolute Error}} \\
\cmidrule(lr){2-3}\cmidrule(lr){4-5}
\textbf{Command} & \textbf{Sim} & \textbf{Real} & \textbf{Sim} & \textbf{Real} \\
\midrule
\midrule
\multicolumn{5}{@{}l}{\textbf{Forward Walk}} \\
\addlinespace[0.2em]
(0.36, 0.0, 0.0)     & (0.07, 0.06) & (0.03, 0.05) & (0.09, 0.20) & (0.10, 0.23) \\
(1.07, 0.0, 0.0)     & (0.01, 0.09) & (0.02, 0.01) & (0.10, 0.23) & (0.09, 0.26) \\
(1.79, 0.0, 0.0)     & (0.07, 0.03) & (0.08, 0.17) & (0.14, 0.28) & (0.11, 0.45) \\
\midrule
\multicolumn{5}{@{}l}{\textbf{Backward Walk}} \\
\addlinespace[0.2em]
(-0.27, 0.0, 0.0)    & (0.07, 0.05) & (0.06, 0.03) & (0.12, 0.19) & (0.10, 0.21) \\
(-0.80, 0.0, 0.0)    & (0.05, 0.00) & (0.05, 0.05) & (0.13, 0.24) & (0.11, 0.27) \\
(-1.34, 0.0, 0.0)    & (0.18, 0.07) & (0.22, 0.01) & (0.16, 0.31) & (0.17, 0.30) \\
\midrule
\multicolumn{5}{@{}l}{\textbf{Sidestepping}} \\
\addlinespace[0.2em]
(0.0, -0.67, 0.0)    & (0.04, 0.01) & (0.15, 0.07) & (0.20, 0.10) & (0.21, 0.14) \\
(0.0, 0.67, 0.0)     & (0.18, 0.06) & (0.20, 0.05) & (0.20, 0.08) & (0.19, 0.20) \\
(0.0, -0.89, 0.0)    & (0.11, 0.04) & (0.21, 0.04) & (0.20, 0.13) & (0.23, 0.24) \\
(0.0, 0.89, 0.0)     & (0.28, 0.01) & (0.21, 0.15) & (0.26, 0.09) & (0.22, 0.24) \\
\midrule
\multicolumn{5}{@{}l}{\textbf{Turn in Place}} \\
\addlinespace[0.2em]
(0.0, 0.0, -1.12)    & (0.05, 0.02) & (0.06, 0.18) & (0.10, 0.14) & (0.10, 0.24) \\
(0.0, 0.0, 1.12)     & (0.08, 0.13) & (0.05, 0.18) & (0.07, 0.20) & (0.06, 0.24) \\
(0.0, 0.0, -2.24)    & (0.07, 0.34) & (0.06, 0.41) & (0.10, 0.29) & (0.11, 0.36) \\
(0.0, 0.0, 2.24)     & (0.06, 0.29) & (0.06, 0.45) & (0.10, 0.29) & (0.12, 0.40) \\
(0.0, 0.0, -3.35)    & (0.04, 1.04) & (0.06, 0.93) & (0.13, 0.72) & (0.15, 0.66) \\
(0.0, 0.0, 3.35)     & (0.09, 0.65) & (0.09, 0.82) & (0.15, 0.49) & (0.16, 0.61) \\
\midrule
\multicolumn{5}{@{}l}{\textbf{Forward Circle}} \\
\addlinespace[0.2em]
(0.89, 0.0, -1.12)   & (0.06, 0.11) & (0.07, 0.18) & (0.10, 0.27) & (0.15, 0.35) \\
(0.89, 0.0, 1.12)    & (0.05, 0.14) & (0.03, 0.18) & (0.11, 0.22) & (0.10, 0.33) \\
(0.89, 0.0, -1.68)   & (0.12, 0.50) & (0.15, 0.61) & (0.12, 0.40) & (0.12, 0.50) \\
(0.89, 0.0, 1.68)    & (0.11, 0.36) & (0.06, 0.49) & (0.12, 0.31) & (0.09, 0.43) \\
\bottomrule
\end{tabular}
\end{minipage}
\endgroup
\else
%
%
\begingroup
\scriptsize
\setlength{\tabcolsep}{3.5pt}
\renewcommand{\arraystretch}{1.05}

\sisetup{
  round-mode          = places,
  round-precision     = 2,
  table-number-alignment = center,
}

\begin{minipage}[t]{0.49\textwidth}
\centering
\begin{tabular}{lcccc}
\toprule
\multicolumn{1}{c}{} & \multicolumn{4}{c}{\textbf{Boston Dynamics Atlas D1}} \\
\cmidrule(lr){2-5}
& \multicolumn{2}{c}{\textbf{Steady-State Error}} & \multicolumn{2}{c}{\textbf{Mean Absolute Error}} \\
\cmidrule(lr){2-3}\cmidrule(lr){4-5}
\textbf{Command} & \textbf{Sim} & \textbf{Real} & \textbf{Sim} & \textbf{Real} \\
\midrule
\midrule
\multicolumn{5}{@{}l}{\textbf{Forward Walk}} \\
\addlinespace[0.2em]
(0.4, 0.0, 0.0)      & (0.03, 0.01) & (0.20, 0.02) & (0.08, 0.12) & (0.14, 0.08) \\
(1.2, 0.0, 0.0)      & (0.08, 0.08) & (0.11, 0.00) & (0.10, 0.18) & (0.16, 0.21) \\
(2.0, 0.0, 0.0)      & (0.31, 0.01) & (0.28, 0.17) & (0.23, 0.17) & (0.23, 0.29) \\
\midrule
\multicolumn{5}{@{}l}{\textbf{Backward Walk}} \\
\addlinespace[0.2em]
(-0.3, 0.0, 0.0)     & (0.09, 0.02) & (0.12, 0.07) & (0.09, 0.06) & (0.11, 0.08) \\
(-0.9, 0.0, 0.0)     & (0.07, 0.03) & (0.02, 0.02) & (0.10, 0.10) & (0.14, 0.15) \\
(-1.5, 0.0, 0.0)     & (0.14, 0.04) & (0.14, 0.03) & (0.15, 0.11) & (0.17, 0.16) \\
\midrule
\multicolumn{5}{@{}l}{\textbf{Sidestepping}} \\
\addlinespace[0.2em]
(0.0, -0.7, 0.0)     & (0.08, 0.00) & (0.07, 0.03) & (0.13, 0.05) & (0.16, 0.08) \\
(0.0, 0.7, 0.0)      & (0.05, 0.05) & (0.07, 0.05) & (0.16, 0.07) & (0.18, 0.07) \\
(0.0, -1.0, 0.0)     & (0.10, 0.03) & (0.13, 0.01) & (0.18, 0.06) & (0.19, 0.09) \\
(0.0, 1.0, 0.0)      & (0.05, 0.00) & (0.24, 0.09) & (0.19, 0.05) & (0.25, 0.08) \\
\midrule
\multicolumn{5}{@{}l}{\textbf{Turn in Place}} \\
\addlinespace[0.2em]
(0.0, 0.0, -1.0)     & (0.09, 0.10) & (0.07, 0.01) & (0.07, 0.10) & (0.06, 0.11) \\
(0.0, 0.0, 1.0)      & (0.04, 0.01) & (0.01, 0.08) & (0.06, 0.07) & (0.06, 0.11) \\
(0.0, 0.0, -2.0)     & (0.04, 0.08) & (0.06, 0.10) & (0.08, 0.13) & (0.08, 0.14) \\
(0.0, 0.0, 2.0)      & (0.06, 0.20) & (0.10, 0.12) & (0.08, 0.19) & (0.10, 0.20) \\
(0.0, 0.0, -3.0)     & (0.03, 0.37) & (0.02, 0.31) & (0.09, 0.29) & (0.10, 0.29) \\
(0.0, 0.0, 3.0)      & (0.06, 0.45) & (0.11, 0.66) & (0.10, 0.39) & (0.09, 0.42) \\
\midrule
\multicolumn{5}{@{}l}{\textbf{Forward Circle}} \\
\addlinespace[0.2em]
(1.0, 0.0, -1.0)     & (0.05, 0.11) & (0.09, 0.16) & (0.11, 0.16) & (0.09, 0.20) \\
(1.0, 0.0, 1.0)      & (0.10, 0.01) & (0.18, 0.17) & (0.09, 0.19) & (0.12, 0.16) \\
(1.0, 0.0, -1.5)     & (0.02, 0.13) & (0.19, 0.39) & (0.10, 0.16) & (0.14, 0.26) \\
(1.0, 0.0, 1.5)      & (0.17, 0.11) & (0.25, 0.19) & (0.12, 0.20) & (0.18, 0.23) \\
\bottomrule
\end{tabular}
\end{minipage}
\hfill
\begin{minipage}[t]{0.49\textwidth}
\centering
\begin{tabular}{lcccc}
\toprule
\multicolumn{1}{c}{} & \multicolumn{4}{c}{\textbf{Unitree G1}} \\
\cmidrule(lr){2-5}
 & \multicolumn{2}{c}{\textbf{Steady-State Error}} & \multicolumn{2}{c}{\textbf{Mean Absolute Error}} \\
\cmidrule(lr){2-3}\cmidrule(lr){4-5}
\textbf{Command} & \textbf{Sim} & \textbf{Real} & \textbf{Sim} & \textbf{Real} \\
\midrule
\midrule
\multicolumn{5}{@{}l}{\textbf{Forward Walk}} \\
\addlinespace[0.2em]
(0.36, 0.0, 0.0)     & (0.07, 0.06) & (0.03, 0.05) & (0.09, 0.20) & (0.10, 0.23) \\
(1.07, 0.0, 0.0)     & (0.01, 0.09) & (0.02, 0.01) & (0.10, 0.23) & (0.09, 0.26) \\
(1.79, 0.0, 0.0)     & (0.07, 0.03) & (0.08, 0.17) & (0.14, 0.28) & (0.11, 0.45) \\
\midrule
\multicolumn{5}{@{}l}{\textbf{Backward Walk}} \\
\addlinespace[0.2em]
(-0.27, 0.0, 0.0)    & (0.07, 0.05) & (0.06, 0.03) & (0.12, 0.19) & (0.10, 0.21) \\
(-0.80, 0.0, 0.0)    & (0.05, 0.00) & (0.05, 0.05) & (0.13, 0.24) & (0.11, 0.27) \\
(-1.34, 0.0, 0.0)    & (0.18, 0.07) & (0.22, 0.01) & (0.16, 0.31) & (0.17, 0.30) \\
\midrule
\multicolumn{5}{@{}l}{\textbf{Sidestepping}} \\
\addlinespace[0.2em]
(0.0, -0.67, 0.0)    & (0.04, 0.01) & (0.15, 0.07) & (0.20, 0.10) & (0.21, 0.14) \\
(0.0, 0.67, 0.0)     & (0.18, 0.06) & (0.20, 0.05) & (0.20, 0.08) & (0.19, 0.20) \\
(0.0, -0.89, 0.0)    & (0.11, 0.04) & (0.21, 0.04) & (0.20, 0.13) & (0.23, 0.24) \\
(0.0, 0.89, 0.0)     & (0.28, 0.01) & (0.21, 0.15) & (0.26, 0.09) & (0.22, 0.24) \\
\midrule
\multicolumn{5}{@{}l}{\textbf{Turn in Place}} \\
\addlinespace[0.2em]
(0.0, 0.0, -1.12)    & (0.05, 0.02) & (0.06, 0.18) & (0.10, 0.14) & (0.10, 0.24) \\
(0.0, 0.0, 1.12)     & (0.08, 0.13) & (0.05, 0.18) & (0.07, 0.20) & (0.06, 0.24) \\
(0.0, 0.0, -2.24)    & (0.07, 0.34) & (0.06, 0.41) & (0.10, 0.29) & (0.11, 0.36) \\
(0.0, 0.0, 2.24)     & (0.06, 0.29) & (0.06, 0.45) & (0.10, 0.29) & (0.12, 0.40) \\
(0.0, 0.0, -3.35)    & (0.04, 1.04) & (0.06, 0.93) & (0.13, 0.72) & (0.15, 0.66) \\
(0.0, 0.0, 3.35)     & (0.09, 0.65) & (0.09, 0.82) & (0.15, 0.49) & (0.16, 0.61) \\
\midrule
\multicolumn{5}{@{}l}{\textbf{Forward Circle}} \\
\addlinespace[0.2em]
(0.89, 0.0, -1.12)   & (0.06, 0.11) & (0.07, 0.18) & (0.10, 0.27) & (0.15, 0.35) \\
(0.89, 0.0, 1.12)    & (0.05, 0.14) & (0.03, 0.18) & (0.11, 0.22) & (0.10, 0.33) \\
(0.89, 0.0, -1.68)   & (0.12, 0.50) & (0.15, 0.61) & (0.12, 0.40) & (0.12, 0.50) \\
(0.89, 0.0, 1.68)    & (0.11, 0.36) & (0.06, 0.49) & (0.12, 0.31) & (0.09, 0.43) \\
\bottomrule
\end{tabular}
\end{minipage}
\endgroup
\fi
\end{table*}

%
%
\begin{table}[ht]
\centering
\caption{\textbf{Summary of observation terms.} This table details the components of the observation space. No scaling or clipping is applied to these raw values. Checkmarks ($\checkmark$) indicates the inclusion of a specific term within the corresponding observation vectors---teacher observations $\oteacher$, deployment observations $\odep$, and critic observation $\ocritic$.}
\label{tab:supplementary_observation}
\ifdefined\isArXiv
%
%
\begin{tabular}{ lcccccc}
\toprule
                        \textbf{Term Name}              & \textbf{Definition}                           & \textbf{Noise}            & \textbf{Dim.}          & \textbf{Teacher} & \textbf{Deploy} & \textbf{Critic} \\  
\midrule
\multicolumn{7}{c}{\textbf{Proprioceptive observations} \; $\oprop$} \\
\midrule
\rowcolor[HTML]{EFEFEF} IMU linear velocity             & ${}_\mathcal{T} \linvel_{\mathcal{IT}}$       & $\mathcal{N}(0,\ 0.1)$    &  $3 \times \numimu$    & \checkmark       & \checkmark$^*$  & \checkmark$^{\dagger}$ \\  
                        IMU angular velocity            & ${}_\mathcal{T} \angvel_{\mathcal{IT}}$       & $\mathcal{N}(0,\ 0.1)$    &  $3 \times \numimu$    & \checkmark       & \checkmark      & \checkmark$^{\dagger}$ \\  
\rowcolor[HTML]{EFEFEF} IMU projected gravity           & ${}_\mathcal{T} \gravity              $       & $\mathcal{N}(0,\ 0.015)$  &  $3 \times \numimu$    & \checkmark       & \checkmark      & \checkmark$^{\dagger}$ \\  
                        Joint positions                 & $\jointpos$                                   & $\mathcal{N}(0,\ 0.005)$  &  $\numjoint$           & \checkmark       & \checkmark      & \checkmark$^{\dagger}$ \\  
\rowcolor[HTML]{EFEFEF} Joint velocities                & $\jointvel$                                   & $\mathcal{N}(0,\ 0.25)$   &  $\numjoint$           & \checkmark       & \checkmark      & \checkmark$^{\dagger}$ \\  
                        Previous action                 & $\action_{t-1}$                               & --                        &  $\numaction$          & \checkmark       & \checkmark      & \checkmark  \\
\midrule
\multicolumn{7}{c}{\textbf{Reference observations} \; $\oref$} \\
\midrule
\rowcolor[HTML]{EFEFEF} Reference torso height              & ${}_\mathcal{I} \hat{\bm r}^z_{\mathcal{IB}}$ & --                        & 1                      & \checkmark       &                 & \checkmark \\  
                        Reference torso linear velocity     & ${}_\mathcal{B} \hat{\linvel}_{\mathcal{IB}}$ & --                        & 3                      & \checkmark       &                 & \checkmark \\  
\rowcolor[HTML]{EFEFEF} Reference torso angular velocity    & ${}_\mathcal{B} \hat{\angvel}_{\mathcal{IB}}$ & --                        & 3                      & \checkmark       &                 & \checkmark \\  
                        Gravity vector in reference torso frame  & ${}_{\mathcal{B}} \gravityref$        & --                        & 3                      & \checkmark       &                 & \checkmark \\
\rowcolor[HTML]{EFEFEF} Reference joint positions           & $\jointposref$                                & --                        & $\numjoint$            & \checkmark       &                 & \checkmark \\
\midrule
\multicolumn{7}{c}{\textbf{Command observations} \; $\ocmd$} \\
\midrule
\rowcolor[HTML]{EFEFEF} Longitudinal velocity command   & $\hat{v}_{\text{long}}$                       & --                        & 1                      & \checkmark       & \checkmark      & \checkmark \\
                        Lateral velocity command        & $\hat{v}_{\text{lat}}$                        & --                        & 1                      & \checkmark       & \checkmark      & \checkmark \\
\rowcolor[HTML]{EFEFEF} Turn rate command               & $\hat{\omega}_{\text{turn}}$                  & --                        & 1                      & \checkmark       & \checkmark      & \checkmark \\
\midrule
\multicolumn{7}{c}{\textbf{Privileged observations} \; $\opriv$} \\
\midrule
\rowcolor[HTML]{EFEFEF} Torso height                        & ${}_\mathcal{I} {\bm r}^z_{\mathcal{IB}}$     & --                        & 1                      &                  &                 & \checkmark \\
                        End effector wrenches               & ${}_\mathcal{I} {\bm w}_e$                & --                        & $6 \times \numee$      &                  &                 & \checkmark \\
\rowcolor[HTML]{EFEFEF} End effector positions in torso frame &${}_\mathcal{B} {\bm r}_{\mathcal{BE}}$   & --                        & $3 \times \numee$      &                  &                 & \checkmark \\        
                        End effector linear velocities in torso frame &${}_\mathcal{B} {\linvel}_{\mathcal{BE}}$   & --              & $3 \times \numee$      &                  &                 & \checkmark \\
\rowcolor[HTML]{EFEFEF} Assistive force norm (curriculum)   & $\|\beta \bm f_{\text{assist}}\|$         & --                        & $1$                    &                  &                 & \checkmark \\                        
                        Assistive torque norm (curriculum)  & $\|\beta \bm \tau_{\text{assist}}\|$      & --                        & $1$                    &                  &                 & \checkmark \\
\rowcolor[HTML]{EFEFEF} Assistive wrench scale              & $\beta$                                       & --                        & $1$                    &                  &                 & \checkmark \\
                        Support observation                 & $\osup$                                       & --                        & $1$                    &                  &                 & \checkmark \\
\rowcolor[HTML]{EFEFEF} Terrain height observation          & $\oterrain$                                   & --                        & $17 \times 11$         &                  &                 & \checkmark$^{\ddag}$ \\
\bottomrule
\end{tabular}
\caption*{
\begin{tabular}{ll}
$^{*}$              & The linear velocity observation is accessible on Boston Dynamics robots, whereas it is unavailable on the Unitree G1. \\
$^{\dagger}$    & Critic observes the marked terms without the addition of sensor noise. \\
$^{\ddag}$    & Terrain height observation is used only for Unitree G1 to support training with terrain roughness randomization. 
\end{tabular}
}
\else
%
%
\footnotesize
\begin{tabular}{ lcccccc}
\toprule
                        \textbf{Term Name}              & \textbf{Definition}                           & \textbf{Noise}            & \textbf{Dim.}          & \textbf{Teacher} & \textbf{Deploy} & \textbf{Critic} \\  
\midrule
\multicolumn{7}{c}{\textbf{Proprioceptive observations} \; $\oprop$} \\
\midrule
\rowcolor[HTML]{EFEFEF} IMU linear velocity             & ${}_\mathcal{T} \linvel_{\mathcal{IT}}$       & $\mathcal{N}(0,\ 0.1)$    &  $3 \times \numimu$    & \checkmark       & \checkmark$^*$  & \checkmark$^{\dagger}$ \\  
                        IMU angular velocity            & ${}_\mathcal{T} \angvel_{\mathcal{IT}}$       & $\mathcal{N}(0,\ 0.1)$    &  $3 \times \numimu$    & \checkmark       & \checkmark      & \checkmark$^{\dagger}$ \\  
\rowcolor[HTML]{EFEFEF} IMU projected gravity           & ${}_\mathcal{T} \gravity              $       & $\mathcal{N}(0,\ 0.015)$  &  $3 \times \numimu$    & \checkmark       & \checkmark      & \checkmark$^{\dagger}$ \\  
                        Joint positions                 & $\jointpos$                                   & $\mathcal{N}(0,\ 0.005)$  &  $\numjoint$           & \checkmark       & \checkmark      & \checkmark$^{\dagger}$ \\  
\rowcolor[HTML]{EFEFEF} Joint velocities                & $\jointvel$                                   & $\mathcal{N}(0,\ 0.25)$   &  $\numjoint$           & \checkmark       & \checkmark      & \checkmark$^{\dagger}$ \\  
                        Previous action                 & $\action_{t-1}$                               & --                        &  $\numaction$          & \checkmark       & \checkmark      & \checkmark  \\
\midrule
\multicolumn{7}{c}{\textbf{Reference observations} \; $\oref$} \\
\midrule
\rowcolor[HTML]{EFEFEF} Reference torso height              & ${}_\mathcal{I} \hat{\bm r}^z_{\mathcal{IB}}$ & --                        & 1                      & \checkmark       &                 & \checkmark \\  
                        Reference torso linear velocity     & ${}_\mathcal{B} \hat{\linvel}_{\mathcal{IB}}$ & --                        & 3                      & \checkmark       &                 & \checkmark \\  
\rowcolor[HTML]{EFEFEF} Reference torso angular velocity    & ${}_\mathcal{B} \hat{\angvel}_{\mathcal{IB}}$ & --                        & 3                      & \checkmark       &                 & \checkmark \\  
                        Gravity vector in reference torso frame  & ${}_{\mathcal{B}} \gravityref$        & --                        & 3                      & \checkmark       &                 & \checkmark \\
\rowcolor[HTML]{EFEFEF} Reference joint positions           & $\jointposref$                                & --                        & $\numjoint$            & \checkmark       &                 & \checkmark \\
\midrule
\multicolumn{7}{c}{\textbf{Command observations} \; $\ocmd$} \\
\midrule
\rowcolor[HTML]{EFEFEF} Longitudinal velocity command   & $\hat{v}_{\text{long}}$                       & --                        & 1                      & \checkmark       & \checkmark      & \checkmark \\
                        Lateral velocity command        & $\hat{v}_{\text{lat}}$                        & --                        & 1                      & \checkmark       & \checkmark      & \checkmark \\
\rowcolor[HTML]{EFEFEF} Turn rate command               & $\hat{\omega}_{\text{turn}}$                  & --                        & 1                      & \checkmark       & \checkmark      & \checkmark \\
\midrule
\multicolumn{7}{c}{\textbf{Privileged observations} \; $\opriv$} \\
\midrule
\rowcolor[HTML]{EFEFEF} Torso height                        & ${}_\mathcal{I} {\bm r}^z_{\mathcal{IB}}$     & --                        & 1                      &                  &                 & \checkmark \\
                        End effector wrenches               & ${}_\mathcal{I} {\bm w}_e$                & --                        & $6 \times \numee$      &                  &                 & \checkmark \\
\rowcolor[HTML]{EFEFEF} End effector positions in torso frame &${}_\mathcal{B} {\bm r}_{\mathcal{BE}}$   & --                        & $3 \times \numee$      &                  &                 & \checkmark \\        
                        End effector linear velocities in torso frame &${}_\mathcal{B} {\linvel}_{\mathcal{BE}}$   & --              & $3 \times \numee$      &                  &                 & \checkmark \\
\rowcolor[HTML]{EFEFEF} Assistive force norm (curriculum)   & $\|\beta \bm f_{\text{assist}}\|$         & --                        & $1$                    &                  &                 & \checkmark \\                        
                        Assistive torque norm (curriculum)  & $\|\beta \bm \tau_{\text{assist}}\|$      & --                        & $1$                    &                  &                 & \checkmark \\
\rowcolor[HTML]{EFEFEF} Assistive wrench scale              & $\beta$                                       & --                        & $1$                    &                  &                 & \checkmark \\
                        Support observation                 & $\osup$                                       & --                        & $1$                    &                  &                 & \checkmark \\
\rowcolor[HTML]{EFEFEF} Terrain height observation          & $\oterrain$                                   & --                        & $17 \times 11$         &                  &                 & \checkmark$^{\ddag}$ \\
\bottomrule
\vspace{0.1em}
\end{tabular}
\caption*{
$^{*}$ The linear velocity observation is accessible on Boston Dynamics robots, whereas it is unavailable on the Unitree G1. \\
$^{\dagger}$ Critic observes the marked terms without the addition of sensor noise. \\
$^{\ddag}$ Terrain height observation is used only for Unitree G1 to support training with terrain roughness randomization. 
}
\fi
\end{table}

%
\begin{table}[ht]
\centering
\caption{\textbf{Termination conditions.} Criteria marked with an asterisk ($*$) are applied exclusively to tasks utilizing reference motion data, whereas those marked with a dagger ($\dagger$) are applied exclusively to the goal-conditioned task in Stage~II.}
\label{tab:supplementary_termination}
\vspace{0.1em}
\ifdefined\isArXiv
%
%
\begin{tabular}{lcccc}
\toprule
                        \multirow{2}{*}{\textbf{Term}}                      & \multicolumn{4}{c}{\textbf{Condition}}                                                        \\
                                                                            &                       & \textbf{Unitree G1}   & \textbf{Atlas R1}     & \textbf{Atlas D1}     \\
\midrule
                        Body-ground penetration                             & Penetration           & > \SI{0.03}{\metre}   & > \SI{0.03}{\metre}   & > \SI{0.03}{\metre}   \\
\rowcolor[HTML]{EFEFEF} Large contact force                                 & Contact force         & > \SI{6000}{\newton}  & > \SI{12000}{\newton} & > \SI{12000}{\newton} \\
                        Torso height deviation from reference$^*$      & Deviation             & > \SI{0.3}{\metre}    & > \SI{0.4}{\metre}    & > \SI{0.4}{\metre}    \\
\rowcolor[HTML]{EFEFEF} Torso projected gravity deviation from reference$^*$   & Deviation angle       & > \SI{0.6}{\radian}   & > \SI{0.6}{\radian}   & > \SI{0.6}{\radian}   \\
                        Foot self collision                                 & \multicolumn{4}{c}{If collision between two feet is detected} \\
\rowcolor[HTML]{EFEFEF} Torso height$^\dagger$                              & Torso height          & < \SI{0.5}{\metre}    & < \SI{0.5}{\metre}    & < \SI{0.5}{\metre}    \\
\bottomrule
\end{tabular}
\else
%
%
\footnotesize
\begin{tabular}{lcccc}
\toprule
                        \multirow{2}{*}{\textbf{Term}}                      & \multicolumn{4}{c}{\textbf{Condition}}                                                        \\
                                                                            &                       & \textbf{Unitree G1}   & \textbf{Atlas R1}     & \textbf{Atlas D1}     \\
\midrule
                        Body-ground penetration                             & Penetration           & $>$ \SI{0.03}{\metre}   & $>$ \SI{0.03}{\metre}   & $>$ \SI{0.03}{\metre}   \\
\rowcolor[HTML]{EFEFEF} Large contact force                                 & Contact force         & $>$ \SI{6000}{\newton}  & $>$ \SI{12000}{\newton} & $>$ \SI{12000}{\newton} \\
                        Torso height deviation from reference$^*$      & Deviation             & $>$ \SI{0.3}{\metre}    & $>$ \SI{0.4}{\metre}    & $>$ \SI{0.4}{\metre}    \\
\rowcolor[HTML]{EFEFEF} Torso projected gravity deviation from reference$^*$   & Deviation angle       & $>$ \SI{0.6}{\radian}   & $>$ \SI{0.6}{\radian}   & $>$ \SI{0.6}{\radian}   \\
                        Foot self collision                                 & \multicolumn{4}{c}{If collision between two feet is detected} \\
\rowcolor[HTML]{EFEFEF} Torso height$^\dagger$                              & Torso height          & $<$ \SI{0.5}{\metre}    & $<$ \SI{0.5}{\metre}    & $<$ \SI{0.5}{\metre}    \\
\bottomrule
\end{tabular}
\fi
\end{table}

%
\begin{table}[ht]
\small
\centering
\caption{\textbf{Domain randomization terms.}}
\label{tab:supplementary_domain_randomization}
\ifdefined\isArXiv
%
%
\begin{tabular}{lcccc}
\toprule
                        \multirow{2}{*}{\textbf{Term}}          &   & \multicolumn{3}{c}{\textbf{Value}}                                                                                    \\  
                                                                &   &\textbf{Unitree G1}                        & \textbf{Atlas R1}                             & \textbf{Atlas D1}         \\
                                    
\midrule
                        Feet static friction                    &   & $\mathcal{U}(0.6,\ 1.3)$                  & $\mathcal{U}(0.6,\ 1.0)$                      & $\mathcal{U}(0.6,\ 1.0)$  \\  
\rowcolor[HTML]{EFEFEF} Feet dynamic friction                   &   & $\mathcal{U}(0.5,\ 0.9)$                  & $\mathcal{U}(0.5,\ 0.9)$                      & $\mathcal{U}(0.5,\ 0.9)$  \\  
                        Feet restitution                        &   & $\mathcal{U}(0.0,\ 0.2)$                  & $\mathcal{U}(0.0,\ 0.2)$                      & $\mathcal{U}(0.0,\ 0.2)$  \\  
\rowcolor[HTML]{EFEFEF} Body mass scale factor                  & For all bodies & $\mathcal{U}(0.9,\ 1.15)$    & --                                            & -- \\ 
                        Torso mass perturbation                 &   & --                                        & $\mathcal{U}(-8.0,\ 8.0)\;\SI{}{\kilogram}$    & $\mathcal{U}(-8.0,\ 8.0)\;\SI{}{\kilogram}$ \\ 
\rowcolor[HTML]{EFEFEF} Torso center of mass perturb.           & 
                                                                \begin{tabular}{c}
                                                                ${}_\mathcal{B}\position^x$ \\ 
                                                                ${}_\mathcal{B}\position^y$ \\
                                                                ${}_\mathcal{B}\position^z$ \\
                                                                \end{tabular}
                                                                &
                                                                \begin{tabular}{c} 
                                                                $\mathcal{U}(-0.025,\ 0.025)\;\SI{}{\metre}$\\
                                                                $\mathcal{U}(-0.05,\ 0.05)\;\SI{}{\metre}$\\
                                                                $\mathcal{U}(-0.05,\ 0.05)\;\SI{}{\metre}$\\
                                                                \end{tabular}
                                                                & 
                                                                \begin{tabular}{c} 
                                                                $\mathcal{U}(-0.1,\ 0.1)\;\SI{}{\metre}$\\
                                                                $\mathcal{U}(-0.1,\ 0.1)\;\SI{}{\metre}$\\
                                                                $\mathcal{U}(-0.1,\ 0.1)\;\SI{}{\metre}$\\
                                                                \end{tabular}
                                                                & -- \\  
                        Joint friction                          & 
                                                                \begin{tabular}{c}
                                                                Ankle \\ 
                                                                Waist \\
                                                                \end{tabular}
                                                                &
                                                                \begin{tabular}{c} 
                                                                $\mathcal{U}(0.0,\ 0.05)$ \\
                                                                $\mathcal{U}(0.0,\ 0.1)$ \\
                                                                \end{tabular}                               
                                                                &
                                                                \begin{tabular}{c} 
                                                                $\mathcal{U}(0.0,\ 0.1)$ \\
                                                                -- \\
                                                                \end{tabular}                               
                                                                & -- \\  
\rowcolor[HTML]{EFEFEF} Nominal joint position perturb.         & 
                                                                \begin{tabular}{c}
                                                                Ankle \\
                                                                Others \\
                                                                \end{tabular}                               
                                                                &
                                                                \begin{tabular}{c}
                                                                $\mathcal{U}(-0.03,\ 0.03)$\;\SI{}{\radian} \\
                                                                $\mathcal{U}(-0.015,\ 0.015)$\;\SI{}{\radian} \\
                                                                \end{tabular}
                                                                & --                             & -- \\  
                        Impulsive push                          & 
                                                                \begin{tabular}{c}
                                                                Interval\\
                                                                $\linvel^{x}$,\ $\linvel^{y}$\\
                                                                $\dot{\phi}$,\ $\dot{\theta}$ \\
                                                                \end{tabular}
                                                                &
                                                                \begin{tabular}{c} 
                                                                $\mathcal{U}(0,\ 10)\;\SI{}{\second}$\\
                                                                $\mathcal{U}(-0.3,\ 0.3)\;\SI[per-mode=repeated-symbol]{}{\meter\per\second}$ \\
                                                                $\mathcal{U}(-0.2,\ 0.2)\;\SI[per-mode=repeated-symbol]{}{\radian\per\second}$ \\
                                                                \end{tabular}
                                                                & 
                                                                \begin{tabular}{c} 
                                                                $\mathcal{U}(0,\ 10)\;\SI{}{\second}$\\
                                                                $\mathcal{U}(-0.5,\ 0.5)\;\SI[per-mode=repeated-symbol]{}{\meter\per\second}$ \\
                                                                \\
                                                                \end{tabular}
                                                                &
                                                                \begin{tabular}{c} 
                                                                $\mathcal{U}(0,\ 10)\;\SI{}{\second}$\\
                                                                $\mathcal{U}(-0.5,\ 0.5)\;\SI[per-mode=repeated-symbol]{}{\meter\per\second}$ \\
                                                                \\
                                                                \end{tabular}
                                                                \\ 
\rowcolor[HTML]{EFEFEF} External force to torso                 &
                                                                \begin{tabular}{c}
                                                                Interval\\
                                                                ${}_{\mathcal{I}}{\bm f}^{x}$,\ ${}_{\mathcal{I}}{\bm f}^{y}$,\ ${}_{\mathcal{I}}{\bm f}^{z}$ \\
                                                                \end{tabular}
                                                                &
                                                                \begin{tabular}{c} 
                                                                $\mathcal{U}(2,\ 4)\;\SI{}{\second}$\\
                                                                $\mathcal{U}(-5.0,\ 5.0)\;\SI{}{\newton}$ \\
                                                                \end{tabular}
                                                                & -- & --\\ 
                        External torque to torso                &
                                                                \begin{tabular}{c}
                                                                Interval\\
                                                                ${}_{\mathcal{I}}{\bm \tau}^{x}$,\ ${}_{\mathcal{I}}{\bm \tau}^{y}$,\ ${}_{\mathcal{I}}{\bm \tau}^{z}$ \\
                                                                \end{tabular}
                                                                &
                                                                \begin{tabular}{c} 
                                                                $\mathcal{U}(2,\ 4)\;\SI{}{\second}$\\
                                                                $\mathcal{U}(-5.0,\ 5.0)\;\SI[inter-unit-product = \ensuremath{{}\cdot{}}]{}{\newton\metre}$ \\
                                                                \end{tabular}
                                                                & -- & --\\  
\rowcolor[HTML]{EFEFEF} Initial torso pose perturb.             &   
                                                                \begin{tabular}{c}  
                                                                ${}_\mathcal{I} \position^x$,\ ${}_\mathcal{I} \position^y$,\ ${}_\mathcal{I} \position^z$ \\
                                                                $\phi$,\ $\theta$,\ $\psi$\\
                                                                \end{tabular}
                                                                & 
                                                                \begin{tabular}{c} 
                                                                $\mathcal{U}(-0.1,/ 0.1)$ \SI{}{\metre} \\
                                                                $\mathcal{U}(-0.1,/ 0.1)$ \SI{}{\radian} \\
                                                                \end{tabular}
                                                                & -- & --
                                                                \\ 
                        Initial torso velocity perturb.         &   
                                                                \begin{tabular}{c}  
                                                                ${}_\mathcal{I} \linvel^x$,\ ${}_\mathcal{I} \linvel^y$,\ ${}_\mathcal{I} \linvel^z$ \\
                                                                $\dot\phi$,\ $\dot\theta$,\ $\dot\psi$\\
                                                                \end{tabular}
                                                                & 
                                                                \begin{tabular}{c} 
                                                                $\mathcal{U}(-0.1,/ 0.1)\;\SI[per-mode=repeated-symbol]{}{\metre\per\second}$ \\
                                                                $\mathcal{U}(-0.1,/ 0.1)\;\SI[per-mode=repeated-symbol]{}{\radian\per\second}$ \\
                                                                \end{tabular}
                                                                & 
                                                                \begin{tabular}{c} 
                                                                $\mathcal{U}(-0.1,/ 0.1)\;\SI[per-mode=repeated-symbol]{}{\metre\per\second}$ \\
                                                                $\mathcal{U}(-0.1,/ 0.1)\;\SI[per-mode=repeated-symbol]{}{\radian\per\second}$ \\
                                                                \end{tabular}
                                                                & 
                                                                \begin{tabular}{c} 
                                                                $\mathcal{U}(-0.1,/ 0.1)\;\SI[per-mode=repeated-symbol]{}{\metre\per\second}$ \\
                                                                $\mathcal{U}(-0.1,/ 0.1)\;\SI[per-mode=repeated-symbol]{}{\radian\per\second}$ \\
                                                                \end{tabular}
                                                                \\ 
\rowcolor[HTML]{EFEFEF} Initial joint positions perturb.        &   
                                                                & $\mathcal{U}(-0.2, 0.2)\;\SI{}{\radian}$  
                                                                & $\mathcal{U}(-0.2, 0.2)\;\SI{}{\radian}$  
                                                                & $\mathcal{U}(-0.2, 0.2)\;\SI{}{\radian}$ \\ 
                        Initial joint velocities perturb.       &   
                                                                & $\mathcal{U}(-0.5, 0.5)\;\SI[per-mode=repeated-symbol]{}{\radian\per\second}$ 
                                                                & $\mathcal{U}(-0.5, 0.5)\;\SI[per-mode=repeated-symbol]{}{\radian\per\second}$
                                                                & $\mathcal{U}(-0.5, 0.5)\;\SI[per-mode=repeated-symbol]{}{\radian\per\second}$ \\ 
                        Terrain roughness                       &   & $\mathcal{U}(-0.025, 0.025)\;\SI{}{\metre}$  & -- & -- \\   
\bottomrule
\end{tabular}
\else
%
%
\scriptsize
\begin{tabular}{lcccc}
\toprule
                        \multirow{2}{*}{\textbf{Term}}          &   & \multicolumn{3}{c}{\textbf{Value}}                                                                                    \\  
                                                                &   &\textbf{Unitree G1}                        & \textbf{Atlas R1}                             & \textbf{Atlas D1}         \\
                                    
\midrule
                        Feet static friction                    &   & $\mathcal{U}(0.6,\ 1.3)$                  & $\mathcal{U}(0.6,\ 1.0)$                      & $\mathcal{U}(0.6,\ 1.0)$  \\  
\rowcolor[HTML]{EFEFEF} Feet dynamic friction                   &   & $\mathcal{U}(0.5,\ 0.9)$                  & $\mathcal{U}(0.5,\ 0.9)$                      & $\mathcal{U}(0.5,\ 0.9)$  \\  
                        Feet restitution                        &   & $\mathcal{U}(0.0,\ 0.2)$                  & $\mathcal{U}(0.0,\ 0.2)$                      & $\mathcal{U}(0.0,\ 0.2)$  \\  
\rowcolor[HTML]{EFEFEF} Body mass scale factor                  & For all bodies & $\mathcal{U}(0.9,\ 1.15)$    & --                                            & -- \\ 
                        Torso mass perturbation                 &   & --                                        & $\mathcal{U}(-8.0,\ 8.0)\;\SI{}{\kilogram}$    & $\mathcal{U}(-8.0,\ 8.0)\;\SI{}{\kilogram}$ \\ 
\rowcolor[HTML]{EFEFEF} Torso center of mass perturb.           & 
                                                                \begin{tabular}{c}
                                                                ${}_\mathcal{B}\position^x$ \\ 
                                                                ${}_\mathcal{B}\position^y$ \\
                                                                ${}_\mathcal{B}\position^z$ \\
                                                                \end{tabular}
                                                                &
                                                                \begin{tabular}{c} 
                                                                $\mathcal{U}(-0.025,\ 0.025)\;\SI{}{\metre}$\\
                                                                $\mathcal{U}(-0.05,\ 0.05)\;\SI{}{\metre}$\\
                                                                $\mathcal{U}(-0.05,\ 0.05)\;\SI{}{\metre}$\\
                                                                \end{tabular}
                                                                & 
                                                                \begin{tabular}{c} 
                                                                $\mathcal{U}(-0.1,\ 0.1)\;\SI{}{\metre}$\\
                                                                $\mathcal{U}(-0.1,\ 0.1)\;\SI{}{\metre}$\\
                                                                $\mathcal{U}(-0.1,\ 0.1)\;\SI{}{\metre}$\\
                                                                \end{tabular}
                                                                & -- \\  
                        Joint friction                          & 
                                                                \begin{tabular}{c}
                                                                Ankle \\ 
                                                                Waist \\
                                                                \end{tabular}
                                                                &
                                                                \begin{tabular}{c} 
                                                                $\mathcal{U}(0.0,\ 0.05)$ \\
                                                                $\mathcal{U}(0.0,\ 0.1)$ \\
                                                                \end{tabular}                               
                                                                &
                                                                \begin{tabular}{c} 
                                                                $\mathcal{U}(0.0,\ 0.1)$ \\
                                                                -- \\
                                                                \end{tabular}                               
                                                                & -- \\  
\rowcolor[HTML]{EFEFEF} Nominal joint position perturb.         & 
                                                                \begin{tabular}{c}
                                                                Ankle \\
                                                                Others \\
                                                                \end{tabular}                               
                                                                &
                                                                \begin{tabular}{c}
                                                                $\mathcal{U}(-0.03,\ 0.03)$\;\SI{}{\radian} \\
                                                                $\mathcal{U}(-0.015,\ 0.015)$\;\SI{}{\radian} \\
                                                                \end{tabular}
                                                                & --                             & -- \\  
                        Impulsive push                          & 
                                                                \begin{tabular}{c}
                                                                Interval\\
                                                                $\linvel^{x}$,\ $\linvel^{y}$\\
                                                                $\dot{\phi}$,\ $\dot{\theta}$ \\
                                                                \end{tabular}
                                                                &
                                                                \begin{tabular}{c} 
                                                                $\mathcal{U}(0,\ 10)\;\SI{}{\second}$\\
                                                                $\mathcal{U}(-0.3,\ 0.3)\;\SI[per-mode=repeated-symbol]{}{\meter\per\second}$ \\
                                                                $\mathcal{U}(-0.2,\ 0.2)\;\SI[per-mode=repeated-symbol]{}{\radian\per\second}$ \\
                                                                \end{tabular}
                                                                & 
                                                                \begin{tabular}{c} 
                                                                $\mathcal{U}(0,\ 10)\;\SI{}{\second}$\\
                                                                $\mathcal{U}(-0.5,\ 0.5)\;\SI[per-mode=repeated-symbol]{}{\meter\per\second}$ \\
                                                                \\
                                                                \end{tabular}
                                                                &
                                                                \begin{tabular}{c} 
                                                                $\mathcal{U}(0,\ 10)\;\SI{}{\second}$\\
                                                                $\mathcal{U}(-0.5,\ 0.5)\;\SI[per-mode=repeated-symbol]{}{\meter\per\second}$ \\
                                                                \\
                                                                \end{tabular}
                                                                \\ 
\rowcolor[HTML]{EFEFEF} External force to torso                 &
                                                                \begin{tabular}{c}
                                                                Interval\\
                                                                ${}_{\mathcal{I}}{\bm f}^{x}$,\ ${}_{\mathcal{I}}{\bm f}^{y}$,\ ${}_{\mathcal{I}}{\bm f}^{z}$ \\
                                                                \end{tabular}
                                                                &
                                                                \begin{tabular}{c} 
                                                                $\mathcal{U}(2,\ 4)\;\SI{}{\second}$\\
                                                                $\mathcal{U}(-5.0,\ 5.0)\;\SI{}{\newton}$ \\
                                                                \end{tabular}
                                                                & -- & --\\ 
                        External torque to torso                &
                                                                \begin{tabular}{c}
                                                                Interval\\
                                                                ${}_{\mathcal{I}}{\bm \tau}^{x}$,\ ${}_{\mathcal{I}}{\bm \tau}^{y}$,\ ${}_{\mathcal{I}}{\bm \tau}^{z}$ \\
                                                                \end{tabular}
                                                                &
                                                                \begin{tabular}{c} 
                                                                $\mathcal{U}(2,\ 4)\;\SI{}{\second}$\\
                                                                $\mathcal{U}(-5.0,\ 5.0)\;\SI[inter-unit-product = \ensuremath{{}\cdot{}}]{}{\newton\metre}$ \\
                                                                \end{tabular}
                                                                & -- & --\\  
\rowcolor[HTML]{EFEFEF} Initial torso pose perturb.             &   
                                                                \begin{tabular}{c}  
                                                                ${}_\mathcal{I} \position^x$,\ ${}_\mathcal{I} \position^y$,\ ${}_\mathcal{I} \position^z$ \\
                                                                $\phi$,\ $\theta$,\ $\psi$\\
                                                                \end{tabular}
                                                                & 
                                                                \begin{tabular}{c} 
                                                                $\mathcal{U}(-0.1,/ 0.1)$ \SI{}{\metre} \\
                                                                $\mathcal{U}(-0.1,/ 0.1)$ \SI{}{\radian} \\
                                                                \end{tabular}
                                                                & -- & --
                                                                \\ 
                        Initial torso velocity perturb.         &   
                                                                \begin{tabular}{c}  
                                                                ${}_\mathcal{I} \linvel^x$,\ ${}_\mathcal{I} \linvel^y$,\ ${}_\mathcal{I} \linvel^z$ \\
                                                                $\dot\phi$,\ $\dot\theta$,\ $\dot\psi$\\
                                                                \end{tabular}
                                                                & 
                                                                \begin{tabular}{c} 
                                                                $\mathcal{U}(-0.1,/ 0.1)\;\SI[per-mode=repeated-symbol]{}{\metre\per\second}$ \\
                                                                $\mathcal{U}(-0.1,/ 0.1)\;\SI[per-mode=repeated-symbol]{}{\radian\per\second}$ \\
                                                                \end{tabular}
                                                                & 
                                                                \begin{tabular}{c} 
                                                                $\mathcal{U}(-0.1,/ 0.1)\;\SI[per-mode=repeated-symbol]{}{\metre\per\second}$ \\
                                                                $\mathcal{U}(-0.1,/ 0.1)\;\SI[per-mode=repeated-symbol]{}{\radian\per\second}$ \\
                                                                \end{tabular}
                                                                & 
                                                                \begin{tabular}{c} 
                                                                $\mathcal{U}(-0.1,/ 0.1)\;\SI[per-mode=repeated-symbol]{}{\metre\per\second}$ \\
                                                                $\mathcal{U}(-0.1,/ 0.1)\;\SI[per-mode=repeated-symbol]{}{\radian\per\second}$ \\
                                                                \end{tabular}
                                                                \\ 
\rowcolor[HTML]{EFEFEF} Initial joint positions perturb.        &   
                                                                & $\mathcal{U}(-0.2, 0.2)\;\SI{}{\radian}$  
                                                                & $\mathcal{U}(-0.2, 0.2)\;\SI{}{\radian}$  
                                                                & $\mathcal{U}(-0.2, 0.2)\;\SI{}{\radian}$ \\ 
                        Initial joint velocities perturb.       &   
                                                                & $\mathcal{U}(-0.5, 0.5)\;\SI[per-mode=repeated-symbol]{}{\radian\per\second}$ 
                                                                & $\mathcal{U}(-0.5, 0.5)\;\SI[per-mode=repeated-symbol]{}{\radian\per\second}$
                                                                & $\mathcal{U}(-0.5, 0.5)\;\SI[per-mode=repeated-symbol]{}{\radian\per\second}$ \\ 
\rowcolor[HTML]{EFEFEF} Terrain roughness                       &   & $\mathcal{U}(-0.025, 0.025)\;\SI{}{\metre}$  & -- & -- \\   
\bottomrule
\end{tabular}
\fi
\end{table}

%
\begin{table}[ht]
\centering
    \caption{\textbf{Summary of rewards.} Formal definitions for the reward terms are available in \highlight{Section~S3 and S4} of the Supplementary Materials.}
    \label{tab:supplementary_reward_summary}
    \vspace{0.1em}
    \begin{tabular}{lccccc} 
        \toprule
        \multirow{2}{*}{\textbf{Term}}          & \multirow{2}{*}{\textbf{Notation}}    &    \multicolumn{4}{c}{\textbf{Hyperparameter}}                            \\
                                                &                                       &    & \textbf{Unitree G1}       & \textbf{Atlas R1}   & \textbf{Atlas D1}   \\                                     
        \midrule
        \multicolumn{6}{c}{\textbf{Reference Tracking} \; $\rtrack$} \\
        \midrule
        \rowcolor[HTML]{EFEFEF} Ref. torso position tracking             & $r_{t,\position}$                    & 
                                                                                                                \begin{tabular}{c}
                                                                                                                $c_{t,{\position}}$   \\
                                                                                                                $\sigma_{\position, \text{min}}$
                                                                                                                \end{tabular}
                                                                                                                &
                                                                                                                --
                                                                                                                &
                                                                                                                \begin{tabular}{c}
                                                                                                                1.0\\
                                                                                                                0.4
                                                                                                                \end{tabular}
                                                                                                                &
                                                                                                                \begin{tabular}{c}
                                                                                                                1.0\\
                                                                                                                0.4
                                                                                                                \end{tabular}
                                                                                                                \\
                                Ref. torso orientation tracking         & $r_{t,\Phi}$                          &
                                                                                                                \begin{tabular}{c}
                                                                                                                $c_{t,{\Phi}}$   \\
                                                                                                                $\sigma_{\Phi, \text{min}}$
                                                                                                                \end{tabular}
                                                                                                                &
                                                                                                                --
                                                                                                                &
                                                                                                                \begin{tabular}{c}
                                                                                                                1.0\\
                                                                                                                0.5
                                                                                                                \end{tabular}
                                                                                                                &
                                                                                                                \begin{tabular}{c}
                                                                                                                1.0\\
                                                                                                                0.5
                                                                                                                \end{tabular}
                                                                                                                \\                                                                                                        
        \rowcolor[HTML]{EFEFEF} Ref. torso linear velocity tracking     & $r_{t,\linvel}$                       & 
                                                                                                                \begin{tabular}{c}
                                                                                                                $c_{t,{\linvel}}$   \\
                                                                                                                $\sigma_{\linvel, \text{min}}$
                                                                                                                \end{tabular}
                                                                                                                &
                                                                                                                \begin{tabular}{c}
                                                                                                                1.0\\
                                                                                                                0.2
                                                                                                                \end{tabular}
                                                                                                                &
                                                                                                                \begin{tabular}{c}
                                                                                                                1.0\\
                                                                                                                0.2
                                                                                                                \end{tabular}
                                                                                                                &
                                                                                                                \begin{tabular}{c}
                                                                                                                1.0\\
                                                                                                                0.2
                                                                                                                \end{tabular}
                                                                                                                \\
                                Ref. torso angular velocity tracking    & $r_{t,\angvel}$                       &
                                                                                                                \begin{tabular}{c}
                                                                                                                $c_{t,{\angvel}}$   \\
                                                                                                                $\sigma_{\angvel, \text{min}}$
                                                                                                                \end{tabular}
                                                                                                                &
                                                                                                                \begin{tabular}{c}
                                                                                                                1.0\\
                                                                                                                0.2
                                                                                                                \end{tabular}
                                                                                                                &
                                                                                                                \begin{tabular}{c}
                                                                                                                1.0\\
                                                                                                                0.2
                                                                                                                \end{tabular}
                                                                                                                &
                                                                                                                \begin{tabular}{c}
                                                                                                                1.0\\
                                                                                                                0.2
                                                                                                                \end{tabular}
                                                                                                                \\
        \rowcolor[HTML]{EFEFEF} Ref. Torso projected gravity tracking   & $r_{t,{\bm g}}$                       &
                                                                                                                \begin{tabular}{c}
                                                                                                                $c_{t,{\bm g}}$   \\
                                                                                                                $\sigma_{\bm g, \text{min}}$
                                                                                                                \end{tabular}
                                                                                                                &
                                                                                                                \begin{tabular}{c}
                                                                                                                1.0\\
                                                                                                                0.2
                                                                                                                \end{tabular}
                                                                                                                &
                                                                                                                \begin{tabular}{c}
                                                                                                                1.0\\
                                                                                                                0.2
                                                                                                                \end{tabular}
                                                                                                                &
                                                                                                                \begin{tabular}{c}
                                                                                                                1.0\\
                                                                                                                0.2
                                                                                                                \end{tabular}
                                                                                                                \\
                                Ref. Joint position tracking            & $r_{t,j}$                             & 
                                                                                                                \begin{tabular}{c}
                                                                                                                $c_{t,{\bm g}}$   \\
                                                                                                                $\sigma_{\bm g, \text{min}}$
                                                                                                                \end{tabular}
                                                                                                                &
                                                                                                                \begin{tabular}{c}
                                                                                                                0.5\\
                                                                                                                0.2
                                                                                                                \end{tabular}
                                                                                                                &
                                                                                                                \begin{tabular}{c}
                                                                                                                0.5\\
                                                                                                                0.2
                                                                                                                \end{tabular}
                                                                                                                &
                                                                                                                \begin{tabular}{c}
                                                                                                                0.5\\
                                                                                                                0.2
                                                                                                                \end{tabular}
                                                                                                                \\
        \midrule
        \multicolumn{6}{c}{\textbf{Regularization} \; $\rreg$} \\
        \midrule
        \rowcolor[HTML]{EFEFEF} Action smoothness penalty   & $r_{\text{reg},{\Delta \bm a}}$       & $c_{\text{reg},{\Delta \bm a}}$       & $-0.5$    & $-0.5$    & $-0.5$    \\
                                Joint torque penalty        & $r_{\text{reg},{\bm{\tau}_j}}$        & $c_{\text{reg},{\bm{\tau}_j}}$        & $-0.0005$ & $-0.0005$ & $-0.0005$ \\
        \rowcolor[HTML]{EFEFEF} Feet slide penalty          & $r_{\text{reg},{\text{slide}}}$       & $c_{\text{reg},{\text{slide}}}$       & $-0.25$   & $-0.25$   & $-0.25$   \\
                                Feet jerk penalty           & $r_{\text{reg},{\jerk}}$              & $c_{\text{reg},{\jerk}}$              & $-0.001$  & $-0.001$  & $-0.001$  \\
        \rowcolor[HTML]{EFEFEF} Feet clearance              & $r_{\text{reg},\text{clr}}$           &  
                                                                                                    \begin{tabular}{c}
                                                                                                    $c_{\text{reg},\text{clr}}$                     \\
                                                                                                    ${}_\mathcal{I}\hat\position^z_{\text{clr}}$    \\
                                                                                                    $\lambda_{\text{clr},v}$                        \\
                                                                                                    $\lambda_{\text{clr},e}$
                                                                                                    \end{tabular}
                                                                                                    &            
                                                                                                    \begin{tabular}{c}
                                                                                                    3.0     \\
                                                                                                    0.06    \\
                                                                                                    2.0     \\
                                                                                                    0.0009  
                                                                                                    \end{tabular}
                                                                                                    & --        & --   \\
        \midrule
        \multicolumn{6}{c}{\textbf{Survival} \; $\rsurvival$} \\
        \midrule
        \rowcolor[HTML]{EFEFEF} Survival                            & $\rsurvival$ & $c_{\text{survival}}$     & 30 & 30 & 30    \\
        \midrule                                
        \multicolumn{6}{c}{\textbf{Command Tracking} \; $\rcommand$} \\
        \midrule
        \rowcolor[HTML]{EFEFEF} Linear velocity command tracking    & $r_{\text{cmd}, v}$           & 
                                                                                                    \begin{tabular}{c}
                                                                                                    $c_{\text{cmd}, v}$   \\
                                                                                                    $\lambda_{\text{cmd}, v}$
                                                                                                    \end{tabular}
                                                                                                    &
                                                                                                    \begin{tabular}{c}
                                                                                                    5.0   \\
                                                                                                    0.25
                                                                                                    \end{tabular}
                                                                                                    &
                                                                                                    \begin{tabular}{c}
                                                                                                    5.0   \\
                                                                                                    0.25
                                                                                                    \end{tabular}
                                                                                                    & 
                                                                                                    \begin{tabular}{c}
                                                                                                    5.0   \\
                                                                                                    0.25
                                                                                                    \end{tabular}
                                                                                                    \\
                                Turn rate command tracking          & $r_{\text{cmd}, \omega}$      & 
                                                                                                    \begin{tabular}{c}
                                                                                                    $c_{\text{cmd}, \omega}$   \\
                                                                                                    $\lambda_{\text{cmd}, \omega}$
                                                                                                    \end{tabular}
                                                                                                    &
                                                                                                    \begin{tabular}{c}
                                                                                                    5.0   \\
                                                                                                    0.25
                                                                                                    \end{tabular}
                                                                                                    &
                                                                                                    \begin{tabular}{c}
                                                                                                    5.0   \\
                                                                                                    0.25
                                                                                                    \end{tabular}
                                                                                                    & 
                                                                                                    \begin{tabular}{c}
                                                                                                    5.0   \\
                                                                                                    0.25
                                                                                                    \end{tabular} 
                                                                                                    \\
        \bottomrule
    \end{tabular}
\end{table}

%
\begin{table}[ht]
\centering
\caption{\textbf{Unitree G1 joint gains and action specifications.} The action vector output by the policy is element-wise multiplied by the action scaling factors (action scalers) and offset by the nominal joint configurations to yield the desired target positions for the low-level, joint-level PD controllers, utilizing the specific gain parameters detailed below.}
\label{tab:supplementary_action}
\vspace{0.1em}
\begin{tabular}{ lcccc}
\toprule
                                        & \textbf{P gain}   
                                        & \textbf{D gain}   
                                        & \makecell{\textbf{Nominal angle} \\ $\left[ \SI{}{\radian} \right]$ }                
                                        & \makecell{\textbf{Action scaling} \\ \textbf{factors}}\\  
                                        
\midrule
                        Waist roll      & 150.0             & 3.0               & 0.0                                   & 0.1\\  
\rowcolor[HTML]{EFEFEF} Waist pitch     & 150.0             & 3.0               & 0.0                                   & 0.1\\  
                        Waist yaw       & 100.0             & 2.0               & 0.0                                   & 0.1\\  
\rowcolor[HTML]{EFEFEF} Hip roll        & 40.0              & 3.0               & 0.0                                   & 0.2\\  
                        Hip pitch       & 40.0              & 3.0               & -0.1                                  & 0.2\\  
\rowcolor[HTML]{EFEFEF} Hip yaw         & 40.0              & 3.0               & 0.0                                   & 0.1\\  
                        Knee            & 80.0              & 2.5               & 0.2                                   & 0.2\\  
\rowcolor[HTML]{EFEFEF} Ankle roll      & 40.0              & 2.0               & 0.0                                   & 0.05\\  
                        Ankle pitch     & 40.0              & 2.0               & -0.1                                  & 0.2\\  
\rowcolor[HTML]{EFEFEF} Shoulder roll   & 20.0              & 1.0               & 0.25 (left), -0.25 (right)            & 0.1\\  
                        Shoulder pitch  & 20.0              & 1.0               & 0.1                                   & 0.1\\  
\rowcolor[HTML]{EFEFEF} Shoulder yaw    & 20.0              & 1.0               & 0.0                                   & 0.1\\  
                        Elbow           & 30.0              & 1.0               & 1.0                                   & 0.1\\  
\rowcolor[HTML]{EFEFEF} Wrist roll      & 20.0              & 0.5               & 0.0                                   & 0.01\\  
                        Wrist pitch     & 20.0              & 0.5               & 0.0                                   & 0.01\\  
\rowcolor[HTML]{EFEFEF} Wrist yaw       & 20.0              & 0.5               & 0.0                                   & 0.01\\  
\bottomrule
\end{tabular}
\end{table}

%
\begin{table}[ht]
\centering
\caption{\textbf{Assistive wrench curriculum parameters.} }
\label{tab:curriculum}
\vspace{0.1em}
\begin{tabular}{lc}
\toprule
                        \textbf{Parameter}                             & \textbf{Value} \\
\midrule
                        Force PD gains \;$(k_p^v, k_d^v)$              & $(0.0, 10.0)$\\
\rowcolor[HTML]{EFEFEF} Torque PD gains \;$(k_p^\omega, k_d^\omega)$   & $(200.0, 10.0)$\\
                        Max curriculum gain \;$\beta_{\text{max}}$     & 0.75  \\
\rowcolor[HTML]{EFEFEF} Curriculum gain incremental step \; $\Delta\beta_\uparrow$ & 0.05  \\
                        Curriculum gain decremental step \; $\Delta\beta_\downarrow$ & 0.1  \\
\bottomrule
\end{tabular}
\end{table}

%
%
\begin{table}[ht]
\small
\centering
    \caption{\textbf{Training hyperparameters.}}
    \label{tab:supplementary_ppo_hp}
    \vspace{0.1em}
    \begin{tabular}{lc} 
        \toprule
        \textbf{Hyperparameter}                             & \textbf{Value} \\
        \midrule
        \multicolumn{2}{c}{\textbf{Teacher policy training (PPO)}} \\
        \midrule
        \rowcolor[HTML]{EFEFEF} Teacher policy network      & MLP(1024, 512, 256) with \emph{ELU} activation \\
                                Critic network              & MLP(512, 512, 256)  with \emph{ELU} activation \\
        \rowcolor[HTML]{EFEFEF} Empirical normalization     & True \\
                                Learning rate               & 
                                                            \begin{tabular}{c}
                                                            1e-3, adaptive based on KL-divergence for flat terrain \\
                                                            1e-5, fixed for rough terrain \\
                                                            \end{tabular} \\
        \rowcolor[HTML]{EFEFEF} Discount factor             & 0.99 \\
                                GAE discount factor         & 0.95 \\
        \rowcolor[HTML]{EFEFEF} Desired KL-divergence       & 0.01 \\
                                Clip range                  & 0.2 \\
        \rowcolor[HTML]{EFEFEF} Entropy coefficient         & 0.0015 \\
                                Value function loss coefficient & 0.5 \\
        \rowcolor[HTML]{EFEFEF} Number of epochs            & 5 \\
                                Number of environments      & 4096 \\
        \rowcolor[HTML]{EFEFEF} Batch size                  & 98,304 $(4096 \times 24)$ \\
                                Mini-batch size             & 24,576 (Number of mini-batch = 4) \\       
        \rowcolor[HTML]{EFEFEF} Initial action noise std.   & 1.0 \\
                                Maximum episode length      & 500 (\SI{10}{\second}) \\
        \midrule
        \multicolumn{2}{c}{\textbf{Prior policy training (Teacher-Student distillation)}} \\
        \midrule
        \rowcolor[HTML]{EFEFEF} Prior policy network        & MLP(1024, 1024, 1024) with \emph{ELU} activation \\
                                Empirical normalization     & True \\
        \rowcolor[HTML]{EFEFEF} Learning Rate               & 1e-6, fixed \\
                                Number of Environments      & 4096 \\
        \rowcolor[HTML]{EFEFEF} Batch Size                  & 98,304 $(4096 \times 24)$ \\
                                Mini-Batch Size             & 3,072 (Number of mini-batch = 32) \\       
        \rowcolor[HTML]{EFEFEF} Maximum episode length      & 150 (\SI{3}{\second}) \\
        \midrule
        \multicolumn{2}{c}{\textbf{RL fine-tuning (PPO)}} \\
        \midrule
        \rowcolor[HTML]{EFEFEF} Empirical normalization     & True \\
                                Learning rate               & 1e-6, fixed \\
        \rowcolor[HTML]{EFEFEF} Discount factor             & 0.99 \\
                                GAE discount factor         & 1.0 \\
        \rowcolor[HTML]{EFEFEF} Clip range                  & 0.2 \\
                                Entropy coefficient         & 0.001 \\
        \rowcolor[HTML]{EFEFEF} Value function loss coefficient & 0.5 \\
                                Number of epochs            & 5 \\
        \rowcolor[HTML]{EFEFEF} Number of environments      & 4096 \\
                                Batch size                  & 98,304 $(4096 \times 24)$ \\
        \rowcolor[HTML]{EFEFEF} Mini-batch size             & 24,576 (Number of mini-batch = 4)  \\
                                Initial action noise std.   & 0.16 \\
        \rowcolor[HTML]{EFEFEF} Maximum episode length      & 
                                                            \begin{tabular}{c}
                                                            150 (\SI{3}{\second}) for reference-guided imitation task \\
                                                            400 (\SI{8}{\second}) for goal-conditioned task \\
                                                            \end{tabular} \\ 
        \bottomrule
    \end{tabular}
\end{table}

\end{document}